\documentclass[letterpaper,journal]{IEEEtran}
\usepackage{amsmath,amsfonts}
\usepackage{algorithmic}
\usepackage{algorithm}
\usepackage{array}
\usepackage[caption=false,font=normalsize,labelfont=sf,textfont=sf]{subfig}
\usepackage{textcomp}
\usepackage{stfloats}
\usepackage{url}
\usepackage{verbatim}
\usepackage{graphicx}
\usepackage{cite}
\usepackage{booktabs}
\usepackage{multirow}
\begin{document}

\title{HistRetinex: Optimizing Retinex model in Histogram \\ Domain for Efficient Low-Light Image Enhancement}

\author{Jingtian Zhao, Xueli Xie, Jianxiang Xi, Xiaogang Yang, Haoxuan Sun
\thanks{This work was supported by the National Natural Science Foundation of China under Grants U23B2064, 62176263, 62103434, and 62003363, Youth Talent Promotion Program of Shaanxi Provincial Association for Science and Technology under Grant 20220123, Natural Science Basic Research Program of Shaanxi Province under Grant 2022KJXX-99, National Defense Basic Research Program of Technology and Industry for National Defense under Grant JCKY2021912B001. (Corresponding author: Jianxiang Xi). 
}
\thanks{J.T. Zhao, J.X. Xi, X.L. Xie and X.G. Yang are with the Rocket Force University of Engineering, Xi’an 710025, P.R. China (jingtianzhao@qq.com; xuelixie@163.com; xijx07@mails.tsinghua.edu.cn; doctoryxg@163.com)}
\thanks{Color versions of one or more of the figures in this article are available online at http://ieeexplore.ieee.org}
}

\markboth{IEEE Transactions on Image Processing}%
{Shell \MakeLowercase{\textit{\textit{et al.}}}: A Sample Article Using IEEEtran.cls for IEEE Journals}


\maketitle

\begin{abstract}
Retinex-based low-light image enhancement methods are widely used due to their excellent performance. However, most of them are time-consuming for large-sized images. This paper extends the Retinex model from the spatial domain to the histogram domain, and proposes a novel histogram-based Retinex model for fast low-light image enhancement, named HistRetinex. Firstly, we define the histogram location matrix and the histogram count matrix, which establish the relationship among histograms of the illumination, reflectance and the low-light image. Secondly, based on the prior information and the histogram-based Retinex model, we construct a novel two-level optimization model. Through solving the optimization model, we give the iterative formulas of the illumination histogram and the reflectance histogram, respectively. Finally, we enhance the low-light image through matching its histogram with the one provided by HistRetinex. Experimental results demonstrate that the HistRetinex outperforms existing enhancement methods in both visibility and performance metrics, while executing 1.86 seconds on 1000×664 resolution images, achieving a minimum time saving of 6.67 seconds.
\end{abstract}


\section{Introduction}
\IEEEPARstart{L}{OW-LIGHT} image enhancement is a critical technology in computer vision, which is used to optimize the color distribution and balance brightness to enhance overall contrast. The image enhancement methods can extract extra visual details from low-light images and facilitate subsequent advanced image processing. The application scenarios of low-light image enhancement include autonomous driving [1], medical applications [2]-[3] and lane detection [4], etc. Currently, the requirements for real-time ability and accuracy in above scenarios have put forward higher demands on the performance and computational efficiency of low-light image enhancement algorithms.

\begin{figure}[!t]
\centering
\includegraphics[width=3.5in]{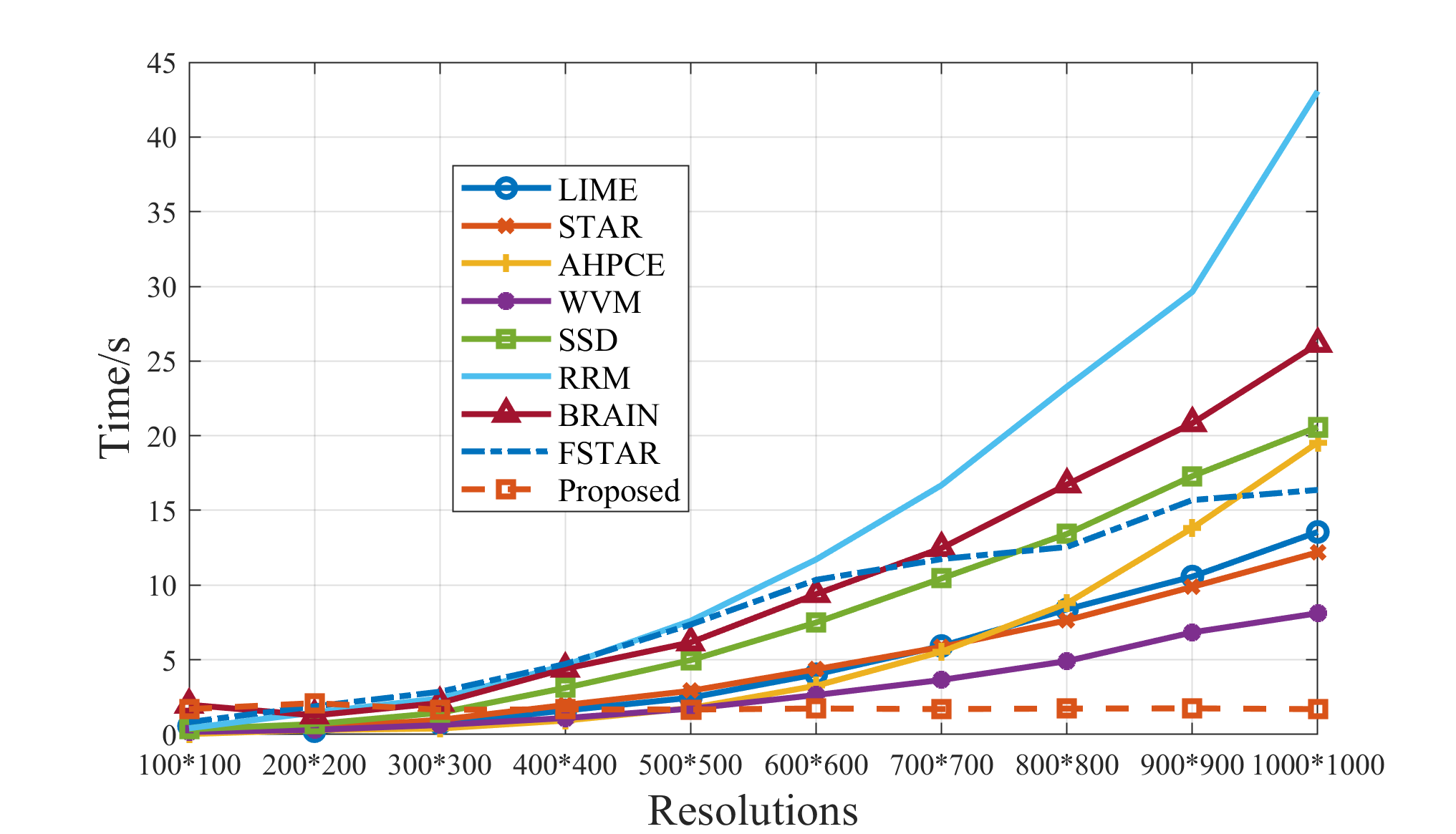}
\caption{Execution time of existing low-light enhancement algorithms in low-light images with different resolutions.}
\label{fig_1}
\end{figure}

Recently, the low-light image enhancement undergoes rapid development. The popular low-light image enhancement methods mainly include learning-based methods [5], Retinex-based methods [6]-[7] and histogram-based methods [8]. Compared with learning-based methods and histogram-based methods, Retinex-based methods do not require prior training and receive widespread attention due to their better interpretability and performance. The Retinex theory was first proposed by Land and McCann in [9], which decomposes a scene into illuminance and reflectance components and realizes visual enhancement through the illumination adjustment. This theory is widely applied in various fields, including scene enhancement [10]-[11] and color correction [12]-[13]. Actually, Retinex-based methods include two classes: classical ones and variational ones. In particular, the variational methods [14]-[15] formulate image decomposition as a variational optimization problem, and construct prior terms to smooth the illumination component and to sparsity the reflectance component, which avoids over-enhancement and has detail preservation ability. Kimmel \textit{et al.} [14] designed an objective function that incorporates the regularization term about the illumination component, where the illumination can be estimated and integrated with the illumination adjustment for image enhancement by solving the related optimization model. This approach is a new paradigm for Retinex-based methods. Similarly, Guo \textit{et al.} [15] proposed the low-light image enhancement via illumination map estimation (LIME), which constructs an objective function to estimate the illumination component and utilizes the fast Fourier transform (FFT) to obtain the solution fast. Based on the above research foundations, many improvements are made in noise removal [16]-[17], enhancement performance [18]-[19] and computational efficiency [20]-[21], etc.

Although the variational Retinex-based algorithms achieve excellent enhancement performance and computational efficiency, they have a common limitation: the long computational time when the algorithms processing the large-scale images. As shown in Fig. 1, the RRM requires merely 0.27s for the 100×100 resolution images but needs 43s for the 1000×1000 resolution images. It is clear that the computational time of existing algorithms grows substantially with the increasing image resolution. The main reason is that existing Retinex-based enhancement algorithms directly process each pixel in the image, which make computational complexity be strongly correlated with the image resolution. In the other head, these methods typically employ the alternating direction multiplier method (ADMM) in [22] to optimize the objective function, which requires multiple iterations to converge the local optimal solution. Therefore, we need to propose a new method to fast realize large-size low-light image enhancement.

Histogram processing handles the images in the histogram domain rather than the spatial domain and maps the results back to the spatial domain to enhance the image. Compared with the Retinex-based methods, this method has high computational efficiency due to the traditional histogram generally only contain 256 gray levels. Based on this method, we transform the Retinex model in the histogram domain and replace the spatial-domain matrix multiplication by the histogram-domain vector operation.

Specifically, we construct a corresponding optimization model in the histogram domain and derive iterative formulas for both histograms of the illumination and the reflectance through systematic solving the optimization model. This domain transformation reduces the computational complexity of the proposed algorithm from the pixel-related one to the grayscale-related one, where the grayscale number much smaller than the pixel number. This method significantly improves computational efficiency and has promising enhancement effect. Furthermore, the HistRetinex inherently preserves the fundamental properties of the Retinex model and has promising scalability and superior enhancement performance.

The contributions of this work are summarized as follows:

1.We define the histogram count matrix and the histogram location matrix to describe the relationship among the illumination, the reflectance and the low-light image. Furthermore, the histogram count matrix is approximated and classified based on the histogram location matrix to estimate the histogram of the low-light image.

2. Based on the histogram-based Retinex model and the prior information, we construct a novel two-level optimization model. Through solving the optimization model, the iterative formulas of the illumination histogram and the reflectance histogram are provided.

3. We realize the Retinex model in the histogram domain and propose a new efficient histogram-based Retinex model, which consumes 1.86s for images with the 1000×664 resolution and saves 6.67s compared with other related algorithms with the shortest processing time.

The rest of this paper is organized as follows. Section II gives the Retinex-based image enhancement and histogram equalization algorithms. Section III constructs a histogram-based Retinex model and an associated optimization model. Section IV analyzes the experimental results. The conclusion of this paper is shown in Section V.

\section{Related Work}

In this section, we introduce the theory related to Retinex-based enhancement algorithms and the motivation for using histogram. The notation $\boldsymbol{I}$ is the test image, the notation $\boldsymbol{S}$ is the value channel of the test image $\boldsymbol{I}$ in the HSV (hue, saturation, value) color space and the notation $l$ is the number of color levels in  $\boldsymbol{S}$.

\subsection{Retinex-based Low-Light Image Enhancement Method}

In the classic Retinex model, it assumes that the $\boldsymbol{S}$ can be decomposed into the illumination component $\boldsymbol{L}$ and the reflectance component $\boldsymbol{R}$, which is given as follows:
\begin{equation}
\label{deqn_ex1a}
\boldsymbol{S}=\boldsymbol{R}\circ\boldsymbol{L},
\end{equation}
where the $\circ $ represents the element-wise multiplication. The objective function of the Retinex model is given by:
\begin{equation}
\label{deqn_ex1a}
\underset{\boldsymbol{R},\boldsymbol{L}}{\mathop{\min }}\,||\boldsymbol{S}-\boldsymbol{R}\circ \boldsymbol{L}||_{F}^{2}+\alpha {{\mathcal{R}}_{1}}(\boldsymbol{R})+\beta {{\mathcal{R}}_{2}}(\boldsymbol{L}),
\end{equation}
where the parameters $\alpha $ and $\beta $ are the positive constants and weight parameters of prior terms about $\boldsymbol{R}$ and $\boldsymbol{L}$, respectively. The prior term ${{\mathcal{R}}_{1}}(\boldsymbol{R})$ ensures that the reflectance component is close to the gradient of the $\boldsymbol{S}$ channel, and the prior term ${{\mathcal{R}}_{2}}(\boldsymbol{L})$ ensure that the illumination component is smooth.

(1) Low-light images usually contain the noise, and the classical methods may amplify noise and generate artifacts when these methods are used to enhance brightness or contrast, which may result in blurred or distorted details; (2) the classical methods cannot retain the details in the dark regions without overexposing bright regions. Next, we introduce the existing algorithms that address the above problems.

To solve the first challenge, Li \textit{et al.} [23] incorporated the noise component in the Retinex model and proposed a robust Retinex model (RRM), which removes the noise in the low-light image and has certain noise suppression ability. Ren \textit{et al.} [24] presented a low-rank regularized Retinex model (LR3M), which incorporates a low-rank prior into the Retinex model to suppress noise in the reflectance component and employs the piece-wise smoothed illumination to further eliminate the noise in both the reflectance and illumination components, respectively. Meanwhile, Hao \textit{et al.} [25] designed a Gaussian total variation (GTV) filter to estimate the illumination component and constructed a noise-related Retinex model to suppress the noise of the reflectance component. Du \textit{et al.} [26] introduced the patch-aware low-rank model (PALR) and proposed a dual-constrained Retinex decomposition model (DCRD). The PALR evaluates the noise level for image patches, and control the regularization extent to remove noise. To conquer the second challenge, Ng and Wang [27] introduced the reflectance component into the objective function and proposed a Retinex-based total variation model (TVM), which has better enhancement performance. However, the TVM loses some detail-preservation due to the side effect of the logarithmic transformation. Xu \textit{et al.} [28] employed an exponentialized mean local variance (EMLV) filter to extract structure and texture maps from low-light images and proposed the structure and texture aware Retinex (STAR) algorithm. The STAR achieves superior visual quality and detail preservation. Cai and Chen [29] extracted image edges and separated them into the contour and texture gradient, and incorporated these image details as the prior information to construct a variational Retinex model. The proposed Brian-like Retinex (BRAIN) model has satisfactory scalability and enhancement performance. Li and He [30] proposed a fractional structure and texture-aware Retinex (FSTAR) algorithm, which introduces the maximum fractional difference as a structure-aware measure and constructs fractional structure-texture maps as weighting matrices in the regularization terms for the illumination and reflectance components. The FSTAR achieves robust enhancement effects for the noise of image and maintains an appropriate illumination range in the low-light images.

Although abovementioned algorithms were widely used due to their desired enhancement effect and extensibility, most of them still need a long execution time for the large-size image. The main reason is that they calculate the illumination and reflectance for each pixel during each iteration, which consumes much time for large-scale images. On the other hand, most of existing Retinex-based methods provide enhancement results with halo artifacts. This is because the blurred prior of illumination component captures redundant reflectance information in the areas of significant illumination change. Therefore, our method decomposes the image into the illumination and reflectance components in the histogram domain and achieves computational complexity independent of the image resolution. Furthermore, we propose the histogram decomposition method, which use vector cross product and classification to separatee the low-light image into the illumination and reflectance components in the histogram domain. This approach eliminates the artifacts caused by the spatial-domain matrix multiplication in traditional approaches. 

\subsection{Histogram-based Low-light Image Enhancement Method}
The Histogram equalization (HE) is a classical image enhancement method, which has high efficiency and enhancement performance to some extent. The HE employs the cumulative distribution function (CDF) to redistribute intensity values and transforms the original probability density function (PDF) into an approximately uniform distribution. This process effectively enhances visibility in the dark regions of the low-light images.

For a digital image with intensity values in the range $[0,l-1]$, its histogram is defined as:
\begin{equation}
\label{deqn_ex1a}
{{H}_{C}}{{(\boldsymbol{S})}_{i}}={{n}_{i}},i\in [0,1,...,l-1],
\end{equation}
where the ${{n}_{i}}$ denotes the number of pixels in the $\boldsymbol{S}$, whose gray value is consistent with  $i$-th gray level.

For the histogram equalization, we first normalize the histogram ${{H}_{C}}(\boldsymbol{S})$ as follows:

\begin{equation}
\label{deqn_ex1a}
{{P}_{C}}{{(\boldsymbol{S})}_{i}}=\frac{{{(\boldsymbol{S})}_{i}}}{N},i\in [0,1,...,l-1],
\end{equation}
where the $N$ denotes the number of pixels in image $\boldsymbol{I}$. The ${{P}_{C}}{{(\boldsymbol{S})}_{i}}$ presents the probability of the $i$-th gray level in the value channel $\boldsymbol{S}$. Then, based on the probability of all gray levels, the histogram equalization constructs a transform function:
\begin{equation}
\label{deqn_ex1a}
{{t}_{i}}=(l-1)\sum\limits_{j=0}^{i}{{{P}_{C}}{{(\boldsymbol{S})}_{j}}},i\in [0,1,...,l-1],
\end{equation}
where the ${{t}_{i}}$ represents a new colorscale value. For a digital image, the old colorscale value ${{r}_{i}}$ always is a integer, and ${{t}_{i}}$ is a radix. We need to round ${{t}_{i}}$ to the nearest old color scale value, and assign each element of ${{H}_{C}}(\boldsymbol{S})$ according to the new color scale value. The proposed algorithm is motivated by this idea of redistribution after approximation for histogram. Next, we will introduce research work of the HE in recent years.

Kim [31] proposed the brightness preserving bi-histogram equalization (BBHE), which constructs two histograms according to the mean intensity of the image and equalizes two sub-histograms independently. It preserves the mean brightness in original image and has enhancement effect to a certain extent. Based on the BBHE, Wang \textit{et al.} [32] proposed the equal area dualistic sub-image histogram equalization (DSIHE), which has better performance and provides the results with more appropriate contrast. In response to the DSIHE, Chen and Ramli [33] utilized the bi-HE (BBHE) to propose a minimum mean brightness error bi-histogram equalization (MMBEBHE). The MMBEBHE sets a threshold level to separate the histogram and yield the minimum absolute mean rightness error. The MMBEBHE avoid the artifacts in the enhanced image and has better enhancement effect. Chen and Ramli [34] also separated the histogram and proposed the recursive mean separate histogram equalization (RMSHE). The RMSHE provides better visual effect since it permits the scalable preservation. Wan \textit{et al.} [35] proposed the adaptive histogram partition and brightness correction enhancement (AHPCE), which generates the foreground and background sub-histograms using a weighted scatter plot smoothing algorithm and processes the corrected mean brightness by utilizing the particle swarm optimization algorithm.

Compared with the Retinex-based methods, the histogram-based methods achieve superior computational efficiency by operating on the histogram of the low-light image rather than operating the all-pixels of low-light image. However, the HE suffers from information loss due to its merging of the distinct gray levels and has unsatisfactory enhancement quality. To improve the enhancement quality of the HE-based methods and the efficiency of Retinex-based methods, we decompose the image into the illumination and reflectance components in the histogram domain, and merge and enhance the gray levels of the illumination components rather than the gray levels of the low-light image. This strategy preserves the key structural information in the image and can achieves effective enhancement.

\section{Porposed Algorithm}
\subsection{Histogram-based Retinex Model}
To extend the classic Retinex model and improve computational efficiency, we plan to transform the Retinex model to the histogram domain and use the histograms of the reflectance and illumination components to estimate the histogram of the original image. The framework of the proposed model is shown in Fig. 2. The classic Retinex model processes the illumination and reflectance components through the spatial-domain matrix multiplication to reconstruct the original image. However, the histogram merely represents the statistical distributions of color intensities and inherently lacks the spatial information. This critical limitation prevents the identification of correspondences between specific reflectance and the illumination levels during the component recombination. Currently, no existing method can effectively replicate the spatial-domain matrix multiplication in the histogram domain, which is the fundamental obstacle to preventing the transformation of Retinex model from the spatial domain to the histogram domain.

\begin{figure}[!t]
\centering
\includegraphics[width=3.5in]{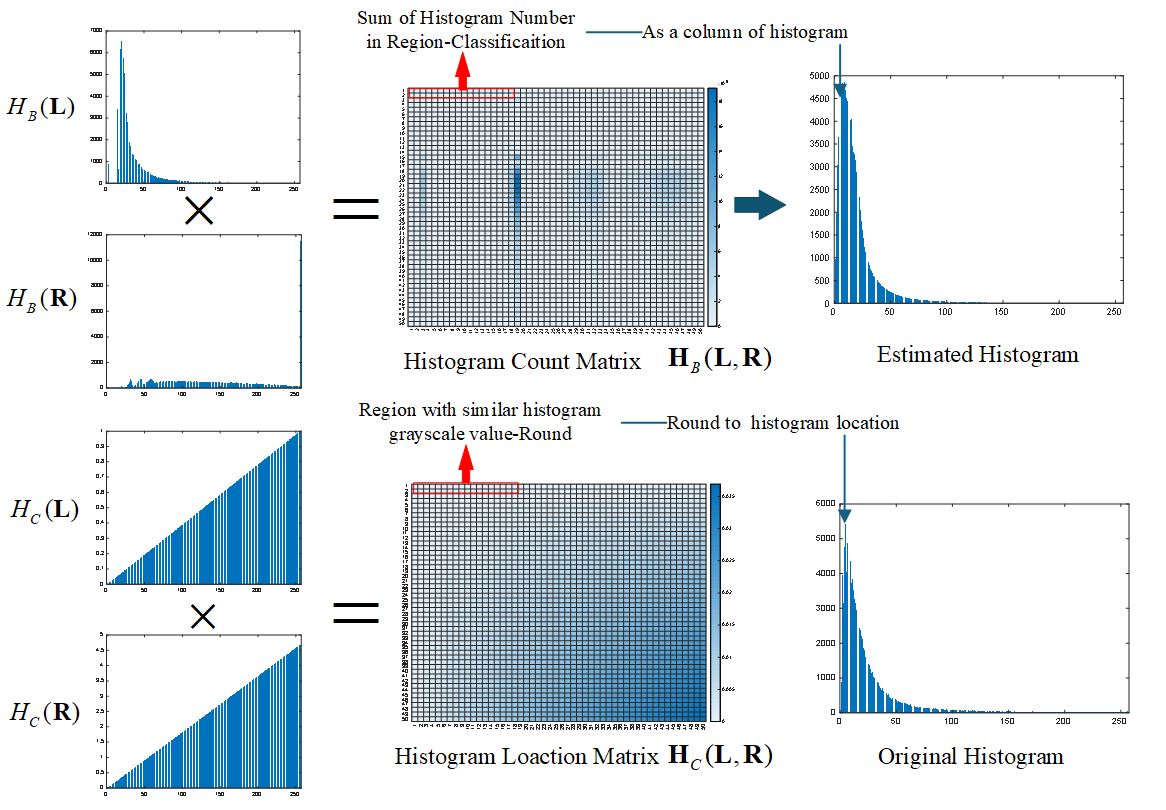}
\caption{The framework of the histogram-based Retinex model.}
\label{fig_1}
\end{figure}

To address this fundamental challenge, we introduce two novel matrix representations to describe the relationship between the illumination component and the reflectance component: the Histogram Location Matrix and Histogram Count Matrix. The definitions of these matrices are as follows:

\begin{figure*}[!t]
\centering
\subfloat{\includegraphics[width=1.3in]{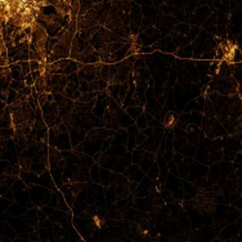}%
\label{fig_first_case}}
\hfil
\subfloat{\includegraphics[width=1.3in]{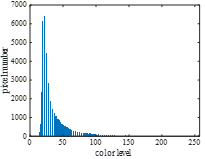}%
\label{fig_second_case}}
\hfil
\subfloat{\includegraphics[width=1.3in]{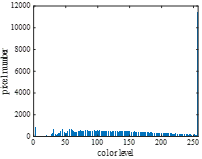}%
\label{fig_second_case}}
\hfil
\subfloat{\includegraphics[width=1.3in]{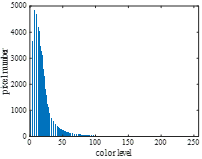}%
\label{fig_second_case}}
\hfil
\subfloat{\includegraphics[width=1.3in]{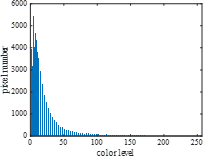}%
\label{fig_second_case}}

\makebox[1.3in][c]{(a)}%
\hspace{0.1in}  
\makebox[1.3in][c]{(b)}%
\hspace{0.1in}
\makebox[1.3in][c]{(c)}%
\hspace{0.1in}
\makebox[1.3in][c]{(d)}%
\hspace{0.1in}
\makebox[1.3in][c]{(e)}
\caption{Histograms of various components in the histogram-based Retinex model. (a) original image; (b) illumination component; (c) reflectance component; (d) estimate V channel component; (e) original V channel component.}
\label{fig_sim}
\end{figure*}

\begin{equation}
\label{deqn_ex1a}
	\boldsymbol{H}_{C}(\boldsymbol{L}, \boldsymbol{R})=H_{C}(\boldsymbol{L}) \times H_{C}(\boldsymbol{R})^{\mathrm{T}} / N,
\end{equation}
\begin{equation}
\label{deqn_ex1a}
\boldsymbol{H}_{B}(\boldsymbol{L}, \boldsymbol{R})=H_{B}(\boldsymbol{L}) \times H_{B}(\boldsymbol{R})^{\mathrm{T}} / N,
\end{equation}
where the ${{H}_{C}}(\cdot )$ and ${{H}_{B}}(\cdot )$ represent the functions, which calculate the count histogram and the location histogram of the image, respectively. The calculated result is a column vector. Through the vector product, the histogram count matrix ${{\boldsymbol{H}}_{C}}(\boldsymbol{L},\boldsymbol{R})$ and histogram location matrix ${{\boldsymbol{H}}_{B}}(\boldsymbol{L},\boldsymbol{R})$ construct the relationship between the illumination and reflectance component, which is the multiplication of each gray level of the illumination component by per gray level of the reflectance component. The graphic representation of the histogram count matrix and histogram location matrix is shown in Fig. 2.

Then, we classify the histogram location matrix into all color levels of the original image. Based on the classified indexes, we divide the histogram count matrix to estimate the histogram of the original image. Thus, we extend the spatial-based Retinex model into the Histogram-based Retinex model:
\begin{equation}
\label{deqn_ex1a}
{{H}_{C}}{{(\boldsymbol{S})}_{i}}=\sum\limits_{j\in \partial {{N}_{i}}}{{{\boldsymbol{H}}_{C}}{{(\boldsymbol{L},\boldsymbol{R})}_{{{x}_{j}},{{y}_{j}}}}/N},
\end{equation}
where the $\partial {{N}_{i}}$ denotes the elements set in the histogram location matrix ${{\boldsymbol{H}}_{B}}(\boldsymbol{L},\boldsymbol{R})$, which have similar values to the $i$-th gray level in the ${{H}_{B}}(\boldsymbol{S})$. The ${{x}_{j}}$ and ${{y}_{j}}$ denote the vertical and horizontal position of $j$-th element in the ${{\boldsymbol{H}}_{C}}(\boldsymbol{L},\boldsymbol{R})$.

Our method replaces the spatial-domain matrix multiplication with three key histogram-domain operations: (1) vector product, (2) value rounding, and (3) intensity-level classification. This paradigm shift from per-pixel to grayscale-level processing decouples computational complexity from image resolution and achieves significant efficiency. As illustrated in Fig. 3, the proposed model successfully estimates the histogram of the original image from the illumination and reflectance component histograms and demonstrates the accuracy and efficiency of proposed method.
\subsection{Optimization Model}
Based on the Histogram-based Retinex model, we construct a novel optimization model, which is as follows:
\begin{equation}
\label{deqn_ex1a}
\begin{aligned}
  & \min {{J}_{\text{HistRetinex}}}({{H}_{C}}(\boldsymbol{R}),{{H}_{C}}(\boldsymbol{L})) \\ 
 & =\sum\limits_{i=0}^{l-1}{{{\left( {{H}_{C}}{{(\boldsymbol{S})}_{i}}-\sum\limits_{j\in \partial {{N}_{i}}}{{{H}_{C}}{{(\boldsymbol{L})}_{{{y}_{j}}}}{{H}_{C}}{{(\boldsymbol{R})}_{{{x}_{j}}}}/N} \right)}^{2}}} \\ 
 & +\alpha \sum\limits_{i=0}^{l-1}{{{\left( {{H}_{C}}{{(\boldsymbol{L})}_{i}}-{{H}_{C}}{{(\boldsymbol{S})}_{i}} \right)}^{2}}}\\
 &+\beta \sum\limits_{i=0}^{l-1}{{{\left( {{H}_{C}}{{(\boldsymbol{R})}_{i}}-{{H}_{C}}{{(\nabla \boldsymbol{S})}_{i}} \right)}^{2}}}, \\ 
\end{aligned}
\end{equation}
where the $\alpha $ and $\beta $ are the weight parameter of the prior terms about the illumination and reflectance components.

This objective function introduces the histogram-based Retinex model, which estimates the gray level of the histogram ${{H}_{C}}(\boldsymbol{S})$ one by one. However, the iterative formulas ${{H}_{C}}(\boldsymbol{R})$ and ${{H}_{C}}(\boldsymbol{L})$ cannot be solved by an optimization model using this estimating approach. Therefore, we need to rewrite the objective function and process the elements of the histogram count matrix ${{\boldsymbol{H}}_{C}}(\boldsymbol{L},\boldsymbol{R})$ one by one. The rewired objective function is as follows:
\begin{equation}
\label{deqn_ex1a}
\begin{aligned}
  & {{J}_{\text{HistRetinex}}}({{H}_{C}}(\boldsymbol{R}),{{H}_{C}}(\boldsymbol{L})) \\ 
 & ={{\sum\limits_{i=1}^{l-1}{\sum\limits_{j=1}^{l-1}{\left( \frac{{{H}_{C}}{{(\boldsymbol{R})}_{i}}\times {{H}_{C}}{{(\boldsymbol{L})}_{j}}}{N}-{{K}_{ij}}\cdot {{H}_{C}}{{(\boldsymbol{S})}_{index(i,j)}} \right)}}}^{2}} \\ 
 & +\alpha \sum\limits_{j=1}^{l-1}{{{\left( {{H}_{C}}{{(\boldsymbol{L})}_{j}}-{{H}_{C}}{{(\boldsymbol{S})}_{j}} \right)}^{2}}}\\
 &+\beta \sum\limits_{i=1}^{l-1}{{{\left( {{H}_{C}}{{(\boldsymbol{R})}_{i}}-{{H}_{C}}{{(\nabla \boldsymbol{S})}_{i}} \right)}^{2}}}, \\ 
\end{aligned}
\end{equation}
where the $index(i,j)$ denotes the function that $index(i,j)=\underset{k}{\mathop{\min }}\,(|{{\boldsymbol{H}}_{B}}{{(\boldsymbol{L},\boldsymbol{R})}_{i,j}}-{{H}_{B}}{{(\boldsymbol{S})}_{k}}|)$, the ${{K}_{ij}}$ is the weight of ${{H}_{C}}{{(\boldsymbol{S})}_{index(i,j)}}$ and calculated by 
\begin{equation}
\label{deqn_ex1a}
\begin{aligned}
 	{{K}_{ij}}={{\boldsymbol{H}}_{C}}{{(\boldsymbol{L},\boldsymbol{R})}_{i,j}}/\sum\limits_{k\in \partial {{N}_{index(i,j)}}}{{{\boldsymbol{H}}_{C}}{{(\boldsymbol{L},\boldsymbol{R})}_{{{x}_{k}},{{y}_{k}}}}},
\end{aligned}
\end{equation}

For Eq. (10), we calculate the partial derivatives of ${{J}_{HRM}}({{H}_{C}}(\boldsymbol{R}),{{H}_{C}}(\boldsymbol{L}))$ with respect to $H{\boldsymbol{(R)}_{i}}$ and $H{\boldsymbol{(L)}_{j}}$, respectively. The iterative formulas of the histogram about the illumination and reflectance are given as follows:

\begin{equation}
\label{deqn_ex1a}
\begin{aligned}
  & {{H}_{C}}{{(\boldsymbol{L})}_{j}} \\ 
 & =\frac{\sum\limits_{i=0}^{l-1}{{{H}_{C}}{{(\boldsymbol{R})}_{i}}{{H}_{C}}{{(\boldsymbol{S})}_{index(i,j)}}}/(N{{K}_{ij}})+\alpha {{H}_{C}}{{(\boldsymbol{S})}_{j}}}{\sum\limits_{i=0}^{l-1}{H_{C}^{2}{{(\boldsymbol{R})}_{i}}}/{{N}^{2}}+\alpha }, \\ 
\end{aligned}
\end{equation}

\begin{equation}
\label{deqn_ex1a}
\begin{aligned}
  & {{H}_{C}}{{(\boldsymbol{R})}_{i}} \\ 
 & =\frac{\sum\limits_{j=0}^{l-1}{{{H}_{C}}{{(\boldsymbol{L})}_{j}}{{H}_{C}}{{(\boldsymbol{S})}_{index(i,j)}}}/(N{{K}_{ij}})+\beta {{H}_{C}}{{(\nabla \boldsymbol{S})}_{i}}}{\sum\limits_{j=0}^{l-1}{H_{C}^{2}{{(\boldsymbol{L})}_{j}}}/{{N}^{2}}+\beta }, \\ 
\end{aligned}
\end{equation}

Due to the limitation of (1) $\sum\limits_{i=1}^{L}{{{H}_{C}}{{(\boldsymbol{R})}_{i}}}=N$ and (2) $\sum\limits_{j=1}^{L}{{{H}_{C}}{{(\boldsymbol{L})}_{j}}}=N$, the final histograms of the illumination and reflectance are obtained as follows:
\begin{equation}
\label{deqn_ex1a}
\begin{aligned}
{{\tilde{H}}_{C}}{{(\boldsymbol{L})}_{j}}=\frac{N\cdot {{H}_{C}}{{(\boldsymbol{L})}_{j}}}{\sum\limits_{k=0}^{l-1}{{{H}_{C}}{{(\boldsymbol{L})}_{k}}}},
\end{aligned}
\end{equation}
\begin{equation}
\label{deqn_ex1a}
\begin{aligned}
{{\tilde{H}}_{C}}{{(\boldsymbol{R})}_{i}}=\frac{N\cdot {{H}_{C}}{{(\boldsymbol{R})}_{i}}}{\sum\limits_{k=0}^{l-1}{{{H}_{C}}{{(\boldsymbol{R})}_{k}}}},
\end{aligned}
\end{equation}
The pseudocode of the proposed algorithm is as follows in Algorithm 1.

\begin{algorithm}[H]
\caption{HistRetinex: Histogram-based Retinex model}\label{alg:alg1}
\begin{algorithmic}
\STATE \textbf{Input:} low-light image $\boldsymbol{S}$, error threshold $\varepsilon $, maximum iteration $T$, weight parameter $\alpha $ and $\beta $.
\STATE \textbf{Output:} Histogram of illumination component ${{H}_{C}}(\boldsymbol{L})$ and histogram of reflectance component ${{H}_{C}}(\boldsymbol{R})$.
\STATE \vspace{-0.5em}\makebox[0pt][l]{\rule{\linewidth}{0.5pt}}
\STATE \textbf{Initialization:} 
\STATE \vspace{-0.5em}\makebox[0pt][l]{\rule{\linewidth}{0.5pt}}
\STATE \hspace{0.5cm} $t=0$
\STATE \hspace{0.5cm} Compute the histogram of the low-light image ${{H}_{C}}(\boldsymbol{S})$
\STATE \hspace{0.5cm} Compute the histogram of the gradient image ${{H}_{C}}(\nabla \boldsymbol{S})$
\STATE \hspace{0.5cm} Initialize the ${{H}_{C}}{{(\boldsymbol{R})}^{(0)}}={{[{{H}_{C}}{{(\nabla \boldsymbol{S})}_{i}}]}_{\boldsymbol{L}\times 1}}$;
\STATE \textbf{repeat:}
\STATE \hspace{0.5cm} Update ${{H}_{C}}{(\boldsymbol{R})}^{(t+1)}$ by Eq. (13);
\STATE \hspace{0.5cm} Update ${{H}_{C}}{(\boldsymbol{L})}^{(t+1)}$ by Eq. (12);
\STATE \hspace{0.5cm} Update ${{\tilde{H}}_{C}}{{(\boldsymbol{R})}^{(t+1)}}$ by Eq. (15);;
\STATE \hspace{0.5cm} Update ${{\tilde{H}}_{C}}{{(\boldsymbol{L})}^{(t+1)}}$ by Eq. (14);
\STATE \hspace{0.5cm} Update ${{K}^{(t+1)}}$ by Eq. (11);
\STATE \hspace{0.5cm}  $t=t+1$;
\STATE \textbf{until}  $||{{H}_{C}}(\boldsymbol{R})_{j}^{(t+1)}-{{H}_{C}}(\boldsymbol{R})_{j}^{(t)}||_{F}^{2}\le \varepsilon $ and $t>T$ or $||{{H}_{C}}(\boldsymbol{L})_{i}^{(t+1)}-{{H}_{C}}(\boldsymbol{L})_{i}^{(t)}||_{F}^{2}\le \varepsilon $
\STATE \textbf{Return} ${{H}_{C}}(\boldsymbol{R})$ and ${{H}_{C}}(\boldsymbol{L})$
\end{algorithmic}
\label{alg1}
\end{algorithm}

\subsection{ Histogram Reprocessing}

After obtaining the final histograms of the illumination and reflectance components, we need to modify the histogram of illumination to improve the visual range and lightness of the low-light image. The traditional Retinex-based image enhancement algorithm improves the illumination and is shown in as follows:

\begin{equation}
\label{deqn_ex1a}
\begin{aligned}
\boldsymbol{\overline{L}}={{\boldsymbol{(L)}}^{\frac{1}{\gamma }}},
\end{aligned}
\end{equation}
where the $\gamma $ is the user-defined parameter, often set to 2.2.

Similar to the traditional Retinex-based enhancement methods, we also improve the contrast of the illumination component in the histogram domain. The enhanced method is given by:
\begin{equation}
\label{deqn_ex1a}
\begin{aligned}
\overline{{{H}_{B}}}(\boldsymbol{L})={{({{H}_{B}}(\boldsymbol{L}))}^{\frac{1}{\gamma }}},
\end{aligned}
\end{equation}

We process the location histogram instead of the count histogram, and construct a new location matrix of the histogram.

\begin{equation}
\label{deqn_ex1a}
\begin{aligned}
\overline{{{H}_{B}}}(\boldsymbol{L},\boldsymbol{R})=\overline{{{H}_{B}}}(\boldsymbol{L})\times {{H}_{B}}{{(\boldsymbol{R})}^{\text{T}}},
\end{aligned}
\end{equation}

Based on the new histogram location matrix, we construct the relationship between the histogram location matrix and the histogram of the low-light image.
\begin{equation}
\label{deqn_ex1a}
\begin{aligned}
inde{{x}_{1}}(i,j)=\underset{k}{\mathop{\min }}\,(|\overline{{{\boldsymbol{H}}_{B}}}{{(\boldsymbol{L},\boldsymbol{R})}_{i,j}}-{{H}_{B}}{{(\boldsymbol{S})}_{k}}|),
\end{aligned}
\end{equation}

Finally, we obtain the histogram of the enhanced results as follows:
\begin{equation}
\label{deqn_ex1a}
\begin{aligned}
	{{H}_{C}}{{(\widetilde{\boldsymbol{S}})}_{k}}=\sum\limits_{\begin{smallmatrix} 
 inde{{x}_{1}}(i,j)=k \\ 
 i,j\in L
\end{smallmatrix}}{{{H}_{C}}{{(\boldsymbol{R},\boldsymbol{L})}_{i,j}}},
\end{aligned}
\end{equation}
where the $\widetilde{\boldsymbol{S}}$ denotes the enhanced results. Due to the lack of the spatial information of the histogram ${{H}_{C}}{{(\boldsymbol{S})}_{k}}$, it is not possible to directly recover the enhanced image. To this end, we utilize the histogram matching to estimate the enhanced image.

\section{Experimental Results And Analysis}
To verify the performance of proposed algorithm, the LIME [15], STAR [28], SSD [25], WVM [36], AHPCE [35], RRM [23], LR3M [24], BRAIN [29] and the FSTAR [30] are compared with the proposed algorithms The experiment platform is an HP laptop equipped with an Intel Core i5-7300HQ processor, 8 GB RAM, and the MATLAB 2017b software environment. The test images come from the MEF [36], NPE [37], LOL [38] and the LSRW [39] datasets. The parameters of the proposed algorithm are set to  ,  , and  . In addition, we use four indexes to evaluate the enhancement effect of algorithm, including the peak signal to noise ratio (PSNR) [40], structural similarity (SSIM) [41], natural image quality evaluator (NIQE) [42] and the lightness order error (LOE) [15].

\subsection{Test and Analysis of Algorithm Effectiveness}

In this section, to verify the effectiveness of these algorithms in the low-light images, we select six images for testing. The test image \#Candle, \#BelgiumHouse, \#House from the MEF dataset, the \#Daybreak\&Nightfall (11), \#night (52) and \#Parking from the NPE dataset. The results of these algorithms are given in Fig.~4, and corresponding performance metrics are shown in Fig.~5.

From Fig.~4, the LIME and AHPCE have the worst enhancement effect. The APHCE obtains results with severe color distortion. Meanwhile, the LIME provides results with an inappropriate range of illumination and noise points in the edges. The STAR gets better enhancement results with more appropriate image illumination, but the results still have many noise points. Based on this, the FSTAR provides a better enhancement effect, which uses the MED to capture significant structural variation. The SSD boosts the shadow of the image beyond the appropriate range, so there will be some color distortion in the enhancement result. The RRM meets the color distortion problem in the background, such as the cloud and mountain in the image \#Parking. This is due to the fact that the RRM and LR3M processes three channels low low-light images in the HSV color space concurrently, and will stretch the illumination of three channels during the illumination stretching processing. The BRAIN and proposed algorithm provide the best enhancement results, which have a suitable illumination range and fewer halo artifacts. In Fig.~5, the proposed algorithm is superior to all compared algorithms in terms of the PSNR and the SSIM.
\begin{figure*}[!t]
\centering
\subfloat{\includegraphics[width=3.5in]{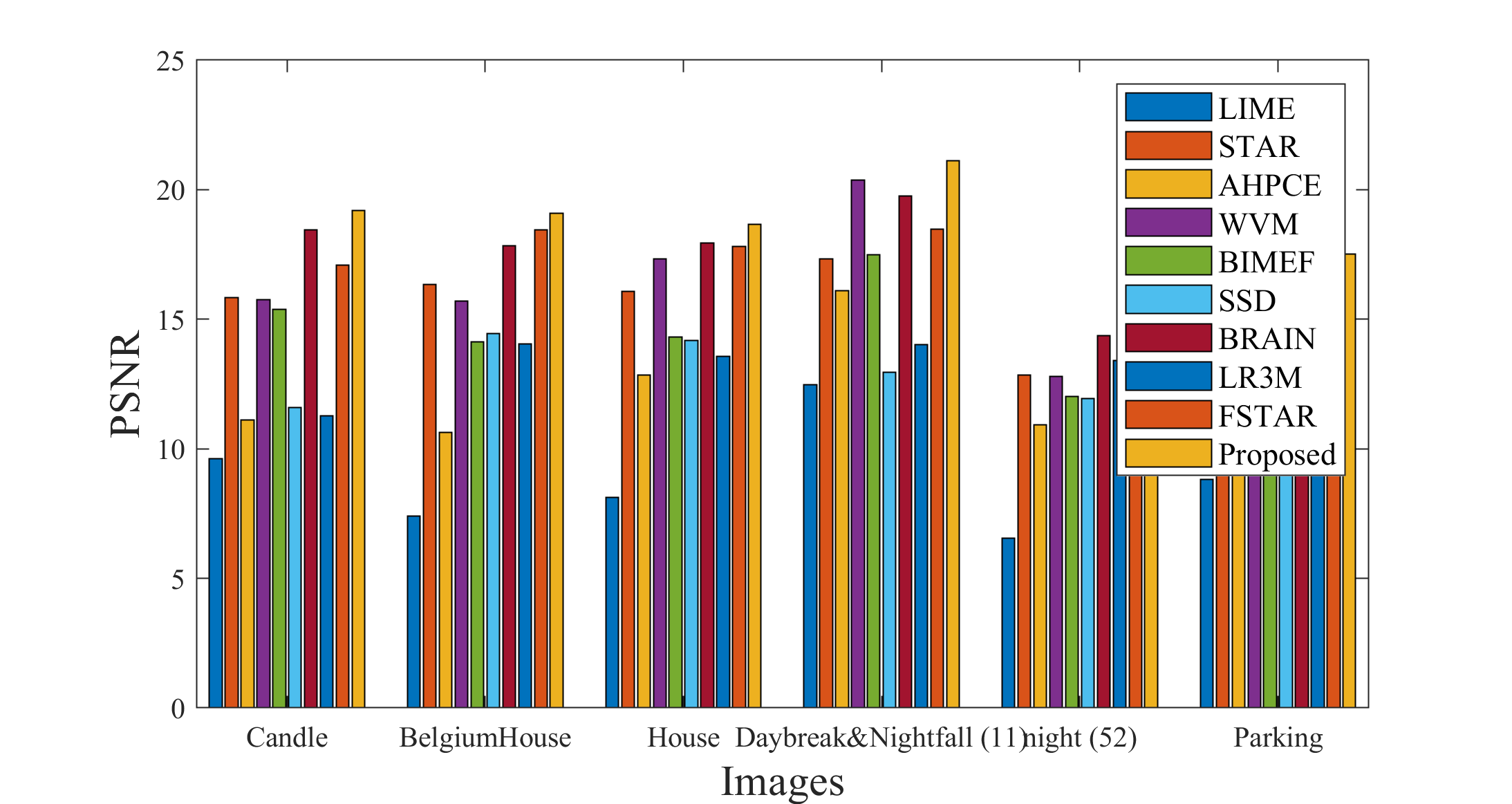}%
\label{fig_first_case}}
\hfil
\subfloat{\includegraphics[width=3.5in]{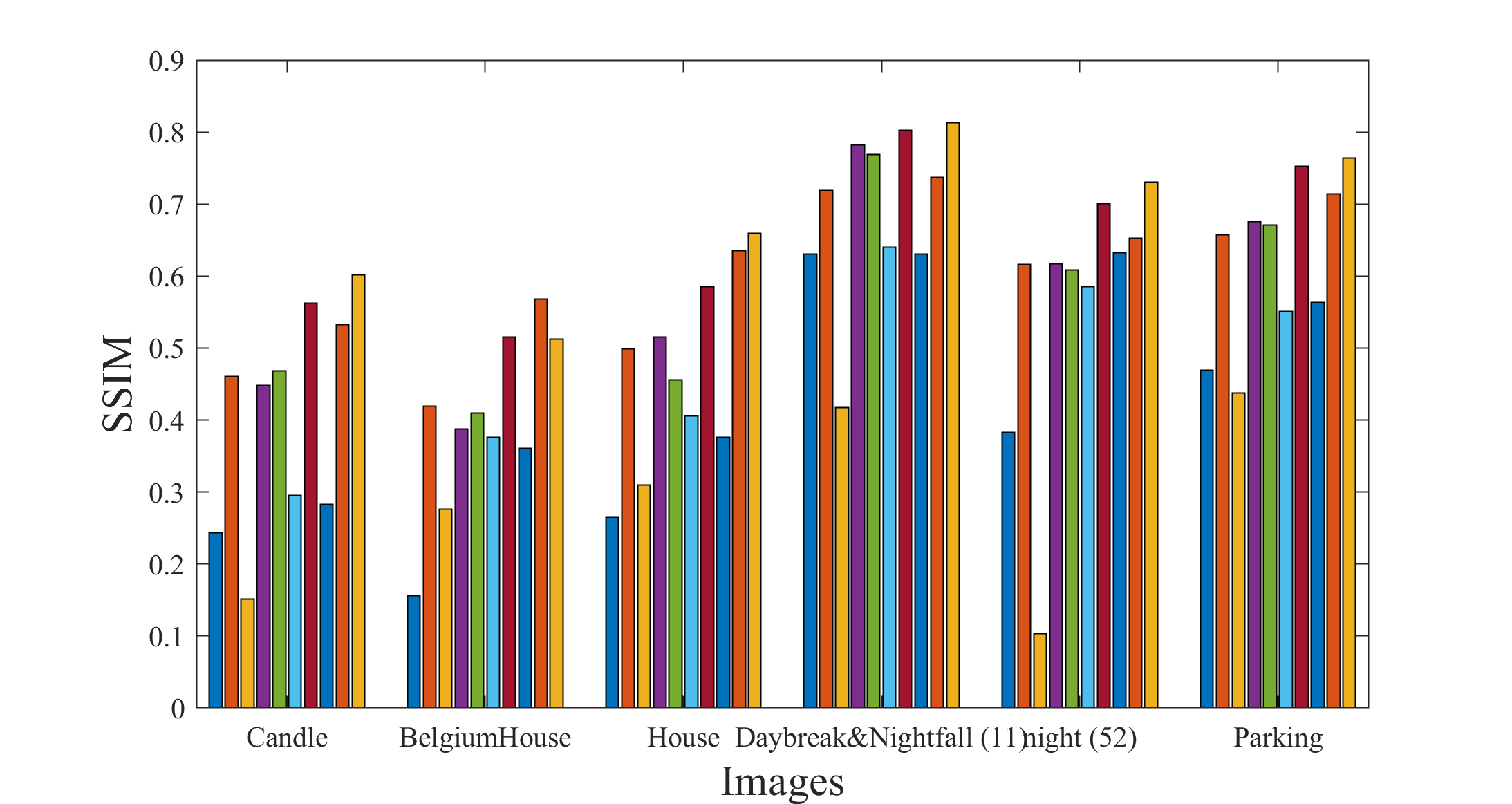}%
\label{fig_second_case}}

(a)\hspace{8.6cm}(b)
\caption{The performance metrics of these algorithms in low-light images. (a) PSNR; (b) SSIM.}
\label{fig_sim}
\end{figure*}

\begin{figure}[htbp]
    \centering
    \subfloat{\includegraphics[width=1.45cm, height=0.97cm]{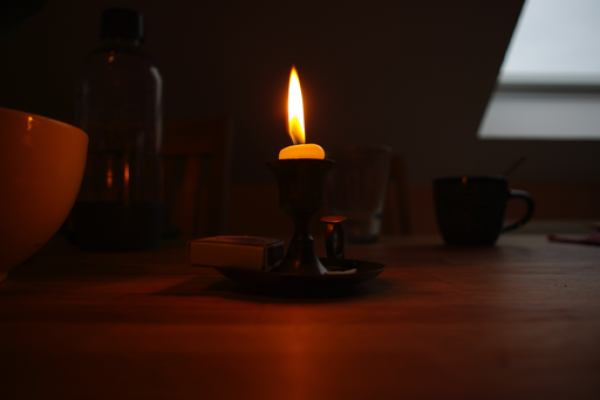}}%
    \subfloat{\includegraphics[width=1.45cm, height=0.97cm]{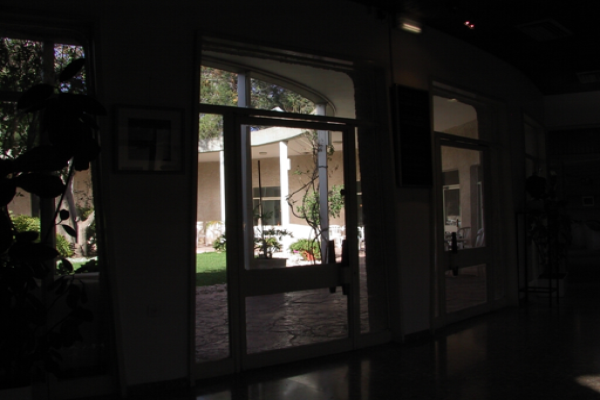}}%
    \subfloat{\includegraphics[width=1.45cm, height=0.97cm]{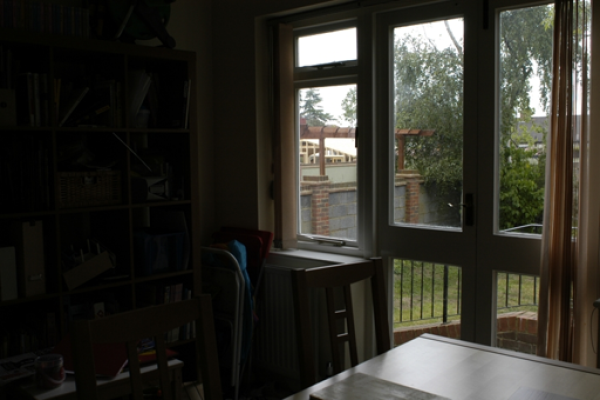}}%
    \subfloat{\includegraphics[width=1.45cm, height=0.97cm]{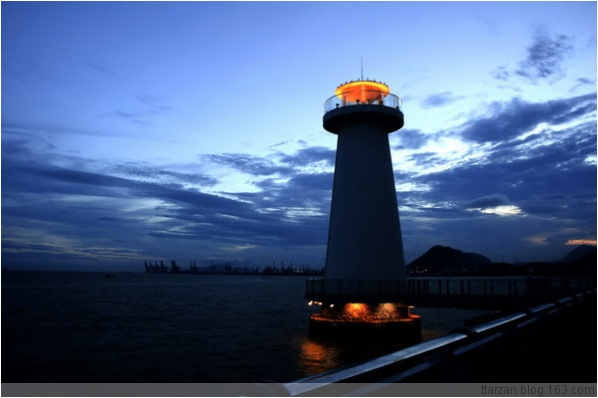}}%
    \subfloat{\includegraphics[width=1.45cm, height=0.97cm]{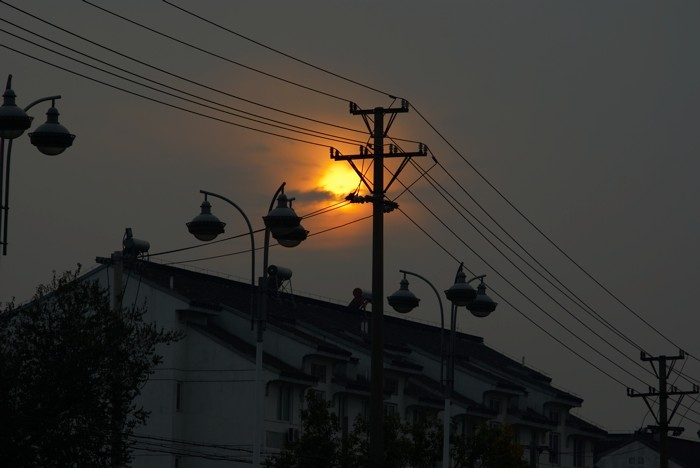}}%
    \subfloat{\includegraphics[width=1.45cm, height=0.97cm]{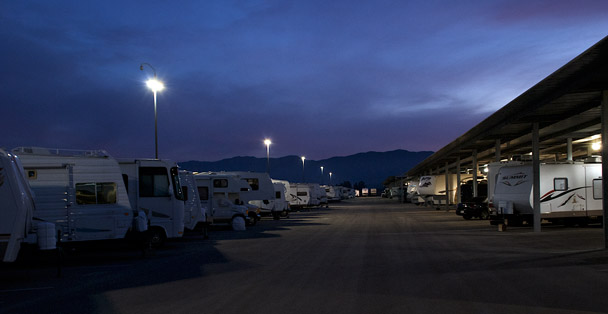}}\\
    \vspace{-1em}
    \subfloat{\includegraphics[width=1.45cm, height=0.97cm]{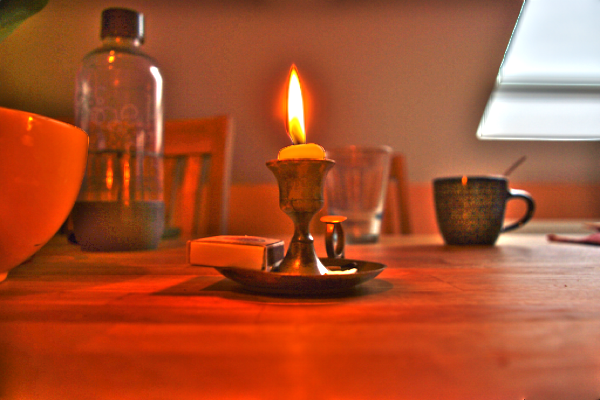}}%
    \subfloat{\includegraphics[width=1.45cm, height=0.97cm]{effectiveness/BelgiumHouseoriginal.png}}%
    \subfloat{\includegraphics[width=1.45cm, height=0.97cm]{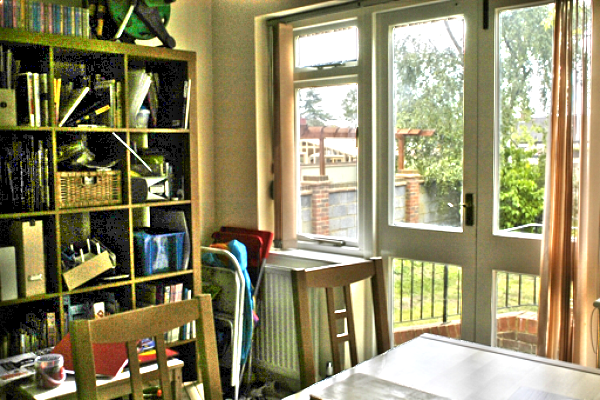}}%
    \subfloat{\includegraphics[width=1.45cm, height=0.97cm]{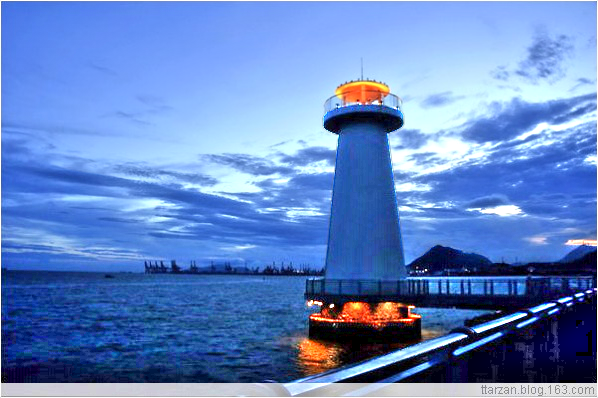}}%
    \subfloat{\includegraphics[width=1.45cm, height=0.97cm]{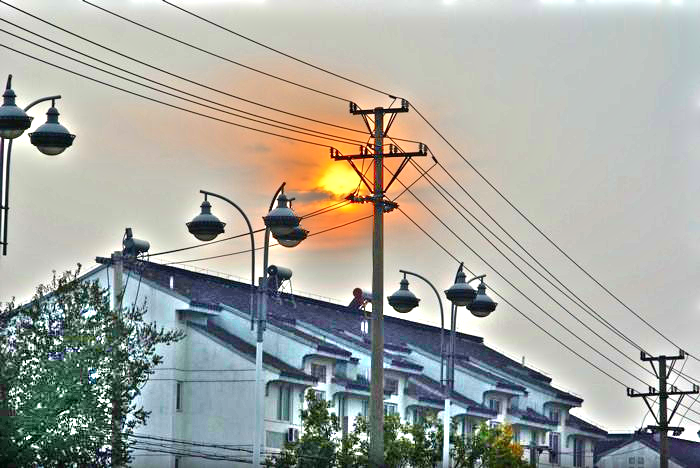}}%
    \subfloat{\includegraphics[width=1.45cm, height=0.97cm]{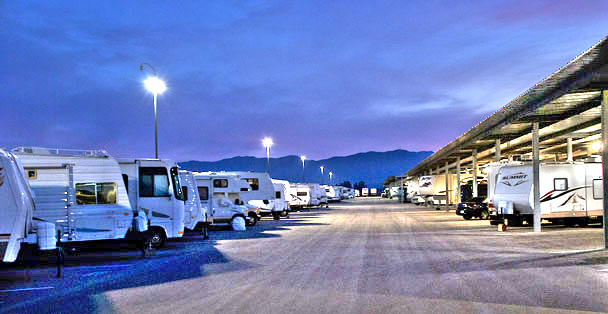}}\\
    \vspace{-1em}
    \subfloat{\includegraphics[width=1.45cm, height=0.97cm]{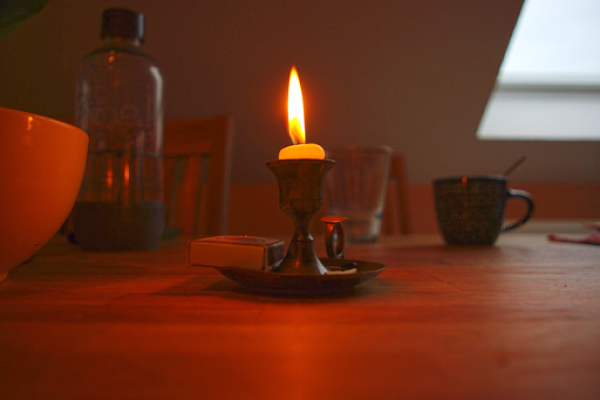}}%
    \subfloat{\includegraphics[width=1.45cm, height=0.97cm]{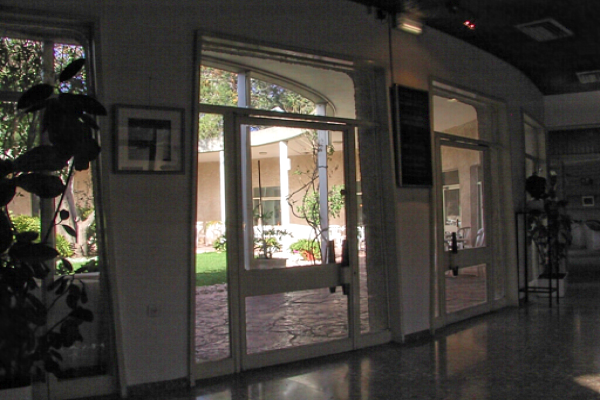}}%
    \subfloat{\includegraphics[width=1.45cm, height=0.97cm]{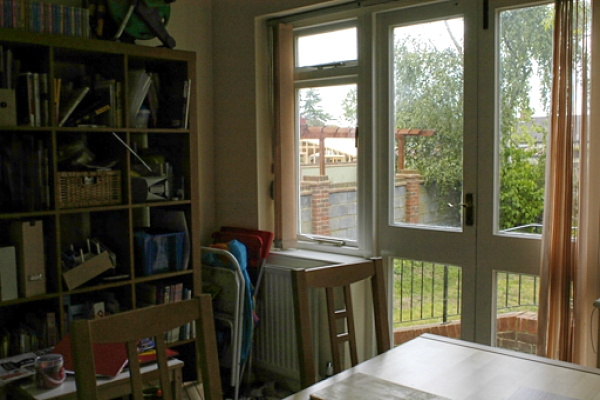}}%
    \subfloat{\includegraphics[width=1.45cm, height=0.97cm]{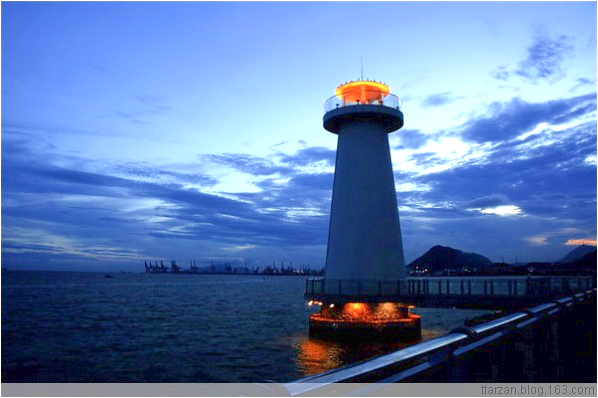}}%
    \subfloat{\includegraphics[width=1.45cm, height=0.97cm]{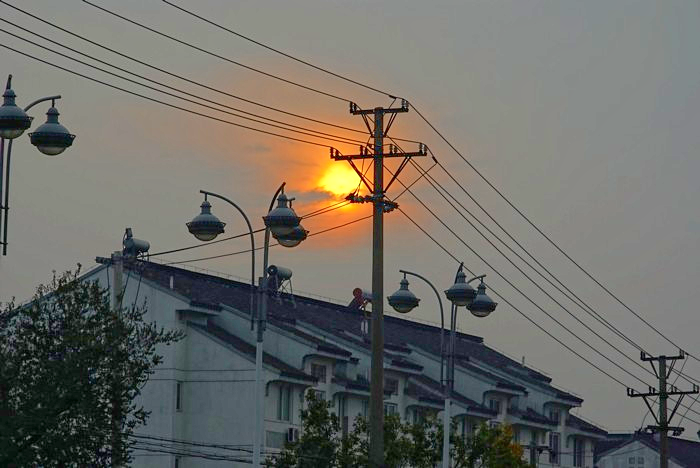}}%
    \subfloat{\includegraphics[width=1.45cm, height=0.97cm]{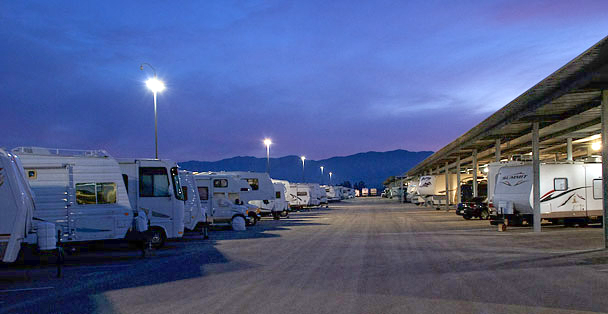}}\\
    \vspace{-1em}
    \subfloat{\includegraphics[width=1.45cm, height=0.97cm]{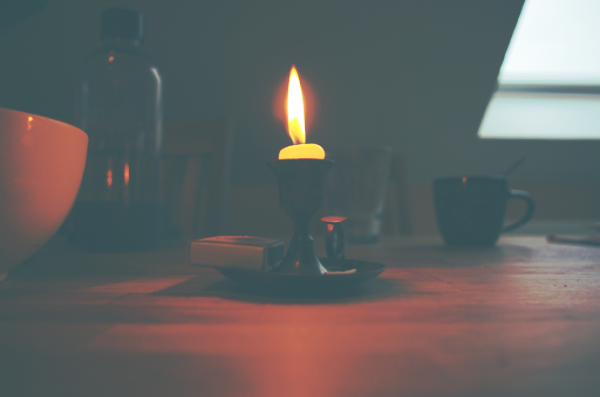}}%
    \subfloat{\includegraphics[width=1.45cm, height=0.97cm]{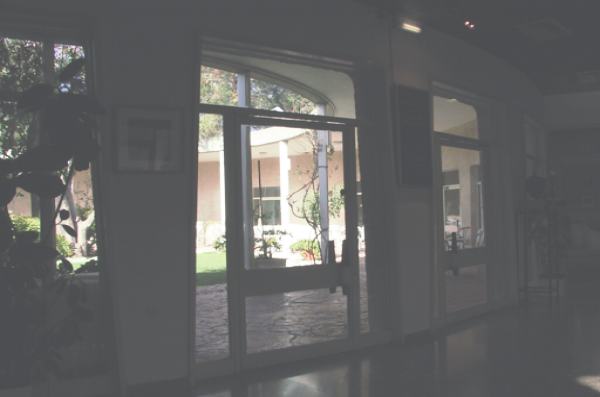}}%
    \subfloat{\includegraphics[width=1.45cm, height=0.97cm]{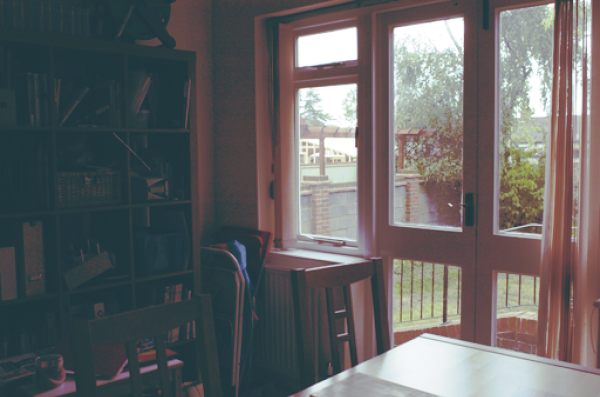}}%
    \subfloat{\includegraphics[width=1.45cm, height=0.97cm]{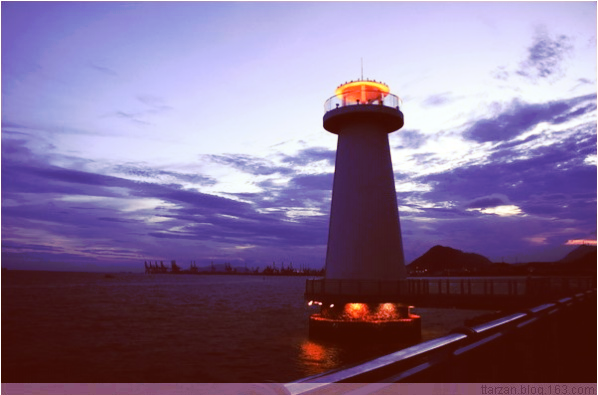}}%
    \subfloat{\includegraphics[width=1.45cm, height=0.97cm]{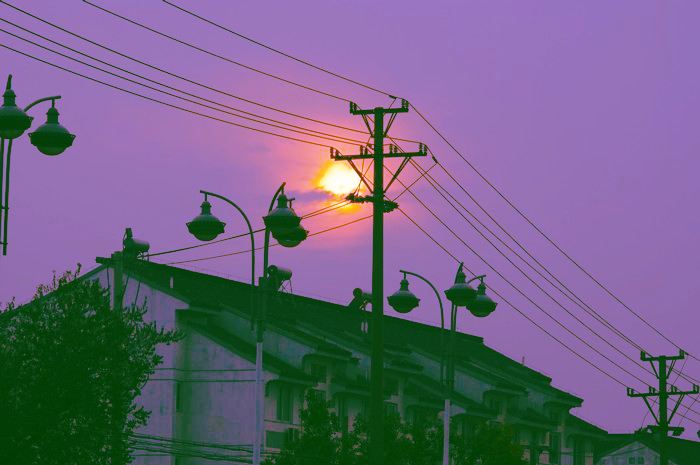}}%
    \subfloat{\includegraphics[width=1.45cm, height=0.97cm]{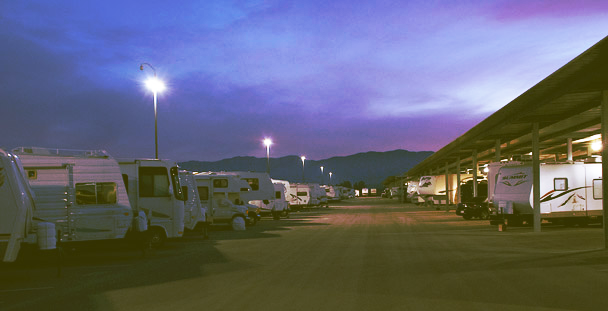}}\\
    \vspace{-1em}
    \subfloat{\includegraphics[width=1.45cm, height=0.97cm]{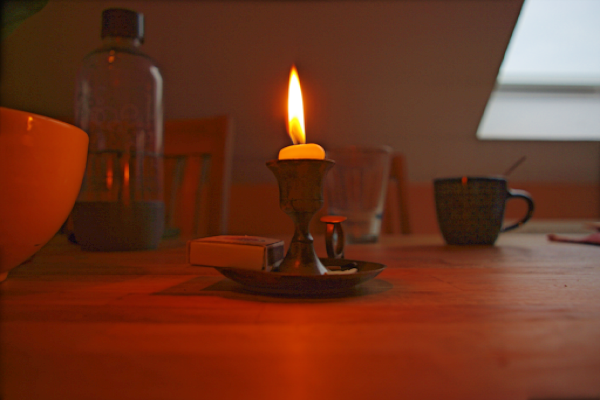}}%
    \subfloat{\includegraphics[width=1.45cm, height=0.97cm]{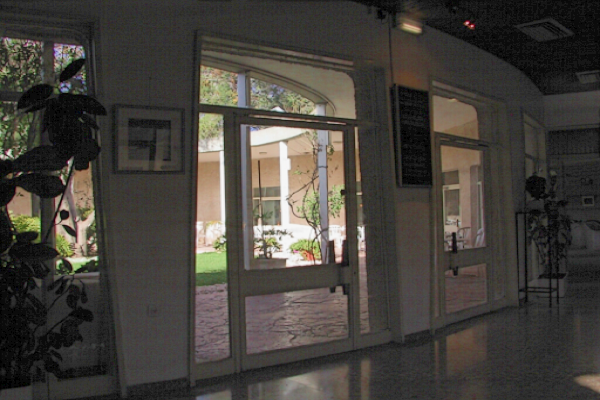}}%
    \subfloat{\includegraphics[width=1.45cm, height=0.97cm]{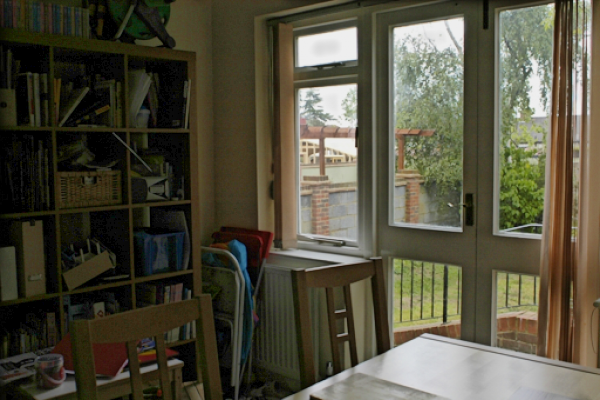}}%
    \subfloat{\includegraphics[width=1.45cm, height=0.97cm]{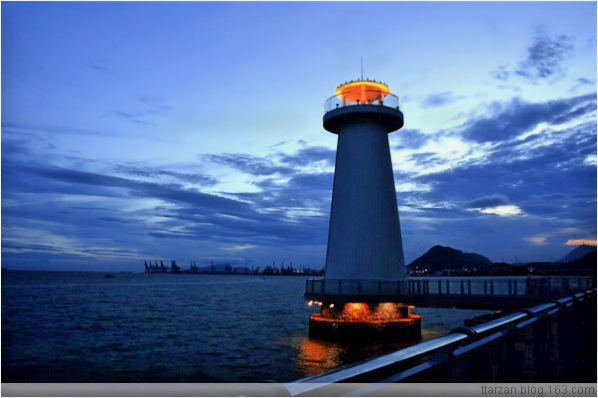}}%
    \subfloat{\includegraphics[width=1.45cm, height=0.97cm]{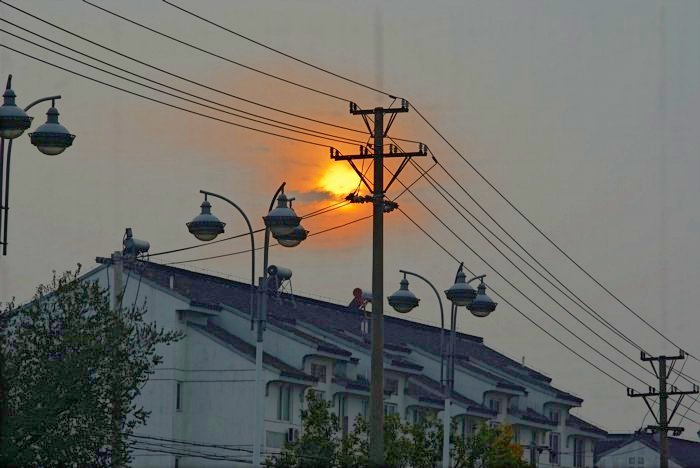}}%
    \subfloat{\includegraphics[width=1.45cm, height=0.97cm]{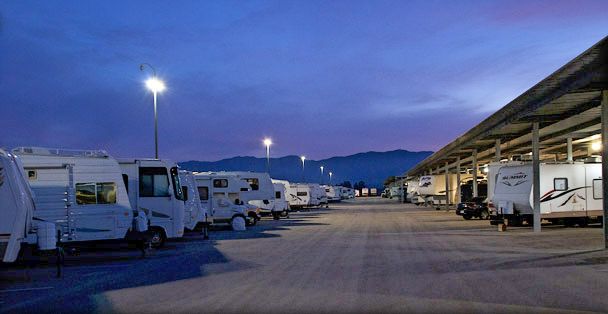}}\\
    \vspace{-1em}
    \subfloat{\includegraphics[width=1.45cm, height=0.97cm]{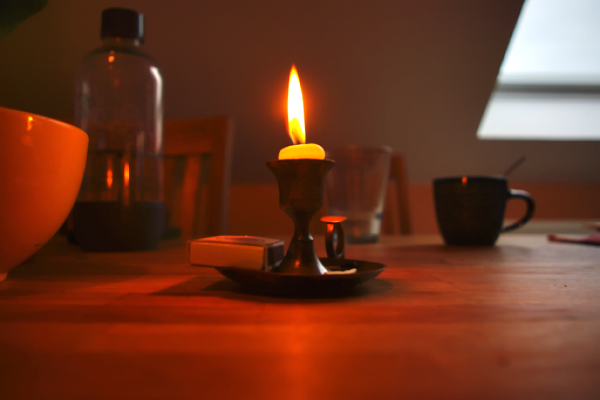}}%
    \subfloat{\includegraphics[width=1.45cm, height=0.97cm]{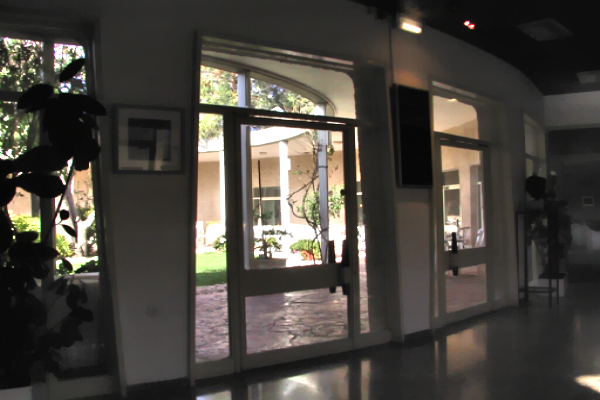}}%
    \subfloat{\includegraphics[width=1.45cm, height=0.97cm]{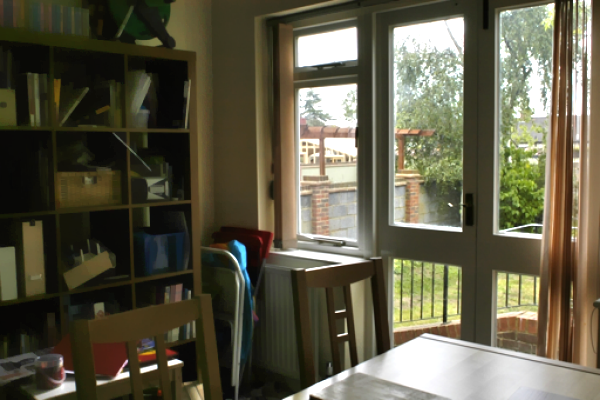}}%
    \subfloat{\includegraphics[width=1.45cm, height=0.97cm]{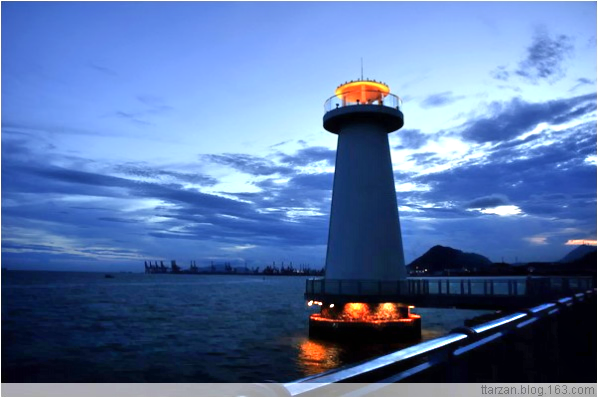}}%
    \subfloat{\includegraphics[width=1.45cm, height=0.97cm]{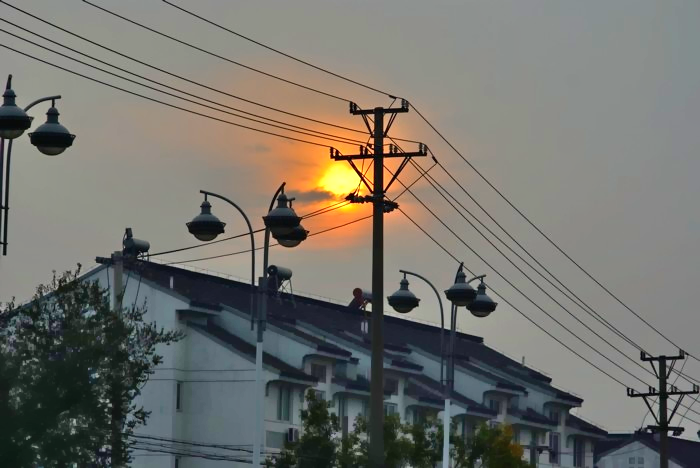}}%
    \subfloat{\includegraphics[width=1.45cm, height=0.97cm]{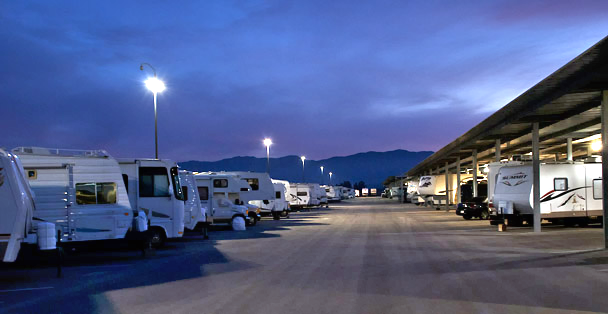}}\\
    \vspace{-1em}
    \subfloat{\includegraphics[width=1.45cm, height=0.97cm]{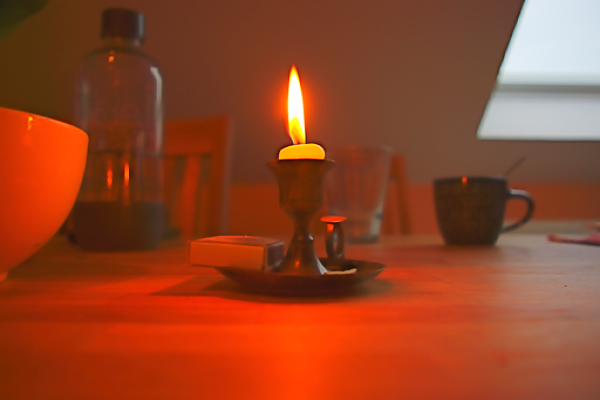}}%
    \subfloat{\includegraphics[width=1.45cm, height=0.97cm]{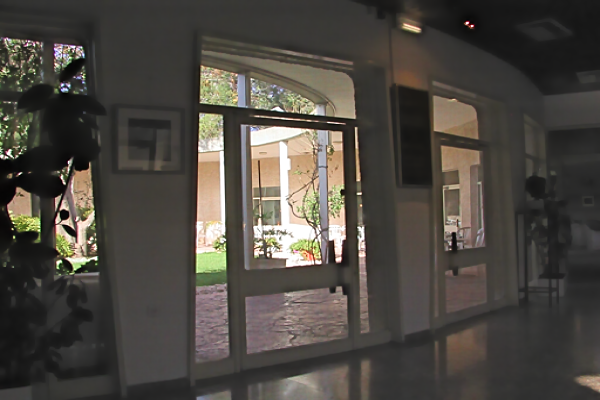}}%
    \subfloat{\includegraphics[width=1.45cm, height=0.97cm]{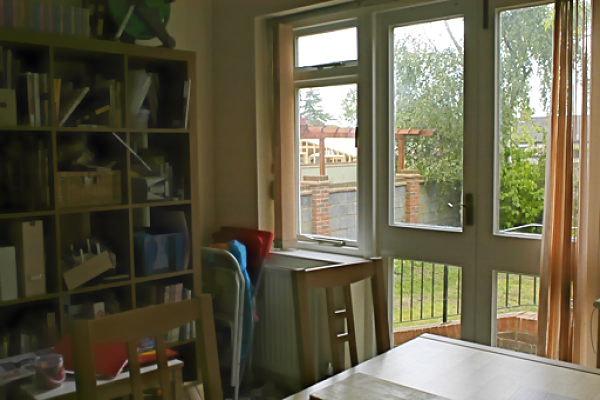}}%
    \subfloat{\includegraphics[width=1.45cm, height=0.97cm]{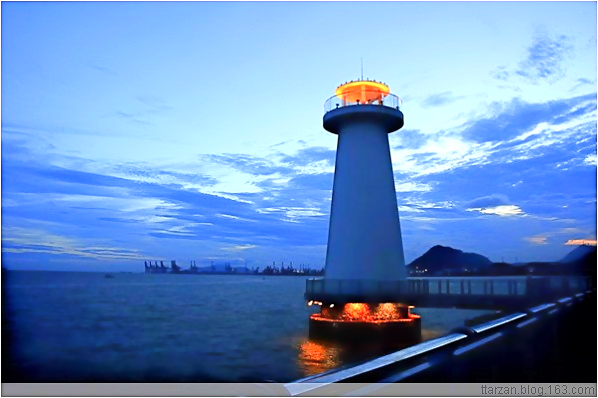}}%
    \subfloat{\includegraphics[width=1.45cm, height=0.97cm]{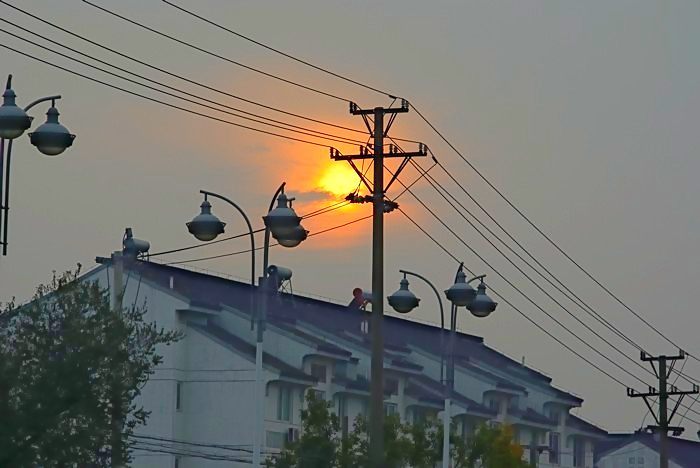}}%
    \subfloat{\includegraphics[width=1.45cm, height=0.97cm]{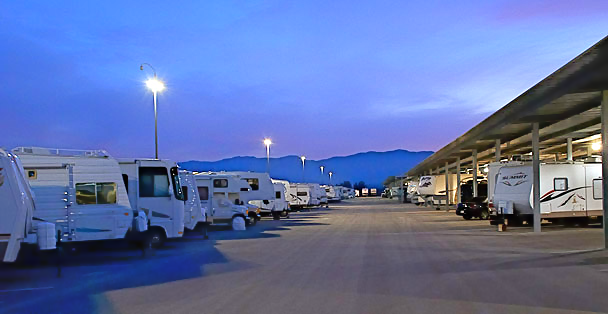}}\\
    \vspace{-1em}
    \subfloat{\includegraphics[width=1.45cm, height=0.97cm]{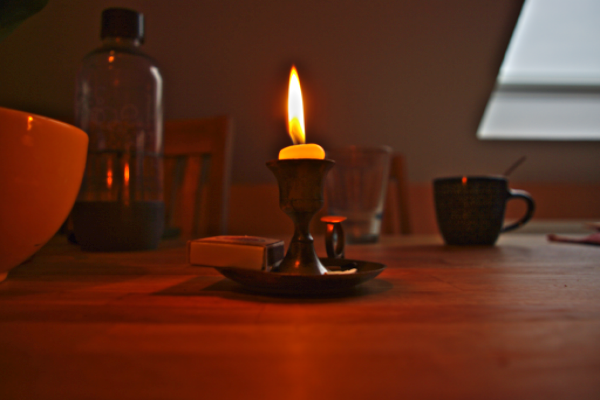}}%
    \subfloat{\includegraphics[width=1.45cm, height=0.97cm]{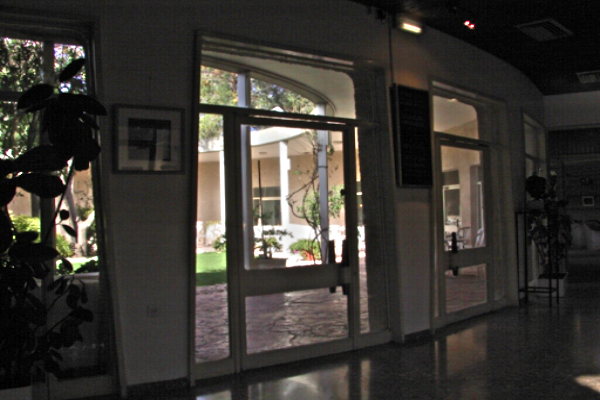}}%
    \subfloat{\includegraphics[width=1.45cm, height=0.97cm]{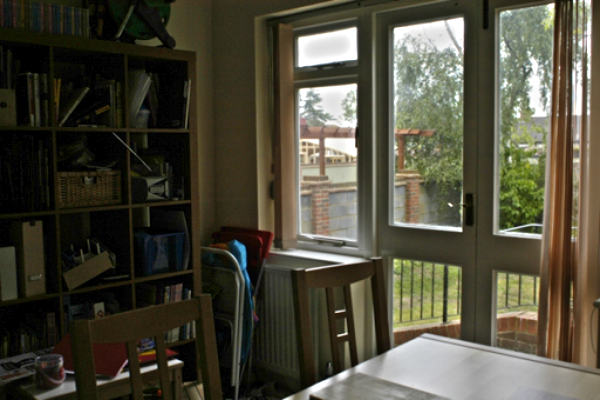}}%
    \subfloat{\includegraphics[width=1.45cm, height=0.97cm]{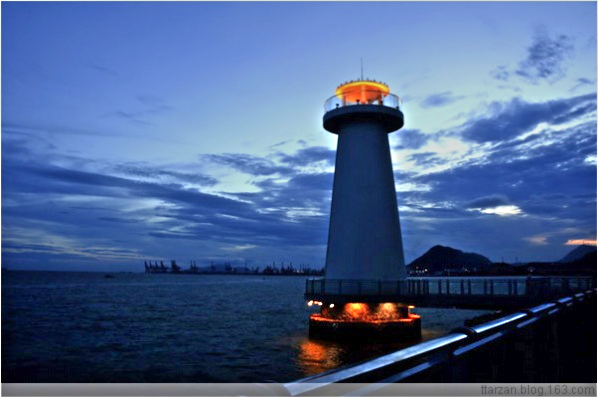}}%
    \subfloat{\includegraphics[width=1.45cm, height=0.97cm]{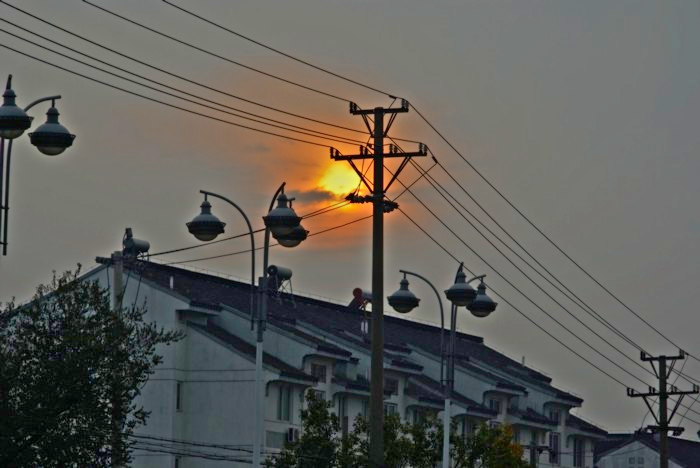}}%
    \subfloat{\includegraphics[width=1.45cm, height=0.97cm]{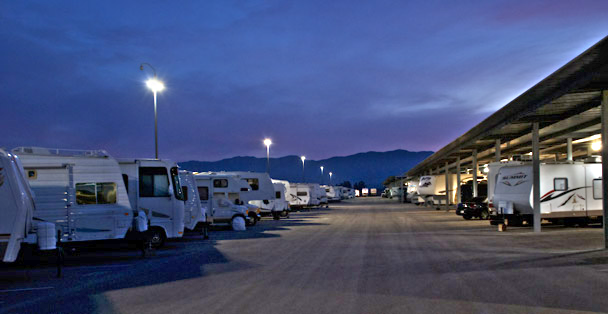}}\\
    \vspace{-1em}
    \subfloat{\includegraphics[width=1.45cm, height=0.97cm]{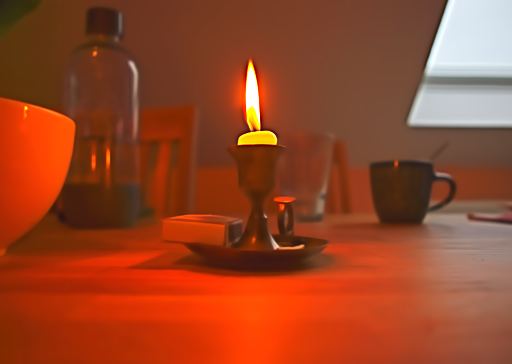}}%
    \subfloat{\includegraphics[width=1.45cm, height=0.97cm]{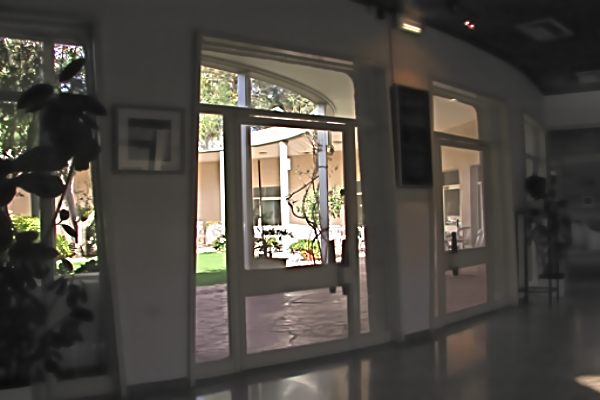}}%
    \subfloat{\includegraphics[width=1.45cm, height=0.97cm]{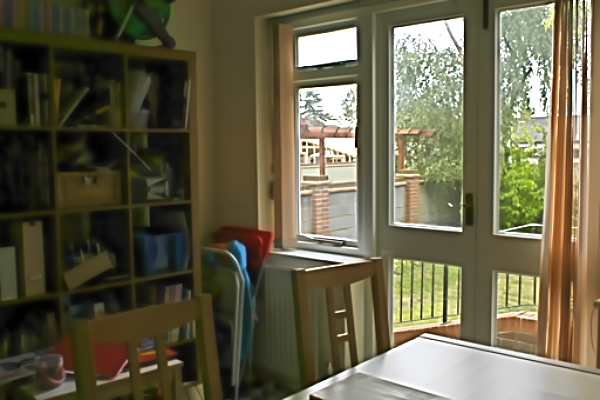}}%
    \subfloat{\includegraphics[width=1.45cm, height=0.97cm]{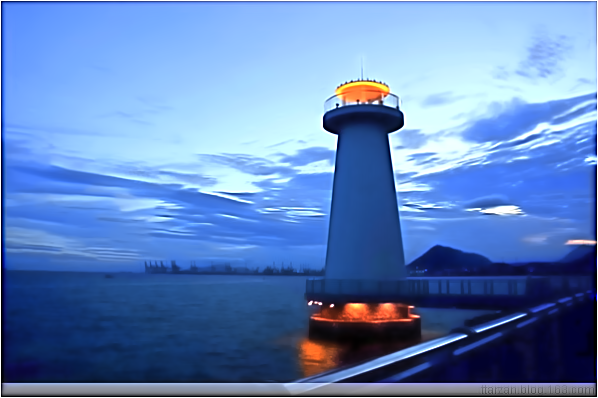}}%
    \subfloat{\includegraphics[width=1.45cm, height=0.97cm]{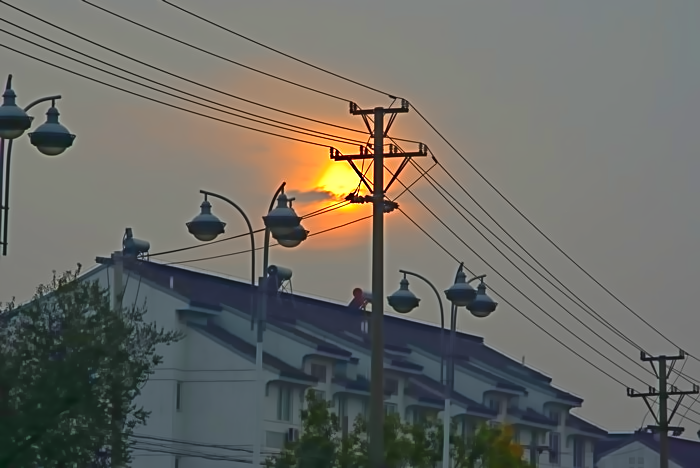}}%
    \subfloat{\includegraphics[width=1.45cm, height=0.97cm]{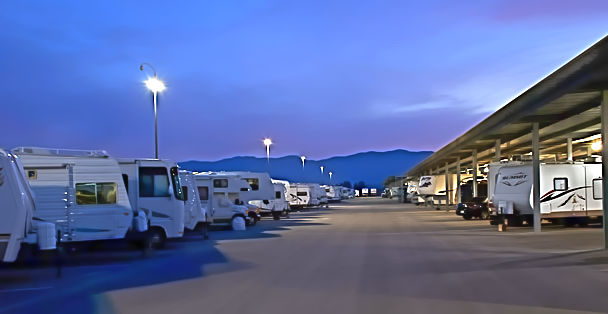}}\\
    \vspace{-1em}
    \subfloat{\includegraphics[width=1.45cm, height=0.97cm]{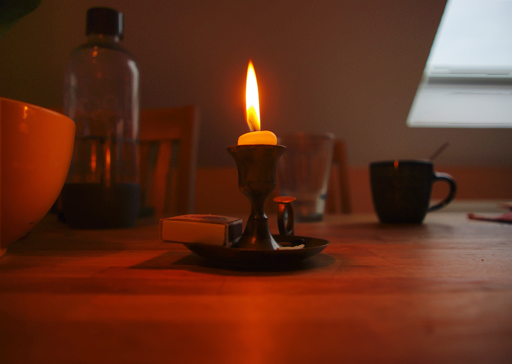}}%
    \subfloat{\includegraphics[width=1.45cm, height=0.97cm]{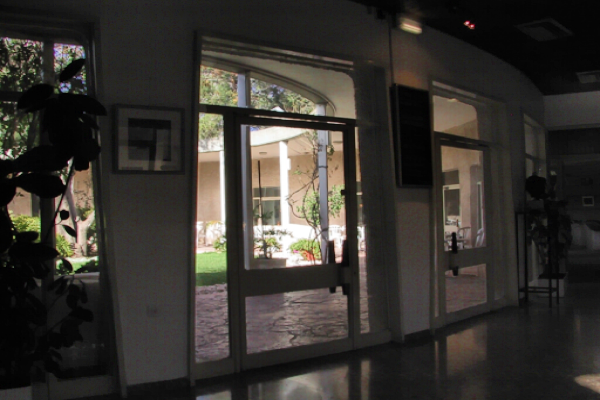}}%
    \subfloat{\includegraphics[width=1.45cm, height=0.97cm]{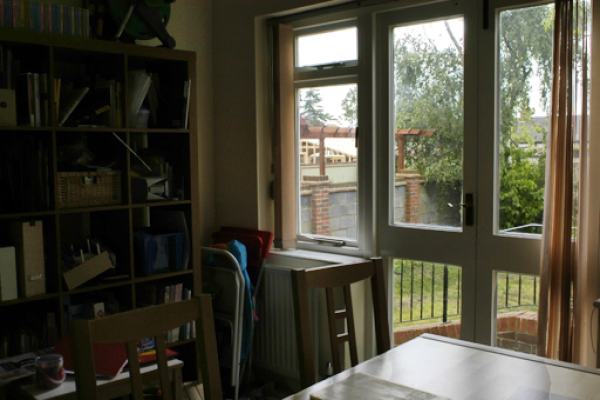}}%
    \subfloat{\includegraphics[width=1.45cm, height=0.97cm]{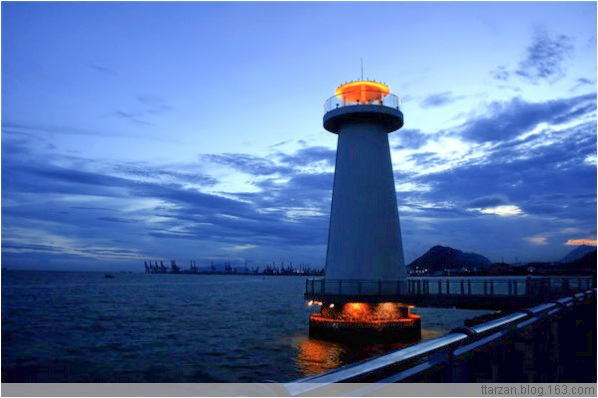}}%
    \subfloat{\includegraphics[width=1.45cm, height=0.97cm]{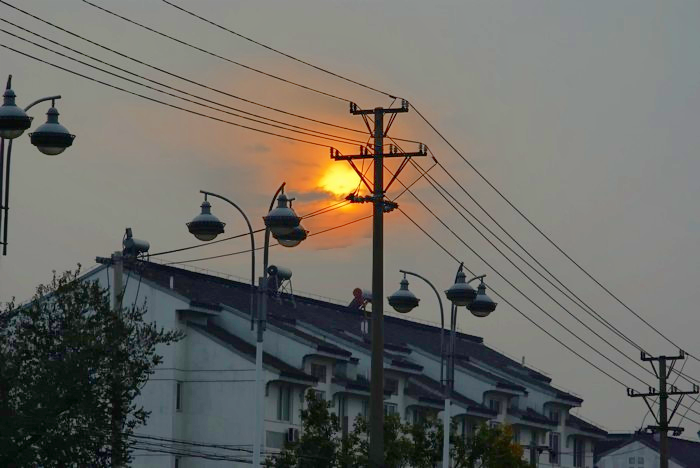}}%
    \subfloat{\includegraphics[width=1.45cm, height=0.97cm]{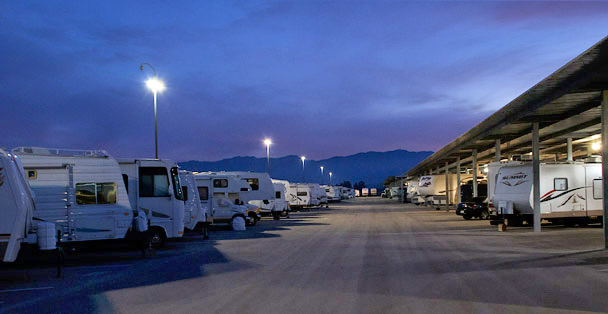}}\\
    \vspace{-1em}
   \subfloat{\includegraphics[width=1.45cm, height=0.97cm]{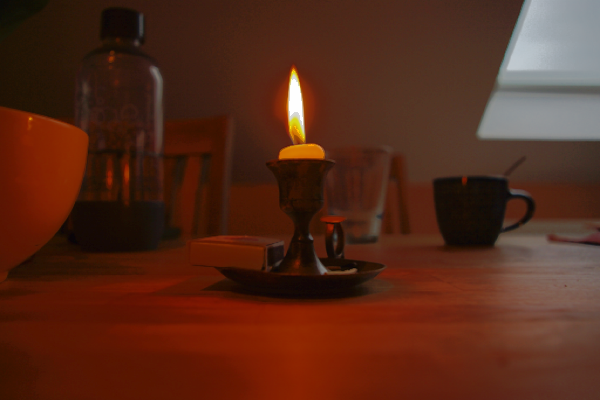}}%
    \subfloat{\includegraphics[width=1.45cm, height=0.97cm]{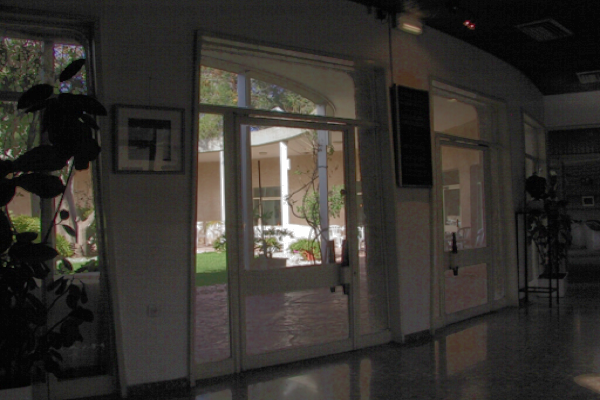}}%
    \subfloat{\includegraphics[width=1.45cm, height=0.97cm]{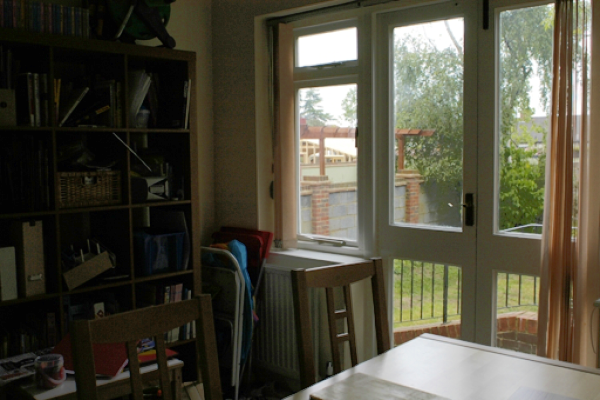}}%
    \subfloat{\includegraphics[width=1.45cm, height=0.97cm]{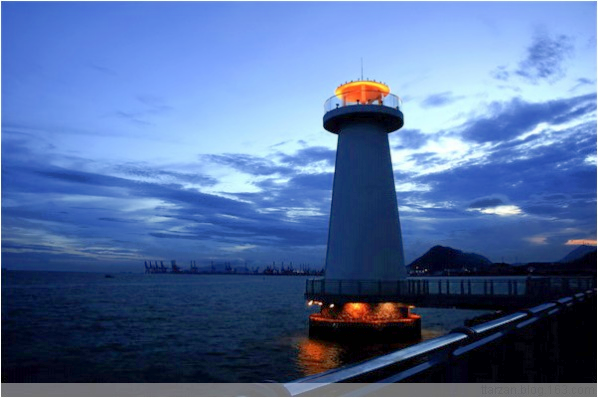}}%
    \subfloat{\includegraphics[width=1.45cm, height=0.97cm]{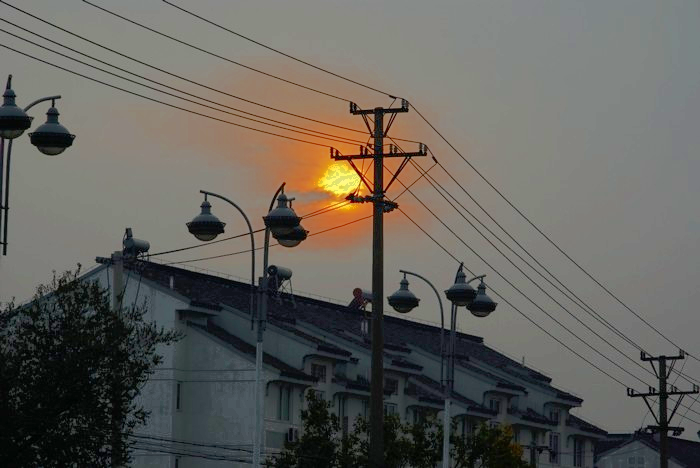}}%
    \subfloat{\includegraphics[width=1.45cm, height=0.97cm]{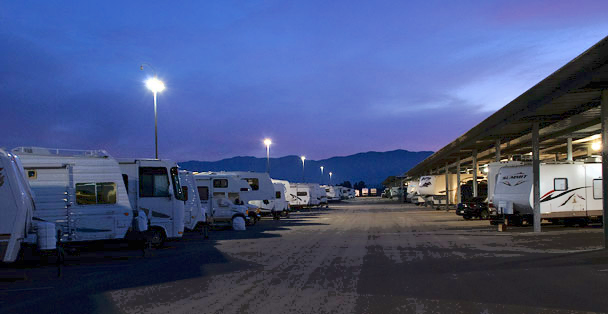}}\\

(a)\hspace{1.1cm}(b)\hspace{1.1cm}(c)\hspace{1.1cm}(d)\hspace{1.1cm}(e)\hspace{1.1cm}(f)
\vspace{-0.2em}
    \caption{Enhancement results of these algorithms in low-light images. (a) \#Candle. (b) \#BelgiumHouse; (c) \#House; (d) \#Daybreak\&Nightfall (11); (e) \#night (52); (f) \#Parking. *From top to bottom, each row denotes low-light image, LIME, STAR, AHPCE, WVM, SSD, RRM, BRAIN, LR3M, FSTAR and proposed algorithm, respectively.}
    \label{fig:grid}
\end{figure}

\subsection{Test and Analysis of Algorithm Generalization}

To further test the performance of those algorithm, we select fifty images from the LOL and LSRW datasets to evaluate the generalization ability of proposed algorithm. The partial enhancement results of these algorithms are given in Fig. 6, and corresponding performance metrics are shown in Table I and Table II.

It can be seen from Table I that the LIME provides the worst performance metrics because it gets enhancement results with an inappropriate illumination range. The AHPCE is better than LIME, but it is still not suitable for the low-light images as an optimal algorithm. The WVM is superior to above compared algorithm in terms of all indexes. However, it still needs to be improved. Although LR3M achieves satisfactory enhancement results, its performance is not stable, as evidenced by the case of image \#2063. The RRM obtains promising performance metrics due to it separates the noise from the low-light images. The FSTAR, STAR and BRAIN provide better performance metrics because they consider the structure and texture information. The proposed algorithm outperforms its peers in PSNR, SSIM and NIQE indexes on the LOL dataset. From Table II, the proposed algorithm is still better than all compared algorithms on the LSRW dataset for the PSNR, SSIM and NIQE. It is clear that the proposed algorithm has promising generalization.

\begin{table}[!t]
\caption{Average Performance Metrics of Ten Algorithms on LOL Dataset\label{tab:table1}}
\centering
\begin{tabular}{ccccc}
\toprule
Algorithm & PSNR & SSIM & LOE & NIQE\\
\midrule
LIME & 7.9358 & 0.1633 & 1554.0136 & 7.7757 \\
STAR & 15.9591 & 0.4012 & 190.5736 & 6.9190 \\
AHPCE & 9.3117 & 0.2176 & 212.7560 & 5.7862 \\
WVM & 10.6284 & 0.2157 & 23.0453 & 8.1876 \\
SSD & 14.2922 & 0.3690 & 339.6008 & 4.2576 \\
RRM & 12.9250 & 0.3048 & 209.7637 & \textbf{3.9505} \\
BRIAN & 17.6792 & 0.4796 & 274.6931 & 6.8172 \\
LR3M & 18.2804 & 0.4797 & 622.3237 & 7.4315 \\
FSTAR & 16.8270 & 0.4866 & 156.4176 & 7.1306 \\
proposed & \textbf{19.3645} & \textbf{0.6771} & \textbf{8.4307} & 3.9669 \\
\bottomrule
\end{tabular}
\end{table}

\begin{table}[!t]
\caption{Average Performance Metrics of Ten Algorithms on LSRW Dataset\label{tab:table2}}
\centering
\begin{tabular}{ccccc}
\toprule
Algorithm & PSNR & SSIM & LOE & NIQE\\
\midrule
LIME & 7.2494 & 0.2229 & 736.2451 & 2.5515 \\
STAR & 14.6078 & 0.4669 & 161.9688 & \textbf{2.4379} \\
AHPCE &12.1680 & 0.3399 & 188.2877 & 2.6628 \\
WVM & 14.2559 & 0.4439 & 13.4240 & 3.1618 \\
SSD & 13.2160 & 0.4604 & 215.2995 & 3.6585  \\
RRM & 13.2133 & 0.4165  & 293.5877 & 3.4590 \\
BRIAN & 16.6532 & 0.5535 	& 260.1856 	&2.9410 \\
LR3M & 13.2282 & 0.4073 	& 368.5032 	&4.4232 \\
FSTAR & 16.0134 & 0.5802 	& 106.2363 	&3.1147 \\
proposed & \textbf{23.8451} & \textbf{0.8630} 	& \textbf{6.0168} 	&2.7427 \\
\bottomrule
\end{tabular}
\end{table}

\subsection{Parameter Sensitivity of Algorithm}
For the proposed algorithm, the illumination weight $\gamma$  is an important parameter, which significantly affects the visual region and lightness of the enhanced image. Therefore, we will discuss and analyze the impact of changes in   on algorithm performance. The number of low-light images are \#Cadik, which is from the \#MEF dataset. We tune $\gamma$  from 1 to 10 in the image \#Cadik. The enhanced results and the histogram of the proposed algorithm are shown in Fig~7, and corresponding performance metrics are given in Table III.

\begin{table}[!t]
\caption{Performance Metrics with Different \(\gamma\) \label{tab:gamma_table}}
\centering
\begin{tabular}{ccccc}
\toprule
Parameters & PSNR & SSIM & LOE & NIQE\\
\midrule
$\gamma=1$ & \textbf{41.5052} & \textbf{0.9949} & \textbf{0.1157} & 3.4215 \\
$\gamma=1.5$ & 28.2906 & 0.7184 & 1.1357 & 3.7289 \\
$\gamma=2.2$ & 21.0543 & 0.5732 & 10.3315 & \textbf{2.4674} \\
$\gamma=5$ & 13.0649 & 0.3044 & 13.9653 & 2.9502 \\
$\gamma=10$ & 9.8295 & 0.2491 & 16.9723 & 2.9909 \\
\bottomrule
\end{tabular}
\end{table}

From Fig. 7, as the increase of $\gamma $, the visual region and the lightness of the enhanced image increase. When $\gamma =1$, the proposed algorithm provides the dark results, which are similar to the test image and have the worst enhancement effect. When $\gamma =1.5$, the proposed algorithm enhances the dark region of the image, but has a poor effect. When $\gamma =2.2$, the lightness of the dark region has further improved and has better enhancement effect. When $\gamma =5$ and $\gamma =10$, the proposed algorithm provides higher lightness, but the enhanced results with many noise points. As can be seen in Table III, the proposed algorithm provides the best performance metrics in $\gamma =1$ in terms of PSNR, SSIM and LOE. 

\begin{figure*}[htbp]
    \centering
    \subfloat{\includegraphics[width=1.45cm, height=0.97cm]{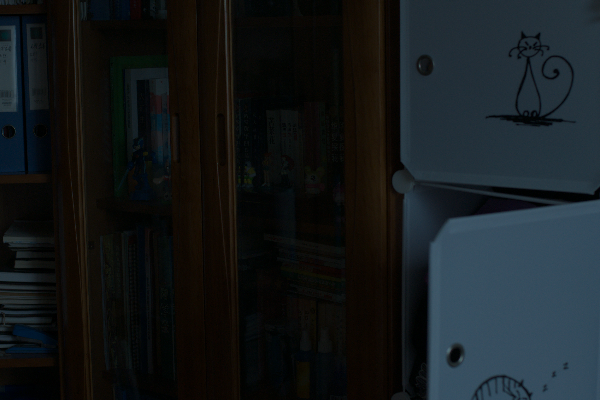}}%
    \subfloat{\includegraphics[width=1.45cm, height=0.97cm]{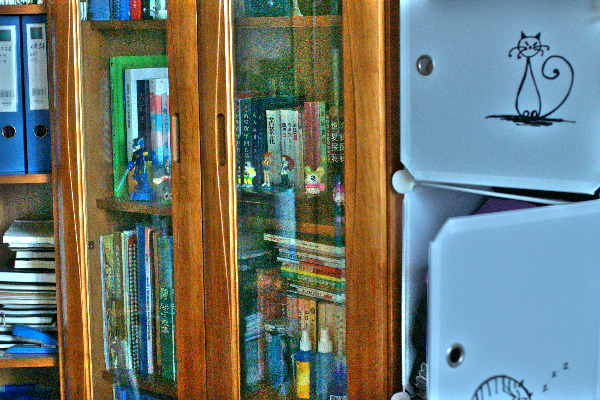}}%
    \subfloat{\includegraphics[width=1.45cm, height=0.97cm]{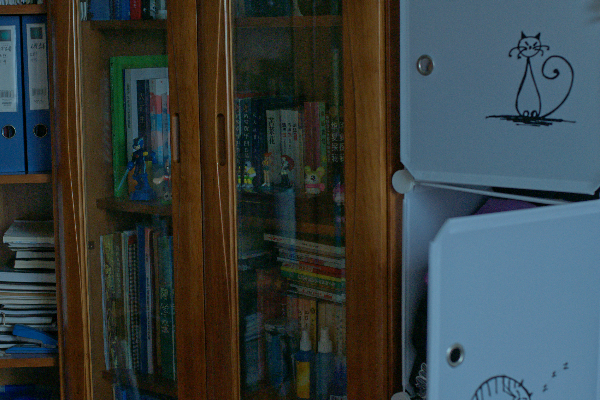}}%
    \subfloat{\includegraphics[width=1.45cm, height=0.97cm]{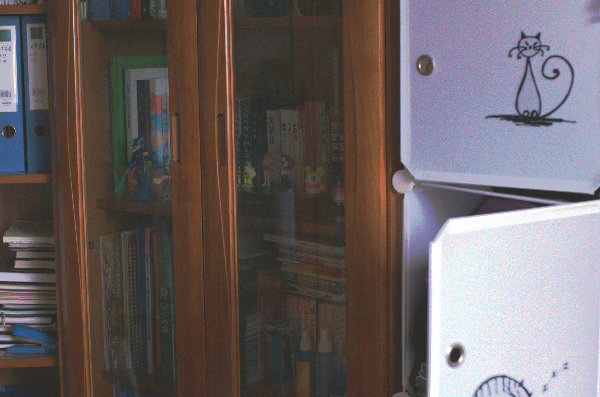}}%
    \subfloat{\includegraphics[width=1.45cm, height=0.97cm]{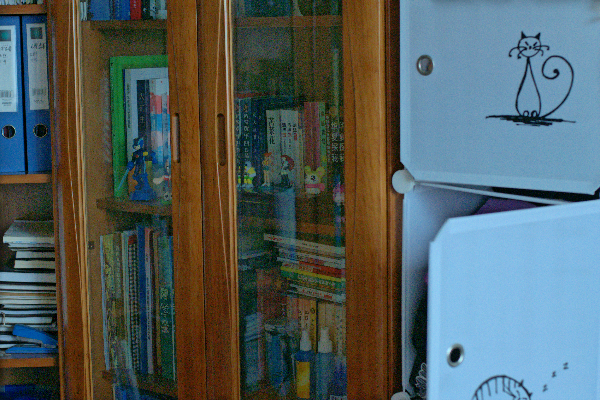}}%
    \subfloat{\includegraphics[width=1.45cm, height=0.97cm]{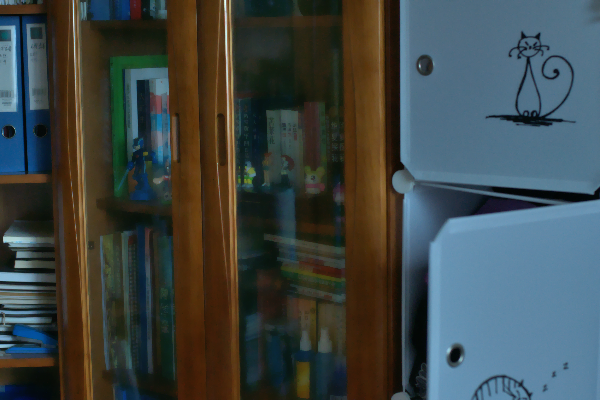}}
    \subfloat{\includegraphics[width=1.45cm, height=0.97cm]{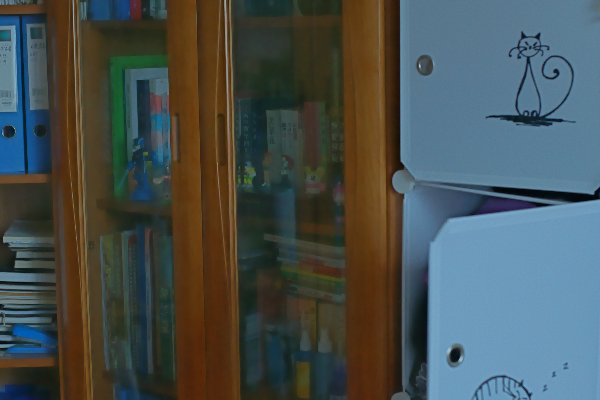}}%
    \subfloat{\includegraphics[width=1.45cm, height=0.97cm]{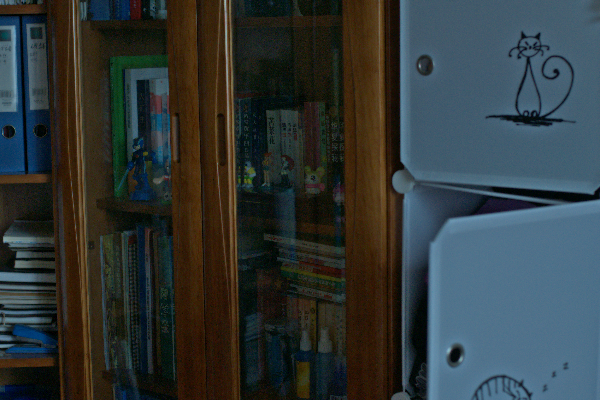}}%
    \subfloat{\includegraphics[width=1.45cm, height=0.97cm]{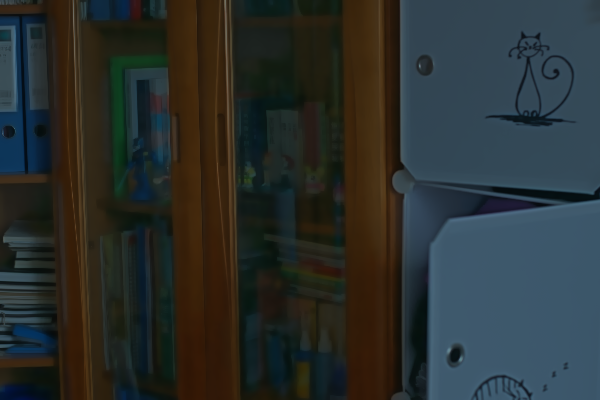}}%
    \subfloat{\includegraphics[width=1.45cm, height=0.97cm]{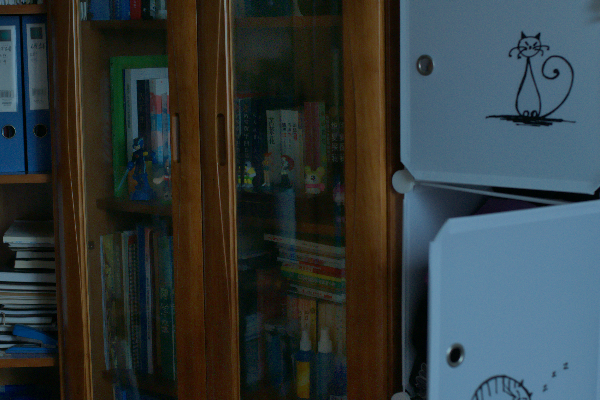}}%
    \subfloat{\includegraphics[width=1.45cm, height=0.97cm]{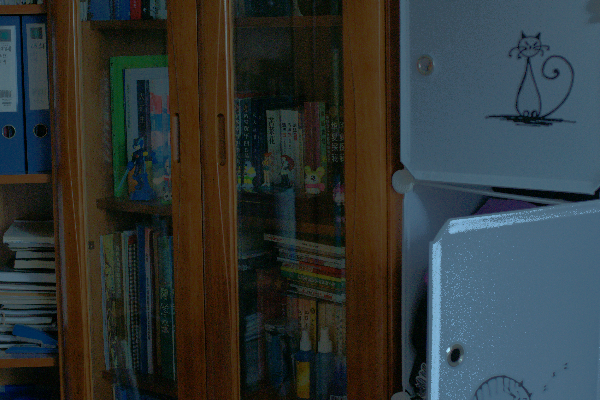}}\\
    \vspace{-1em}
\subfloat{\includegraphics[width=1.45cm, height=0.97cm]{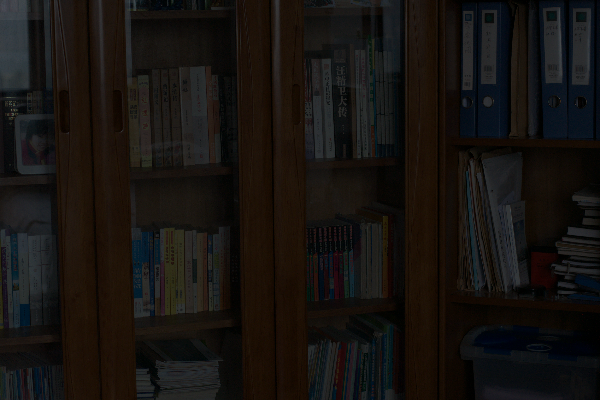}}%
\subfloat{\includegraphics[width=1.45cm, height=0.97cm]{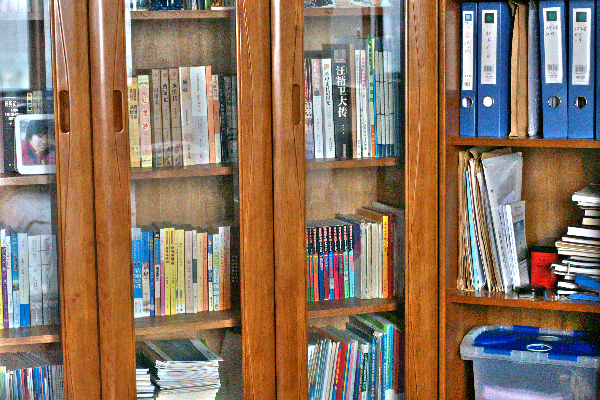}}%
\subfloat{\includegraphics[width=1.45cm, height=0.97cm]{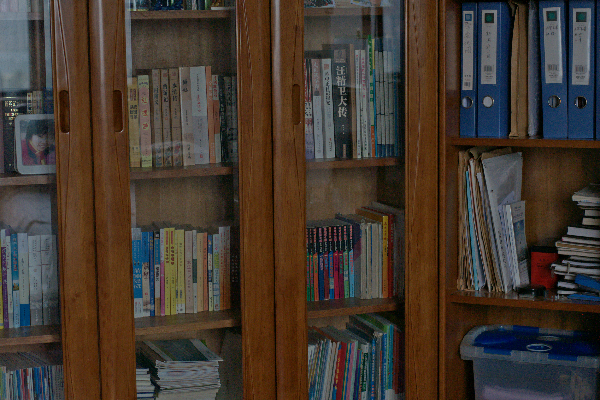}}%
\subfloat{\includegraphics[width=1.45cm, height=0.97cm]{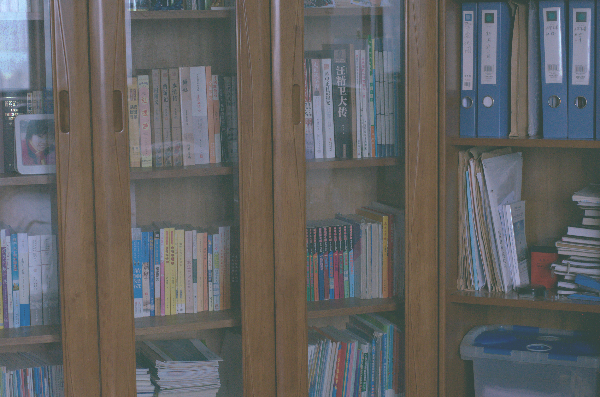}}%
\subfloat{\includegraphics[width=1.45cm, height=0.97cm]{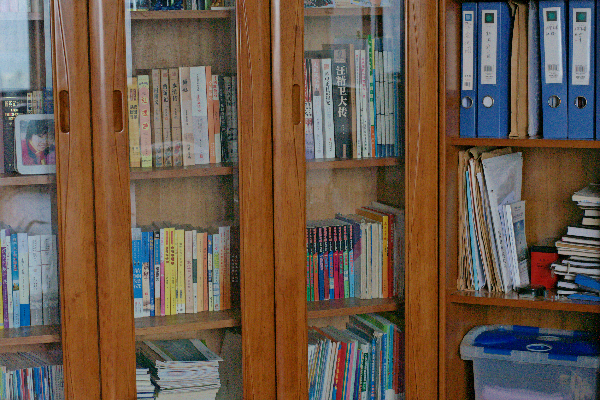}}%
\subfloat{\includegraphics[width=1.45cm, height=0.97cm]{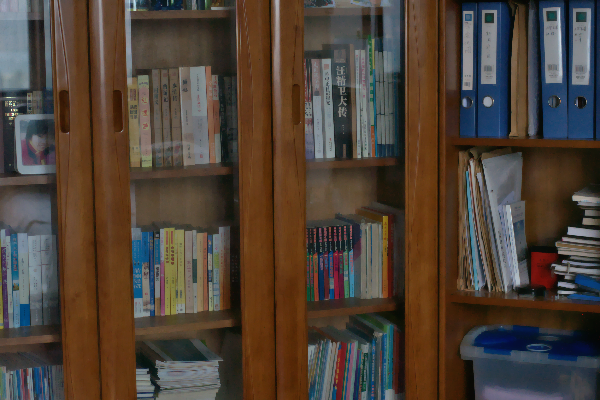}}%
\subfloat{\includegraphics[width=1.45cm, height=0.97cm]{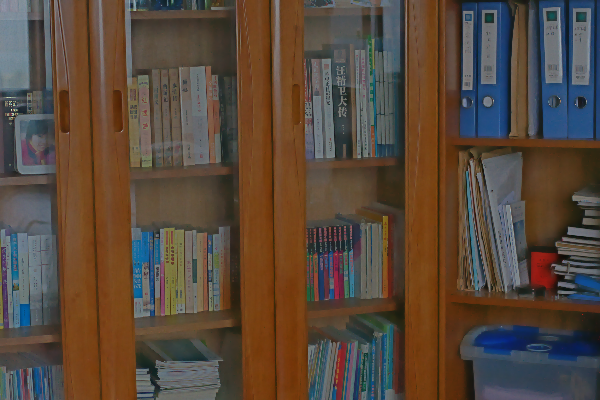}}%
\subfloat{\includegraphics[width=1.45cm, height=0.97cm]{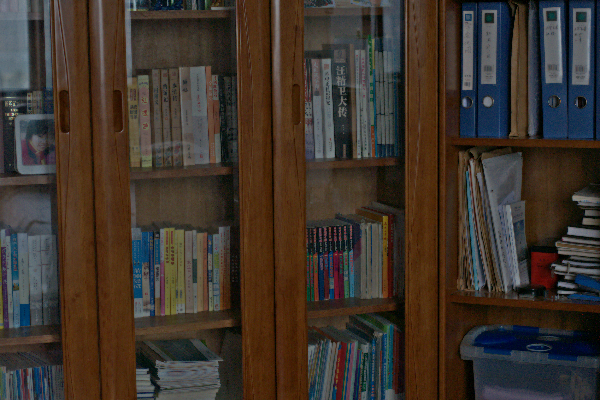}}%
\subfloat{\includegraphics[width=1.45cm, height=0.97cm]{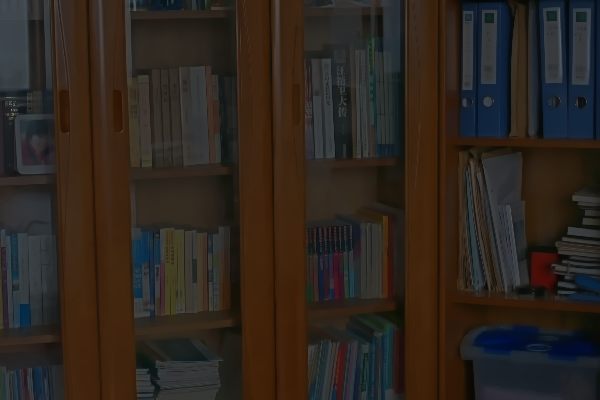}}%
\subfloat{\includegraphics[width=1.45cm, height=0.97cm]{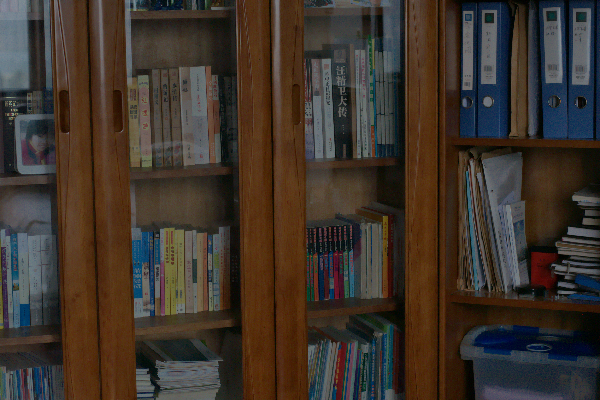}}%
\subfloat{\includegraphics[width=1.45cm, height=0.97cm]{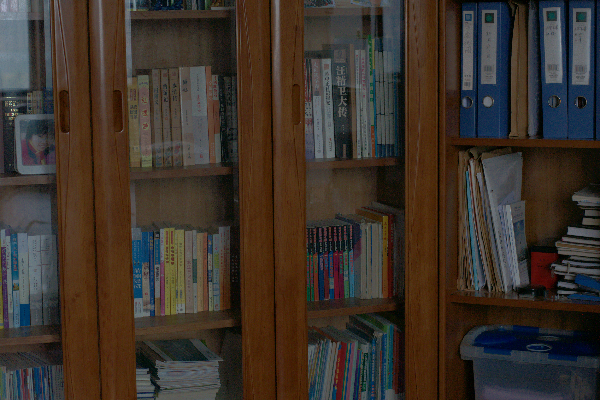}}\\
    \vspace{-1em}
\subfloat{\includegraphics[width=1.45cm, height=0.97cm]{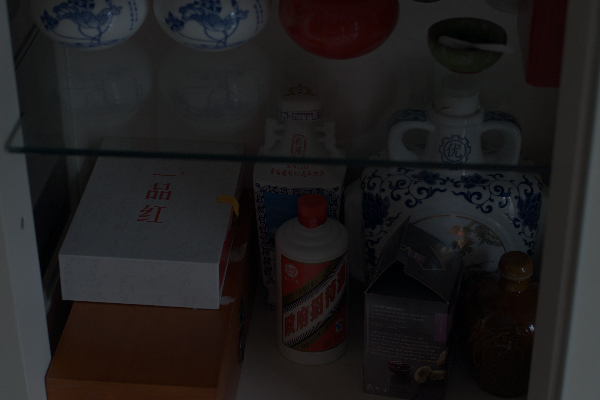}}%
\subfloat{\includegraphics[width=1.45cm, height=0.97cm]{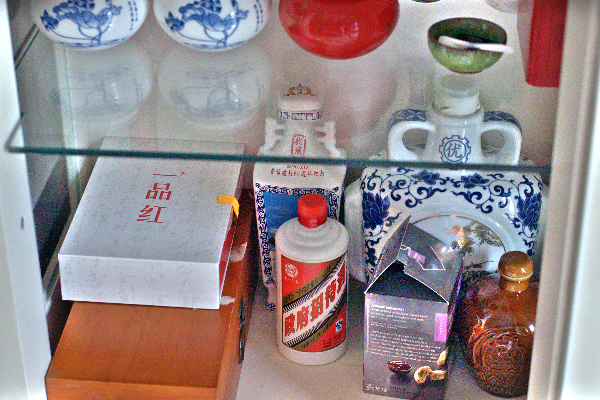}}%
\subfloat{\includegraphics[width=1.45cm, height=0.97cm]{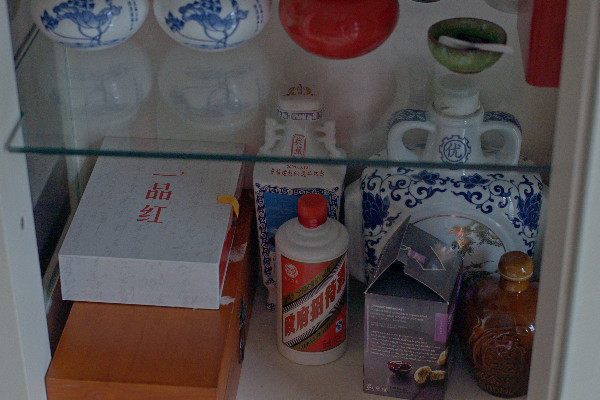}}%
\subfloat{\includegraphics[width=1.45cm, height=0.97cm]{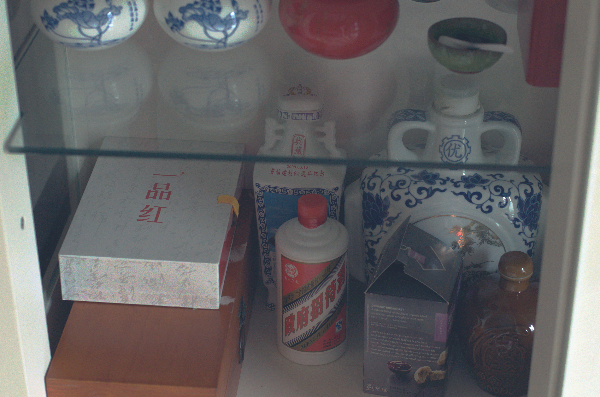}}%
\subfloat{\includegraphics[width=1.45cm, height=0.97cm]{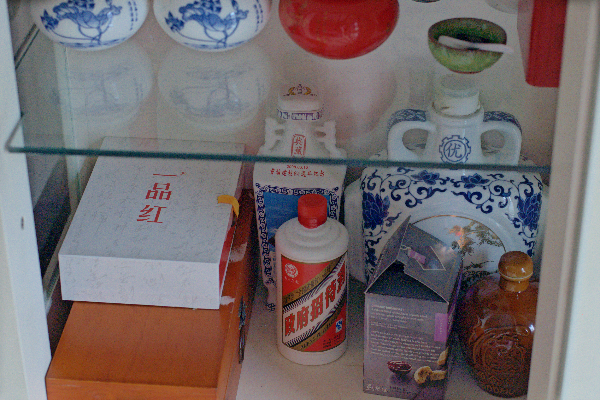}}%
\subfloat{\includegraphics[width=1.45cm, height=0.97cm]{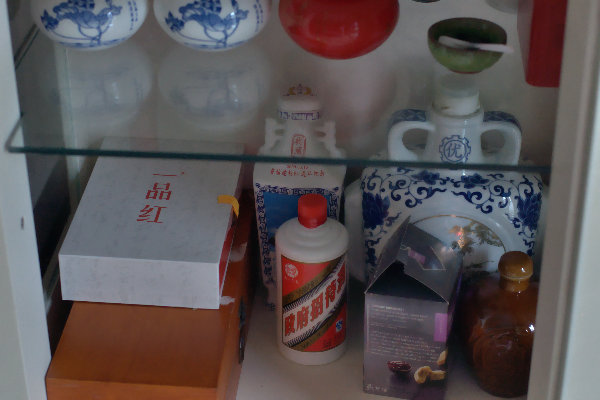}}%
\subfloat{\includegraphics[width=1.45cm, height=0.97cm]{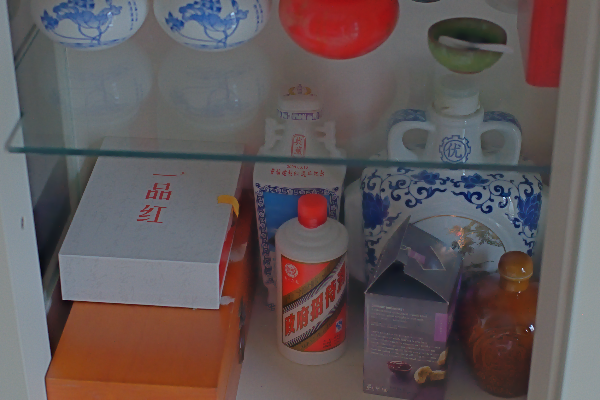}}%
\subfloat{\includegraphics[width=1.45cm, height=0.97cm]{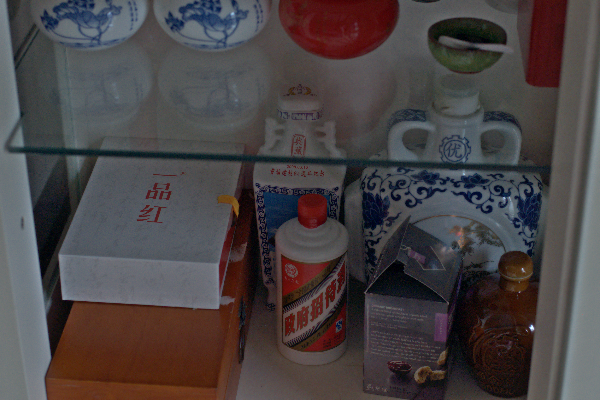}}%
\subfloat{\includegraphics[width=1.45cm, height=0.97cm]{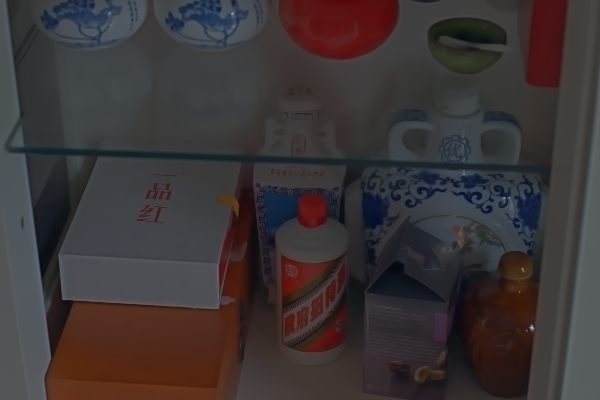}}%
\subfloat{\includegraphics[width=1.45cm, height=0.97cm]{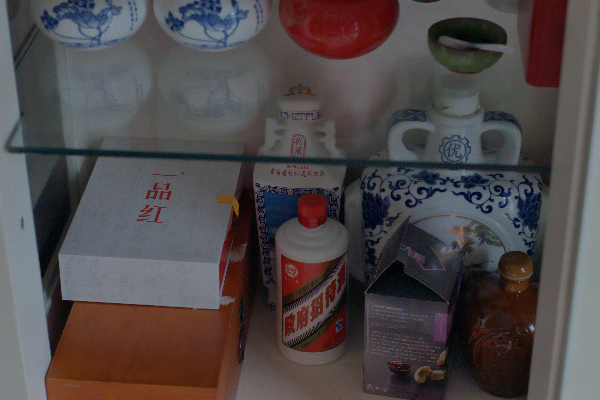}}%
\subfloat{\includegraphics[width=1.45cm, height=0.97cm]{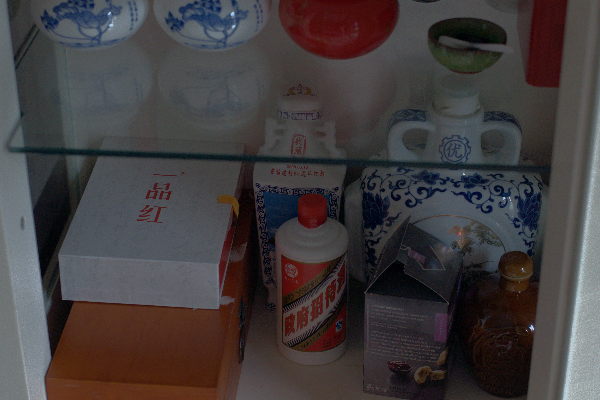}}\\
    \vspace{-1em}
 \subfloat{\includegraphics[width=1.45cm, height=0.97cm]{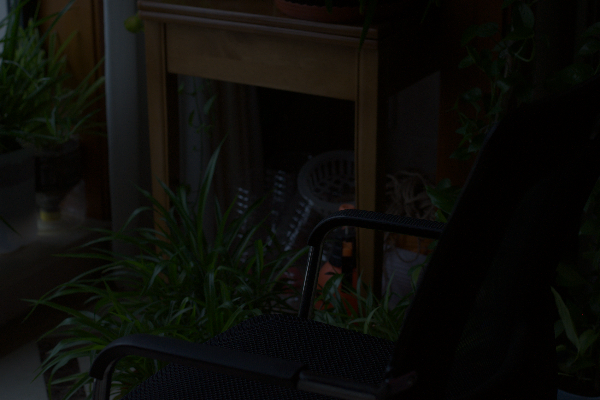}}%
\subfloat{\includegraphics[width=1.45cm, height=0.97cm]{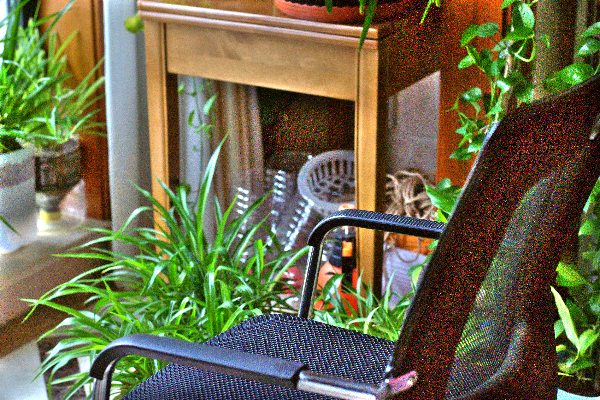}}%
\subfloat{\includegraphics[width=1.45cm, height=0.97cm]{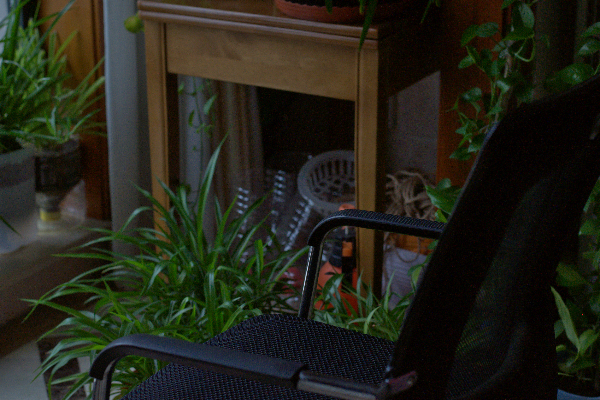}}%
\subfloat{\includegraphics[width=1.45cm, height=0.97cm]{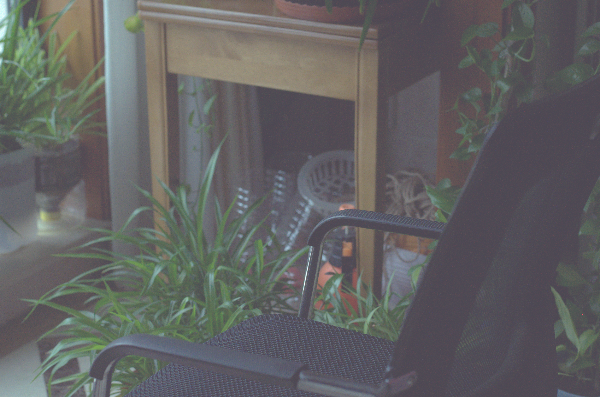}}%
\subfloat{\includegraphics[width=1.45cm, height=0.97cm]{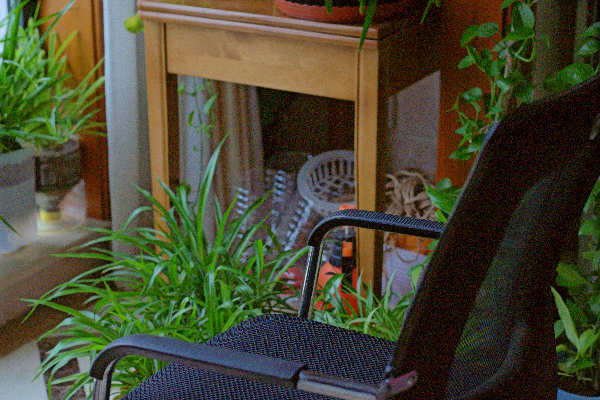}}%
\subfloat{\includegraphics[width=1.45cm, height=0.97cm]{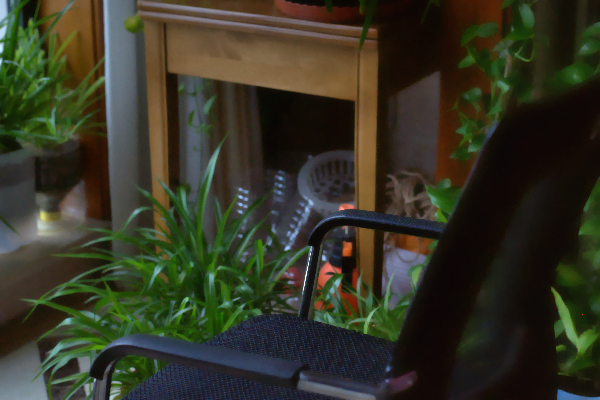}}%
\subfloat{\includegraphics[width=1.45cm, height=0.97cm]{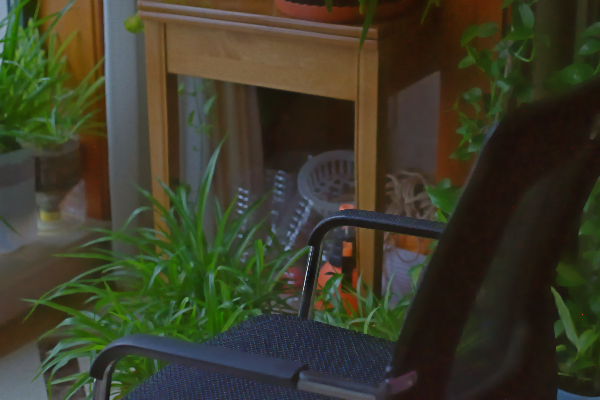}}%
\subfloat{\includegraphics[width=1.45cm, height=0.97cm]{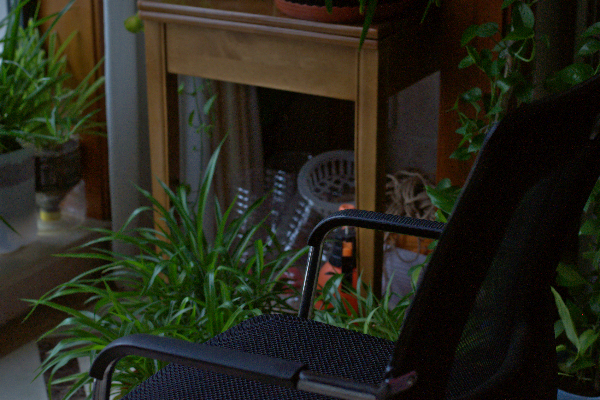}}%
\subfloat{\includegraphics[width=1.45cm, height=0.97cm]{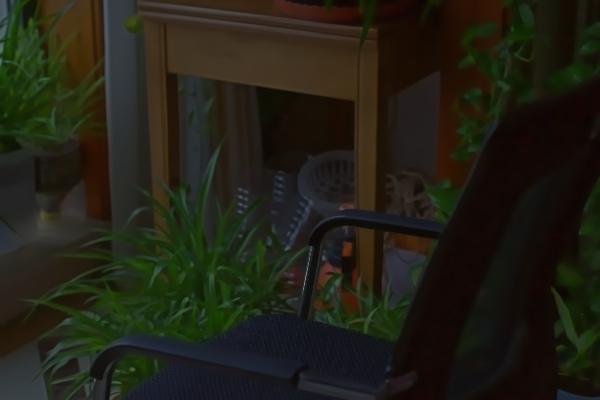}}%
\subfloat{\includegraphics[width=1.45cm, height=0.97cm]{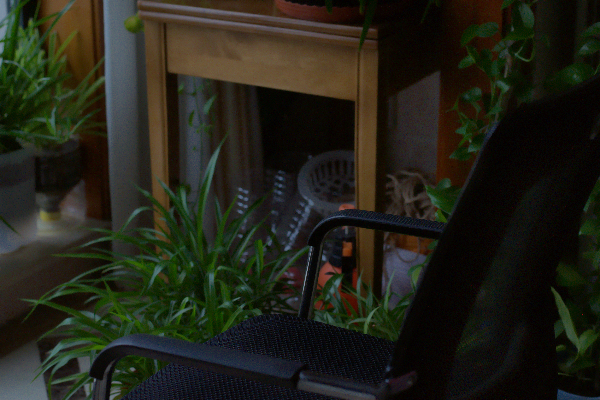}}%
\subfloat{\includegraphics[width=1.45cm, height=0.97cm]{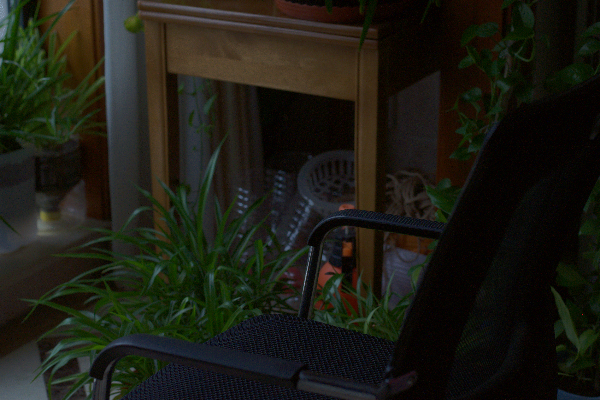}}\\
    \vspace{-1em}
\subfloat{\includegraphics[width=1.45cm, height=0.97cm]{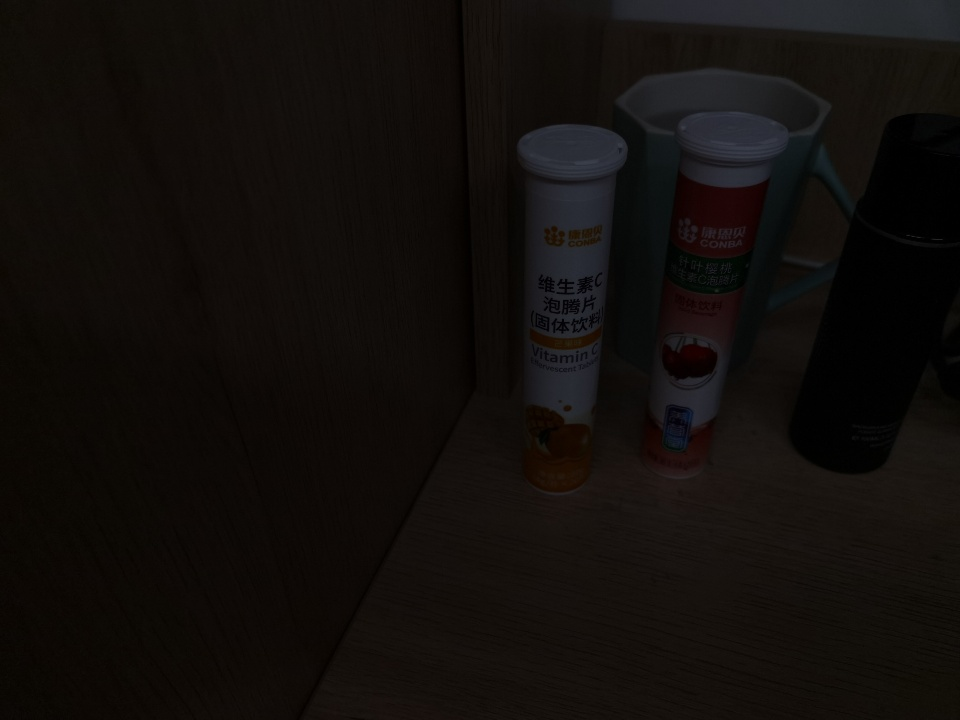}}%
\subfloat{\includegraphics[width=1.45cm, height=0.97cm]{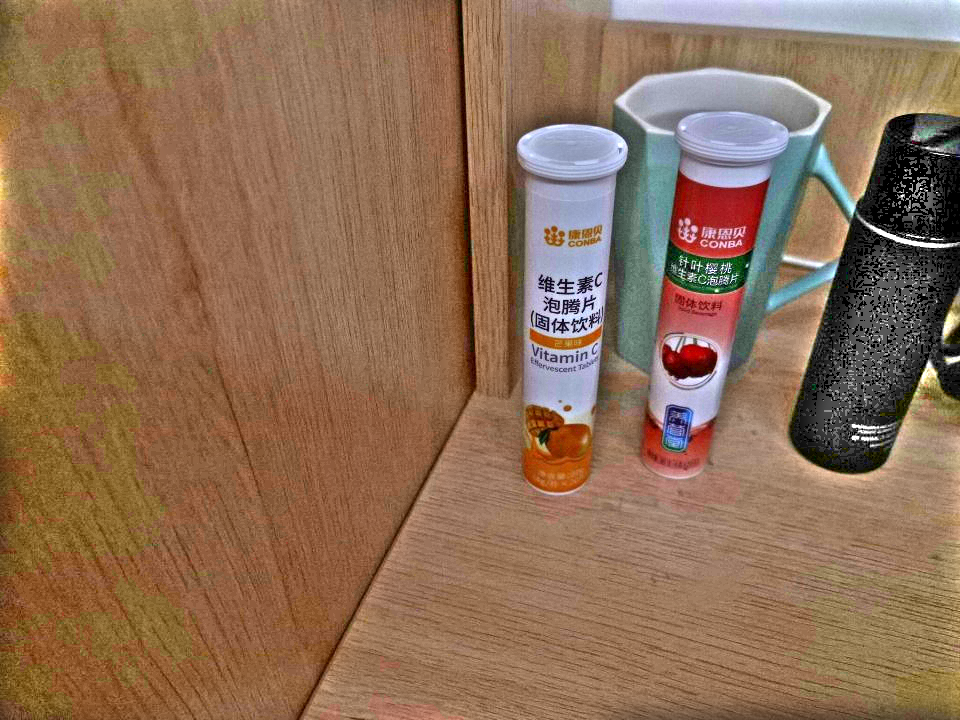}}%
\subfloat{\includegraphics[width=1.45cm, height=0.97cm]{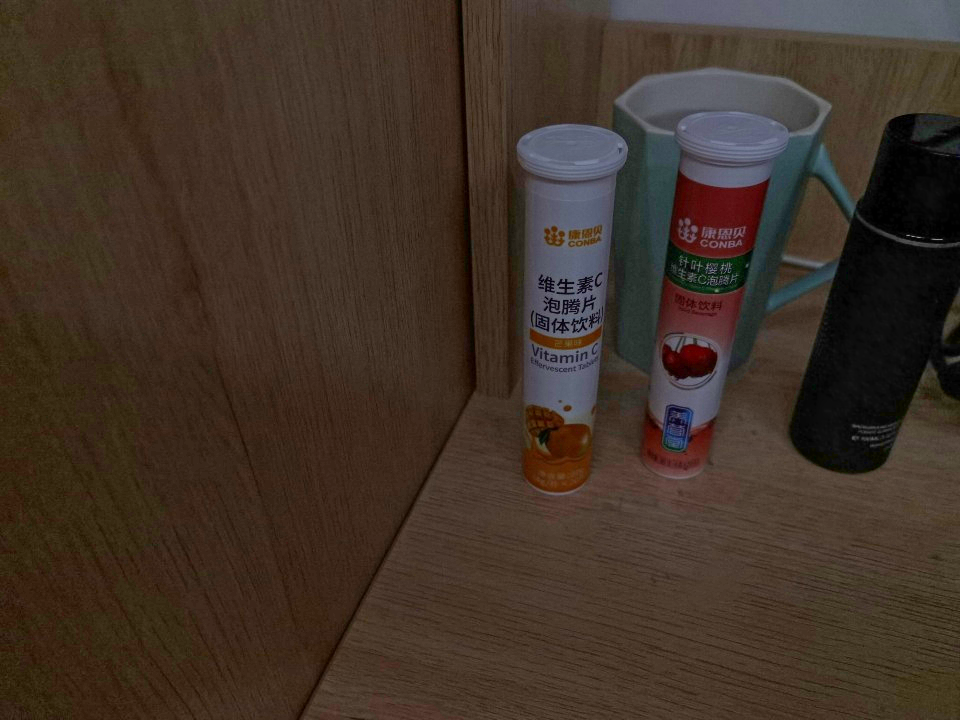}}%
\subfloat{\includegraphics[width=1.45cm, height=0.97cm]{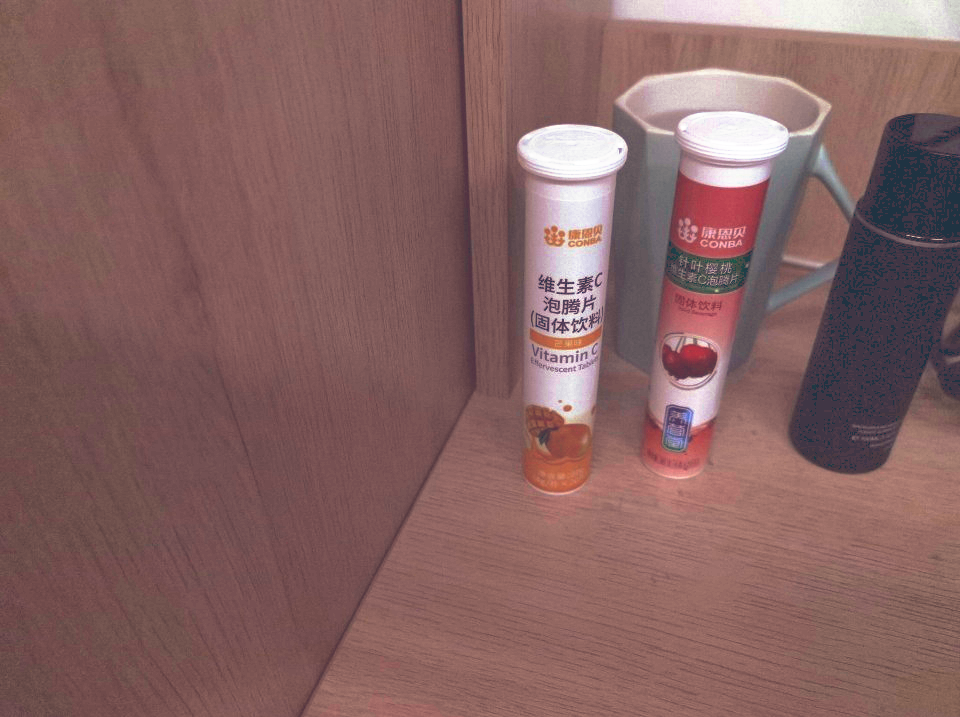}}%
\subfloat{\includegraphics[width=1.45cm, height=0.97cm]{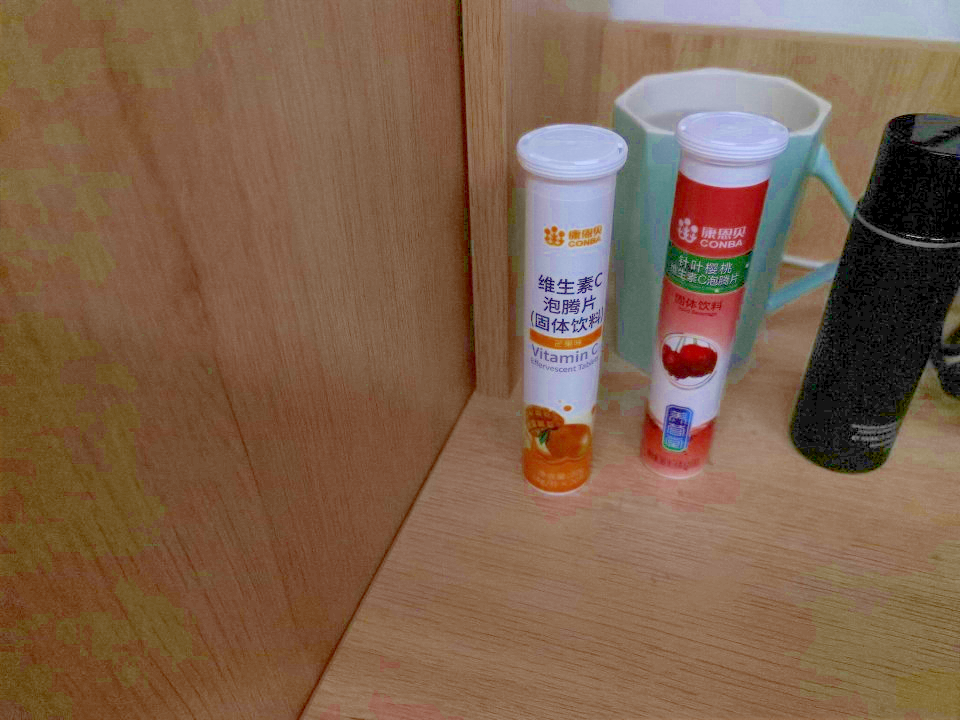}}%
\subfloat{\includegraphics[width=1.45cm, height=0.97cm]{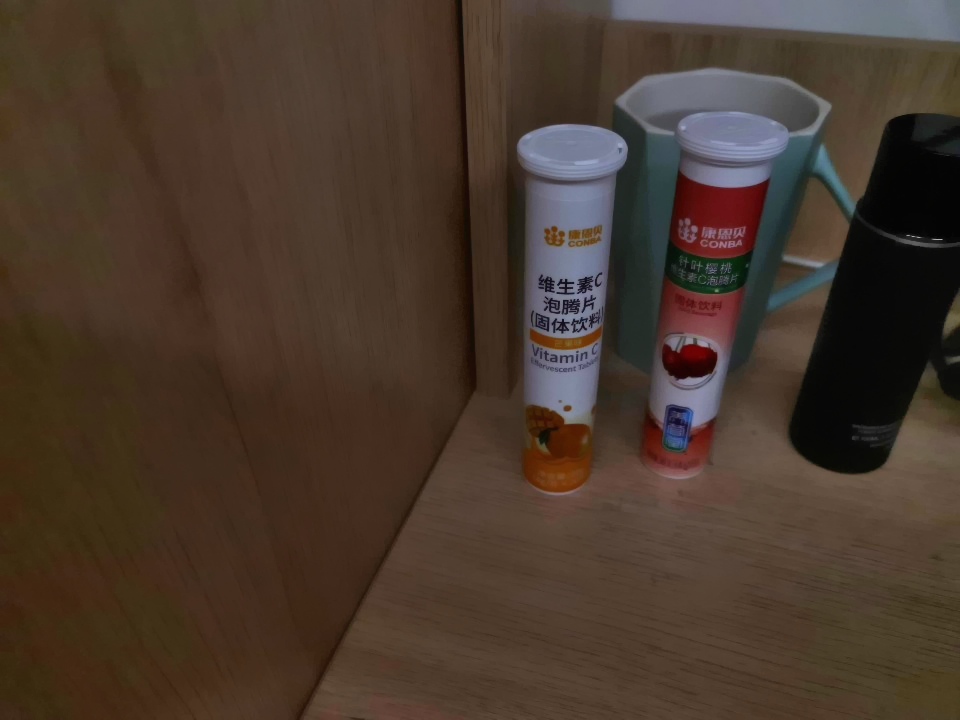}}%
\subfloat{\includegraphics[width=1.45cm, height=0.97cm]{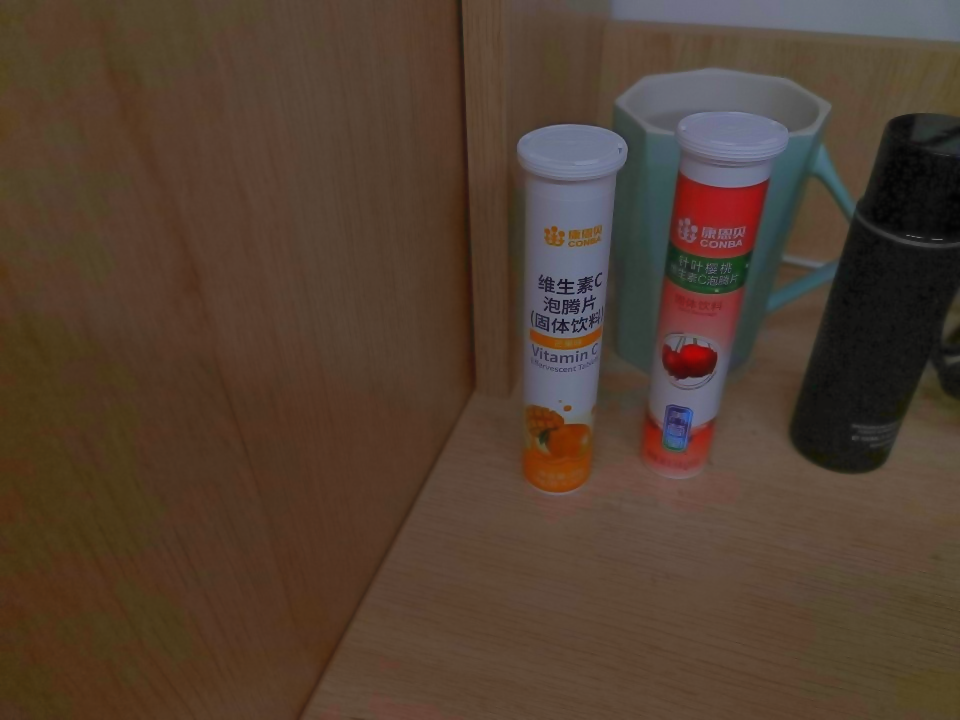}}%
\subfloat{\includegraphics[width=1.45cm, height=0.97cm]{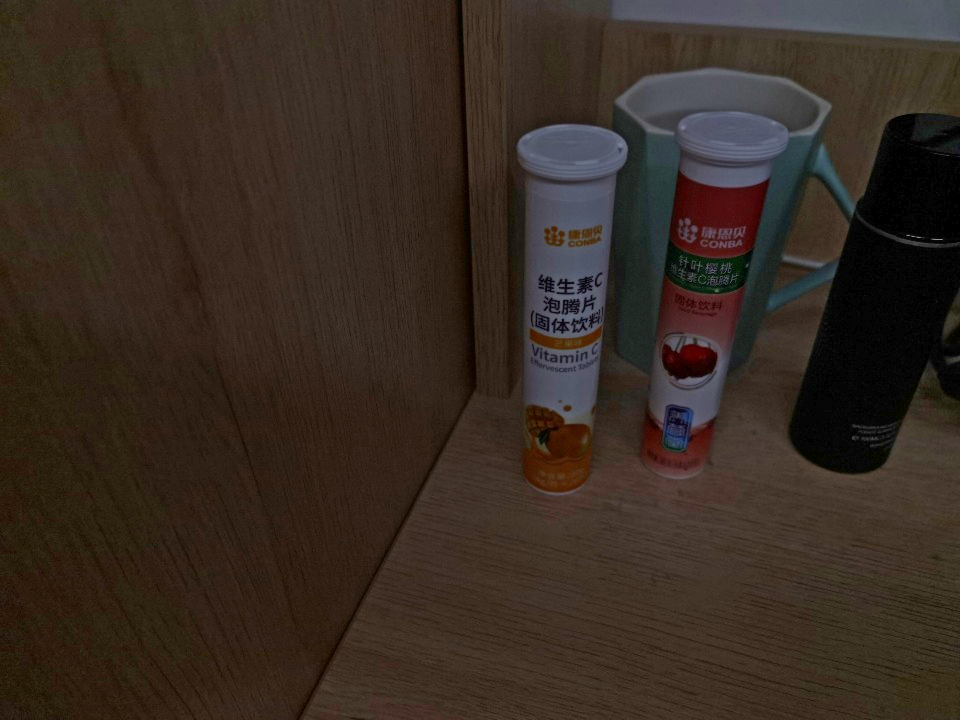}}%
\subfloat{\includegraphics[width=1.45cm, height=0.97cm]{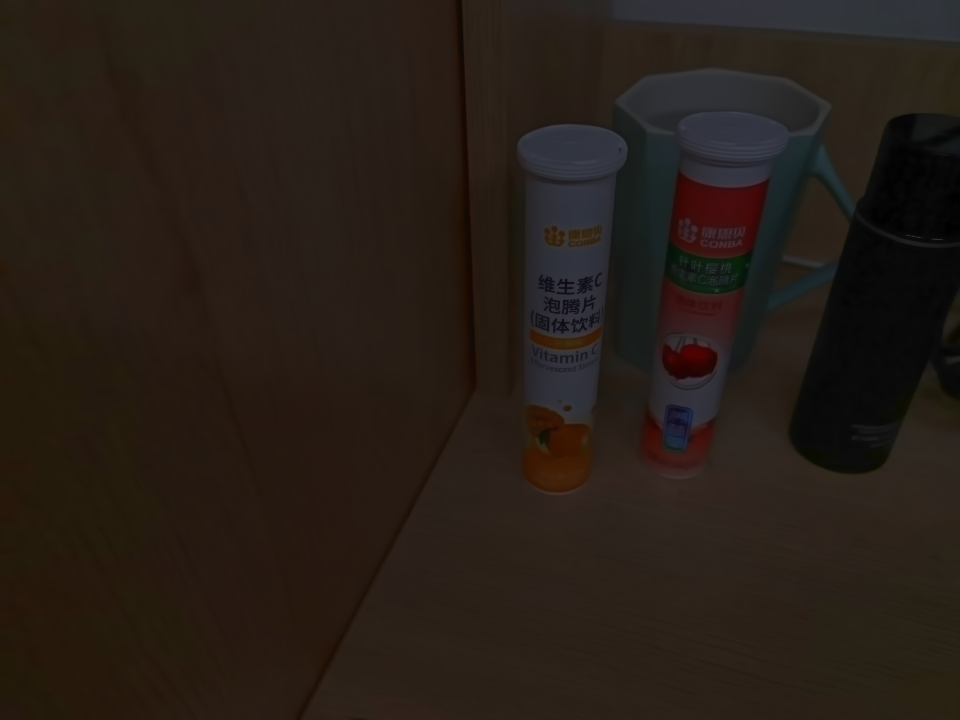}}%
\subfloat{\includegraphics[width=1.45cm, height=0.97cm]{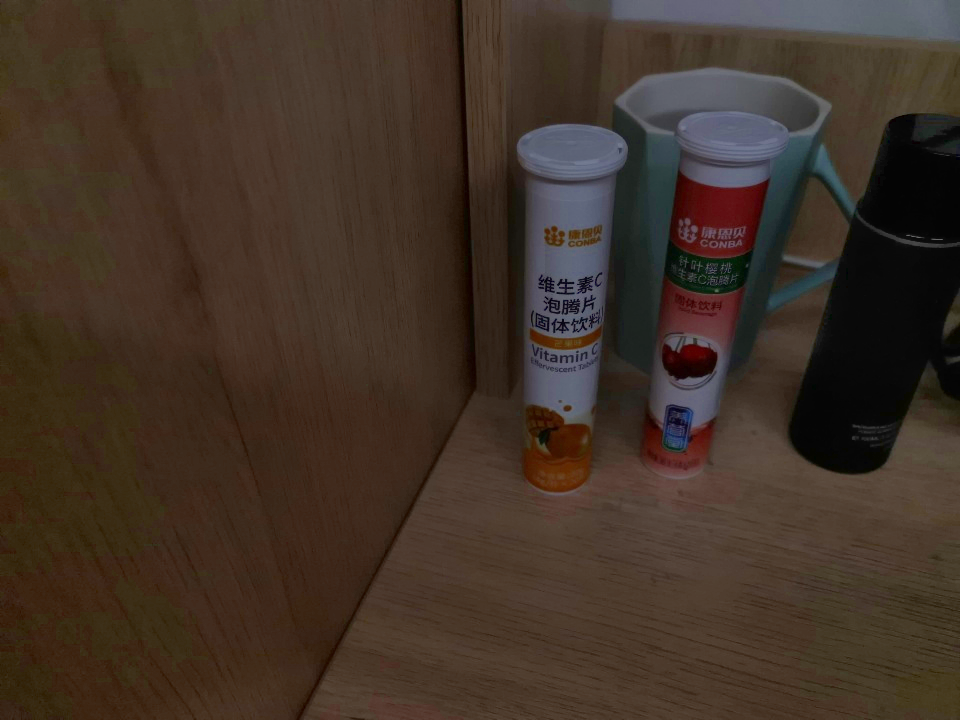}}%
\subfloat{\includegraphics[width=1.45cm, height=0.97cm]{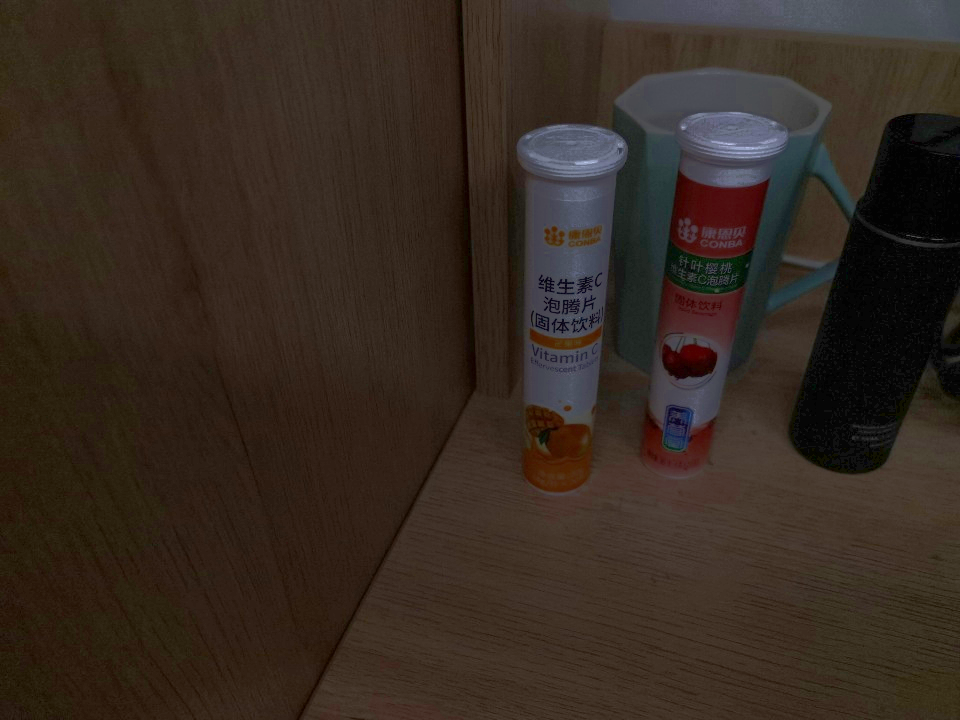}}\\
    \vspace{-1em}
\subfloat{\includegraphics[width=1.45cm, height=0.97cm]{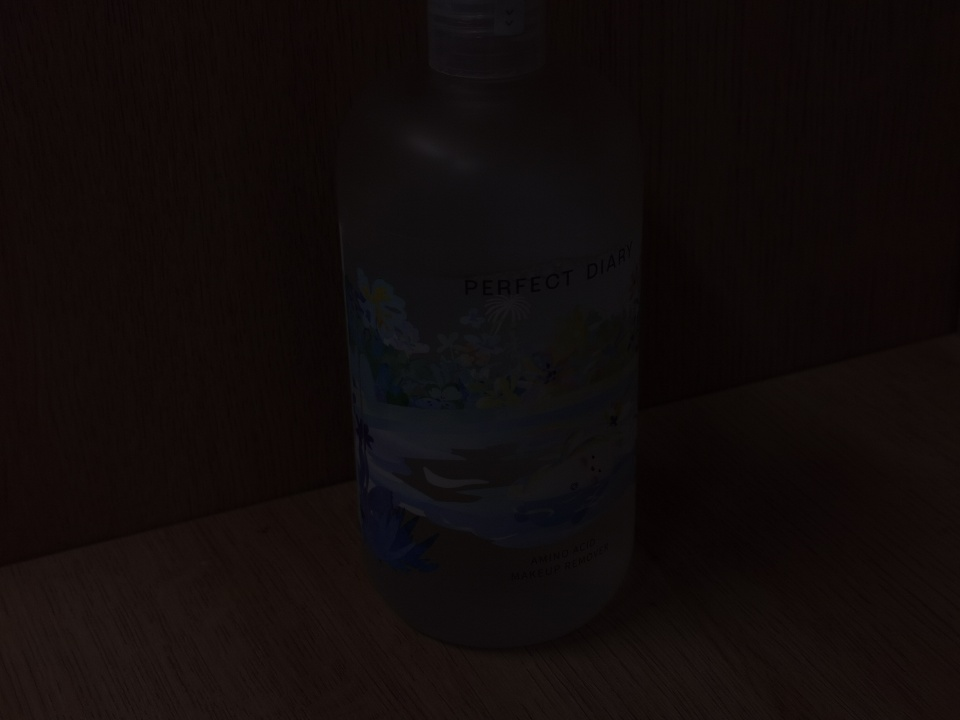}}%
\subfloat{\includegraphics[width=1.45cm, height=0.97cm]{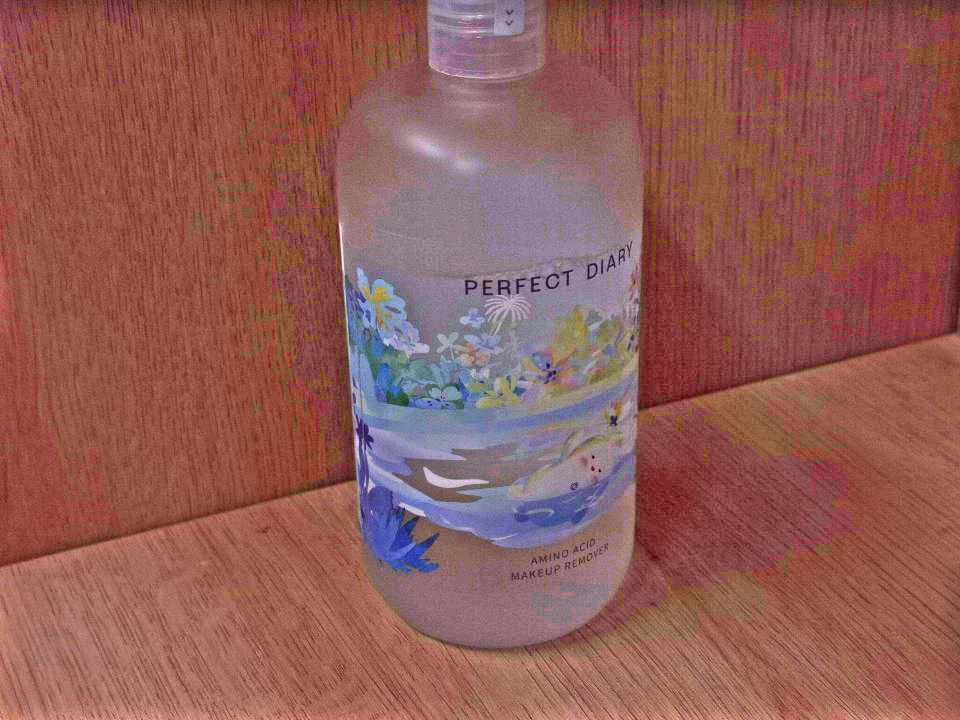}}%
\subfloat{\includegraphics[width=1.45cm, height=0.97cm]{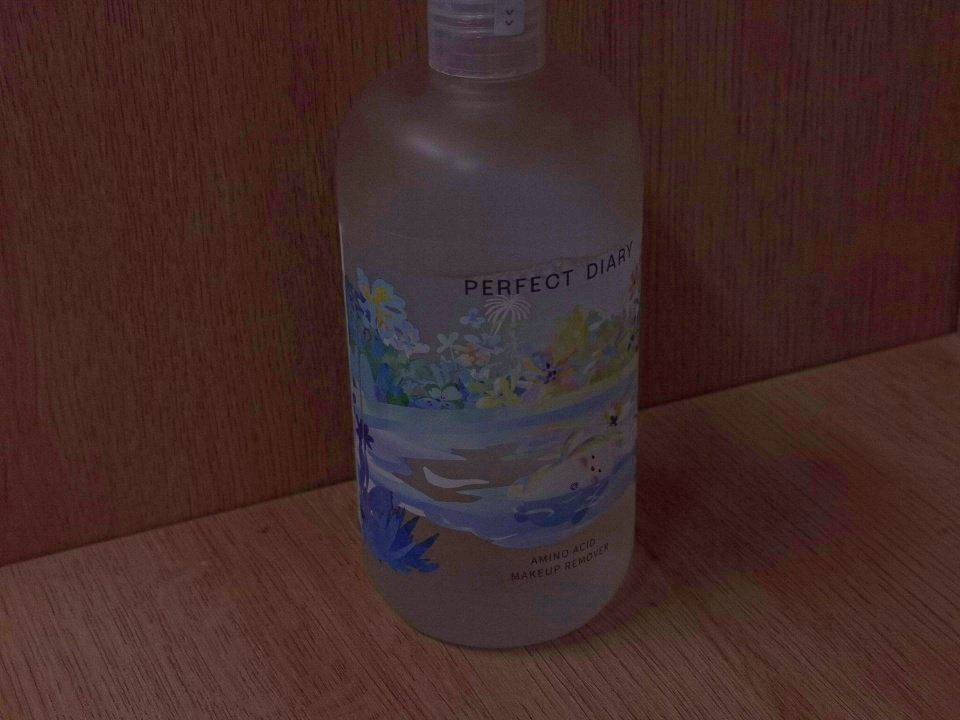}}%
\subfloat{\includegraphics[width=1.45cm, height=0.97cm]{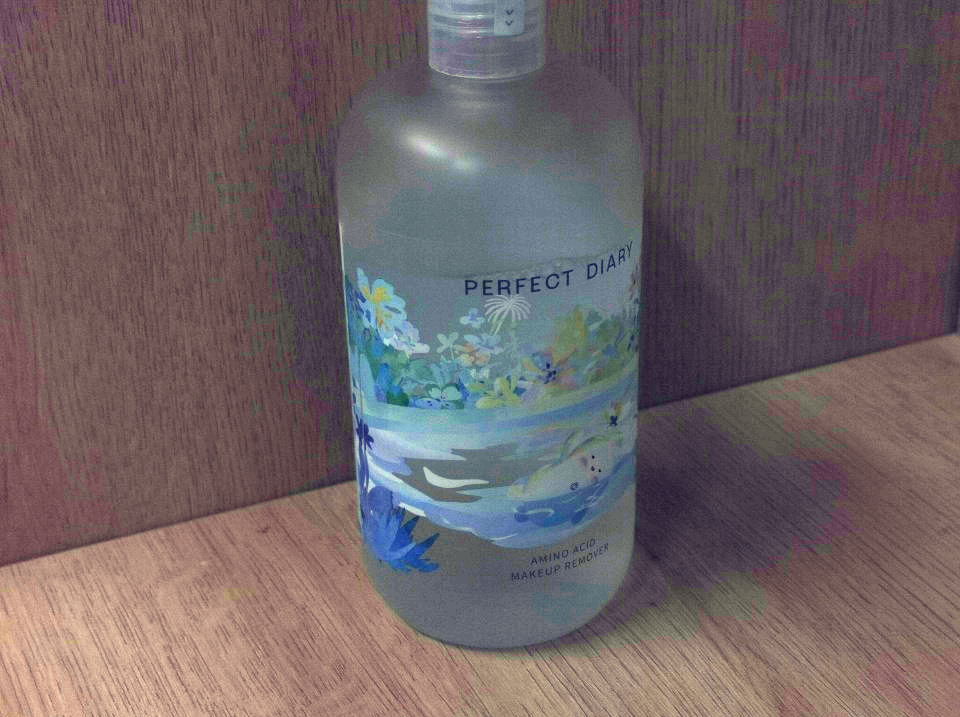}}%
\subfloat{\includegraphics[width=1.45cm, height=0.97cm]{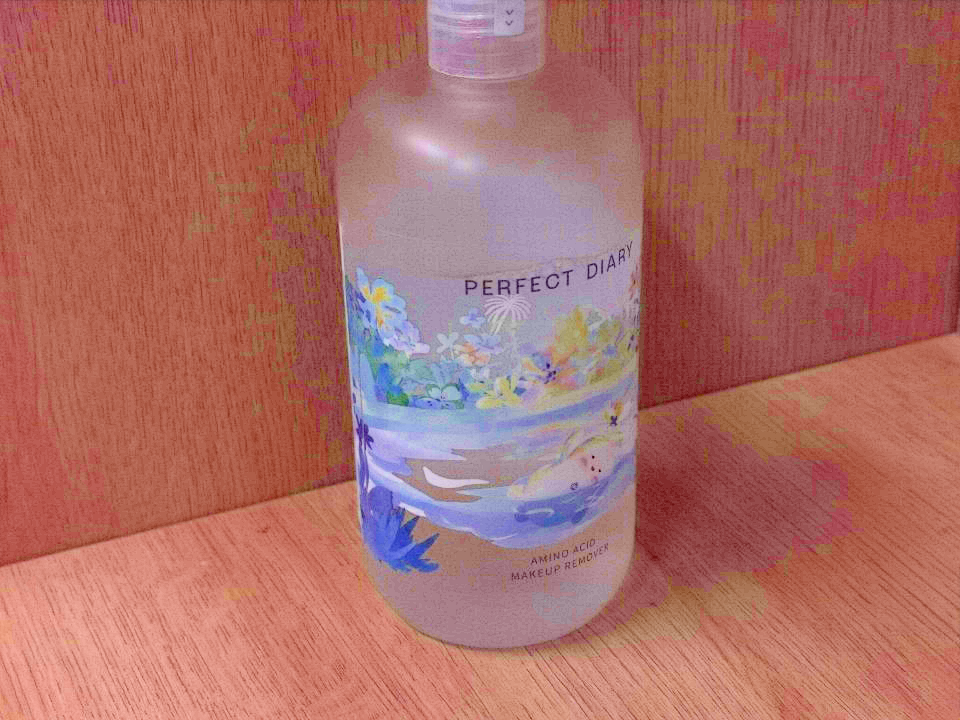}}%
\subfloat{\includegraphics[width=1.45cm, height=0.97cm]{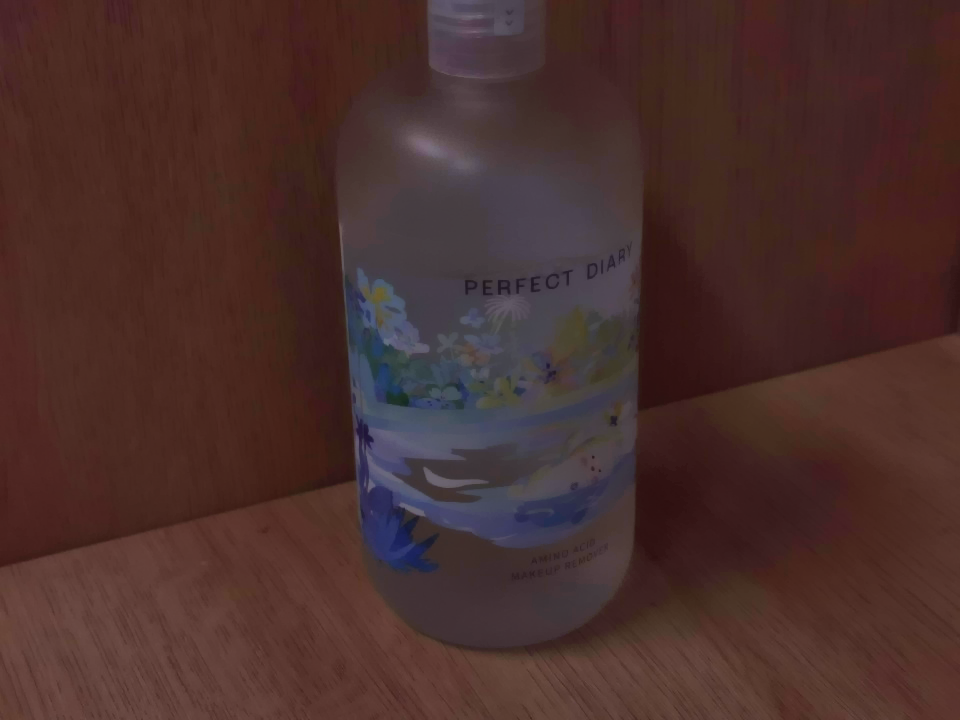}}%
\subfloat{\includegraphics[width=1.45cm, height=0.97cm]{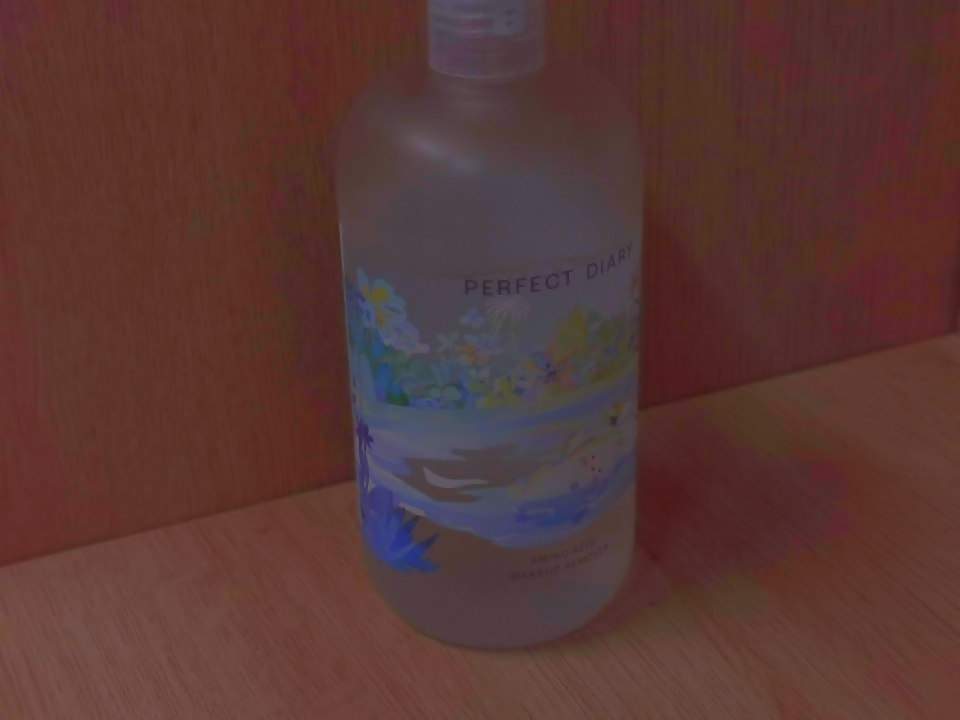}}%
\subfloat{\includegraphics[width=1.45cm, height=0.97cm]{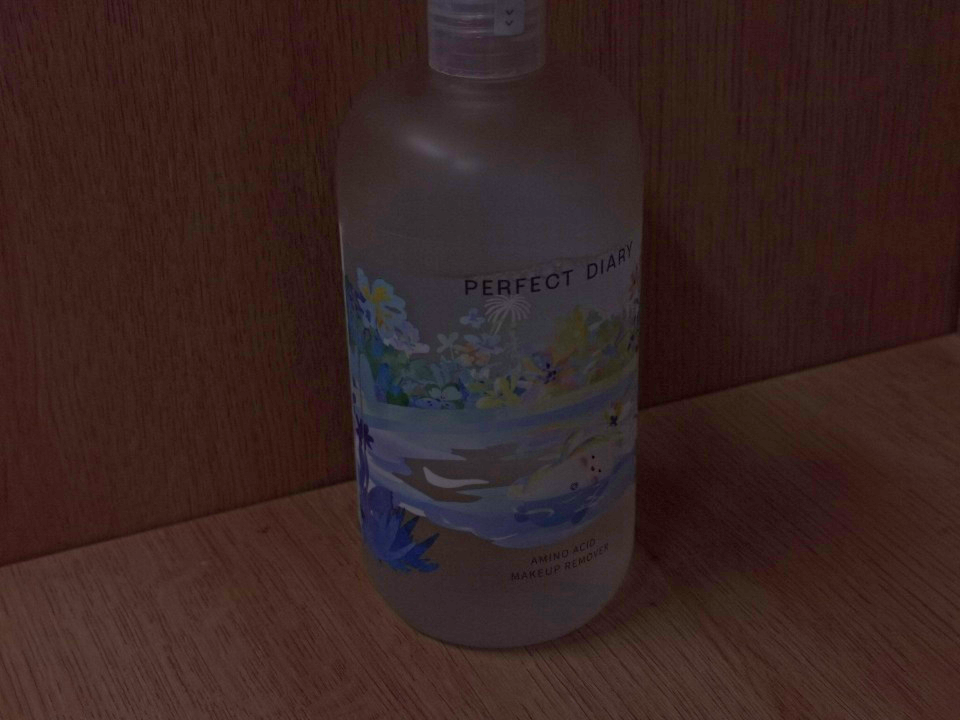}}%
\subfloat{\includegraphics[width=1.45cm, height=0.97cm]{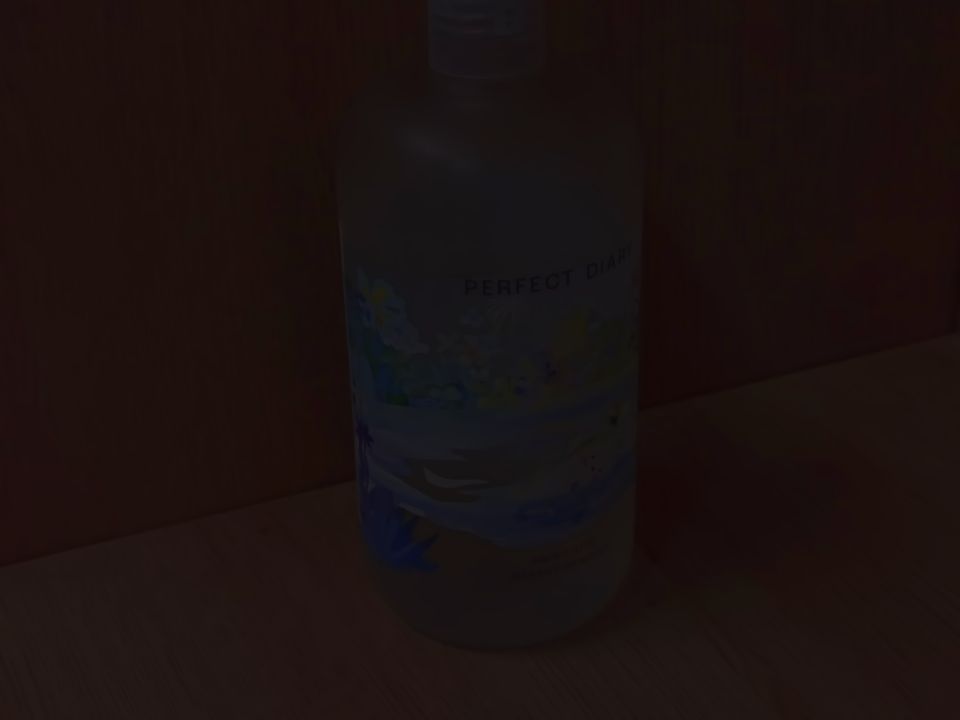}}%
\subfloat{\includegraphics[width=1.45cm, height=0.97cm]{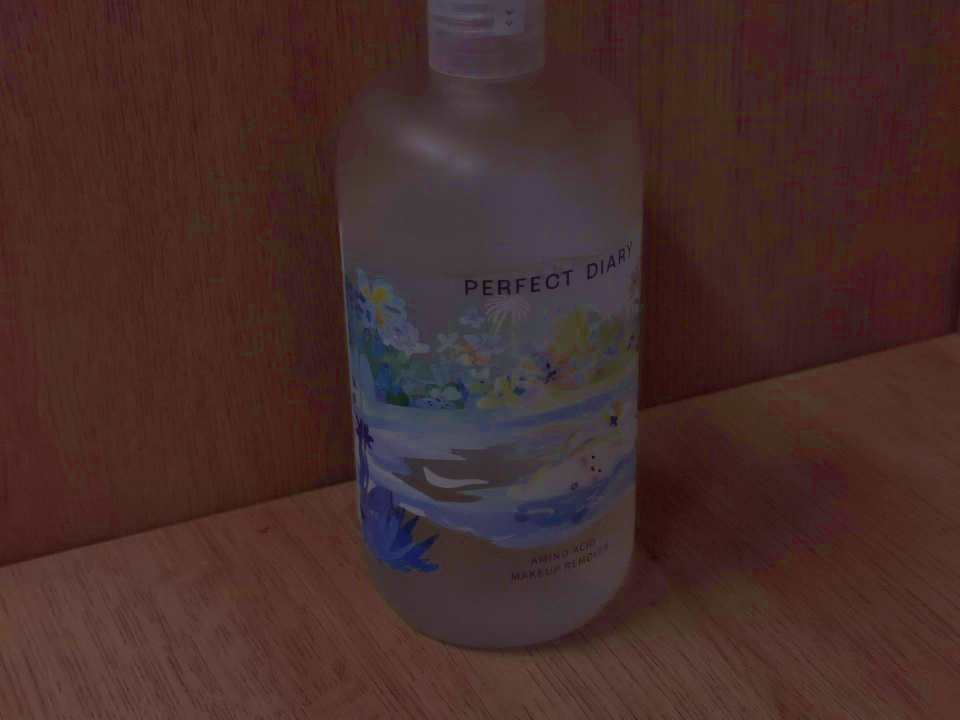}}%
\subfloat{\includegraphics[width=1.45cm, height=0.97cm]{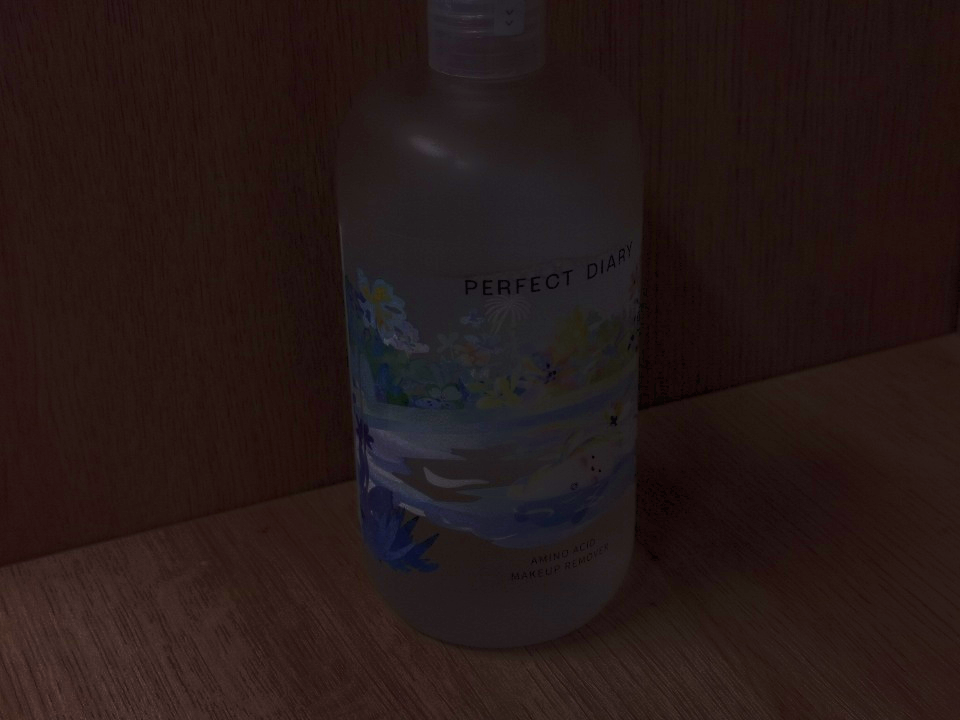}}\\
    \vspace{-1em}
\subfloat{\includegraphics[width=1.45cm, height=0.97cm]{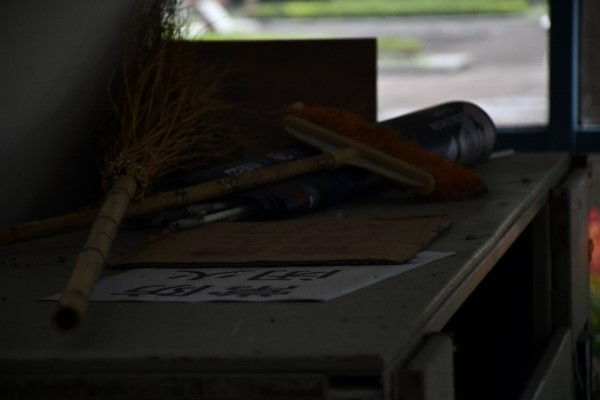}}%
\subfloat{\includegraphics[width=1.45cm, height=0.97cm]{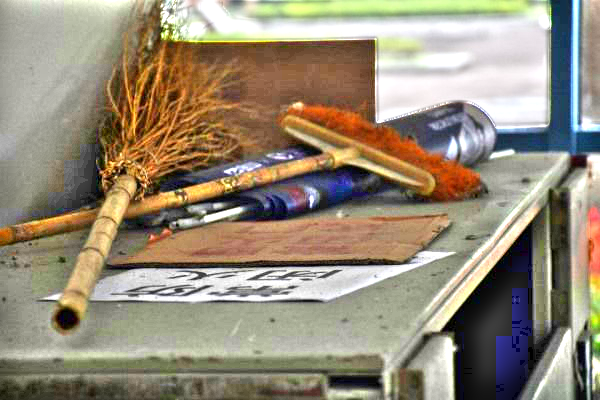}}%
\subfloat{\includegraphics[width=1.45cm, height=0.97cm]{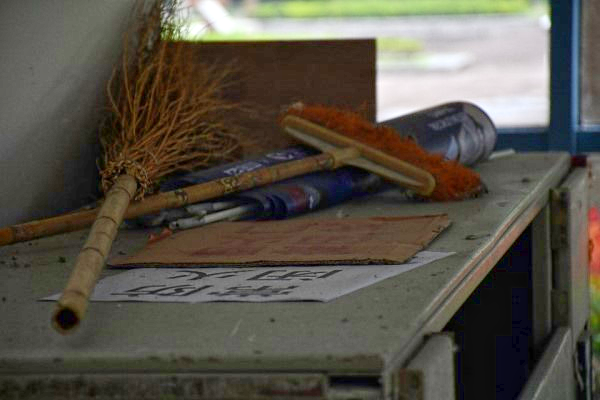}}%
\subfloat{\includegraphics[width=1.45cm, height=0.97cm]{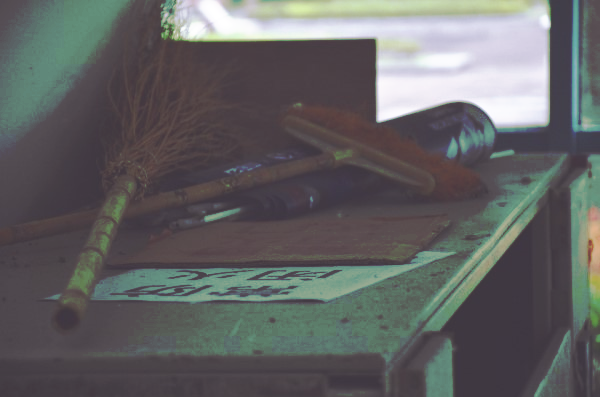}}%
\subfloat{\includegraphics[width=1.45cm, height=0.97cm]{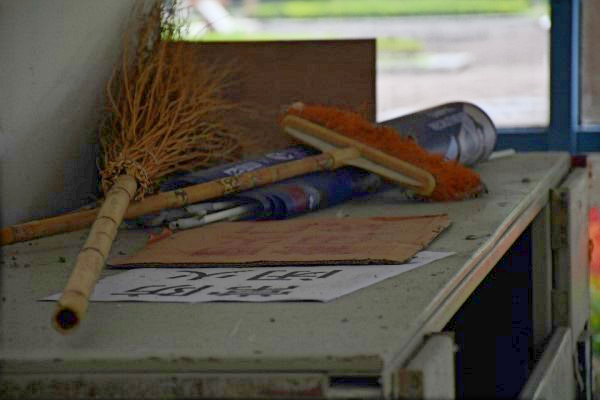}}%
\subfloat{\includegraphics[width=1.45cm, height=0.97cm]{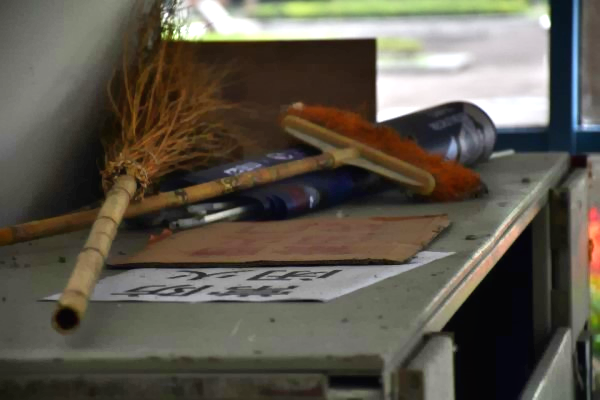}}%
\subfloat{\includegraphics[width=1.45cm, height=0.97cm]{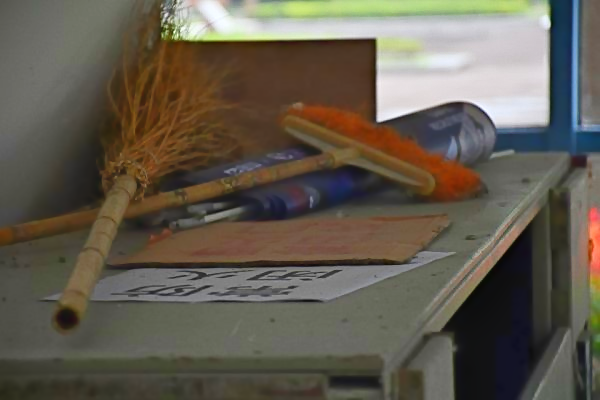}}%
\subfloat{\includegraphics[width=1.45cm, height=0.97cm]{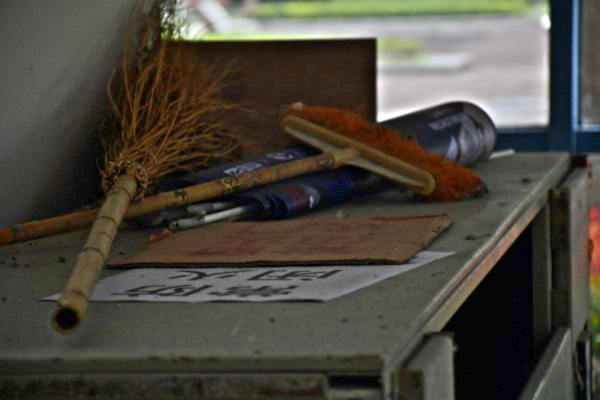}}%
\subfloat{\includegraphics[width=1.45cm, height=0.97cm]{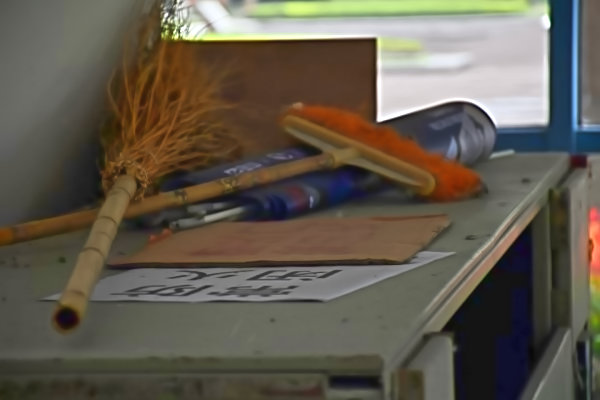}}%
\subfloat{\includegraphics[width=1.45cm, height=0.97cm]{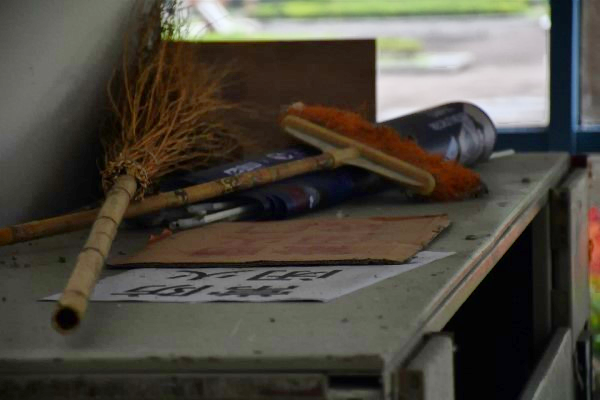}}%
\subfloat{\includegraphics[width=1.45cm, height=0.97cm]{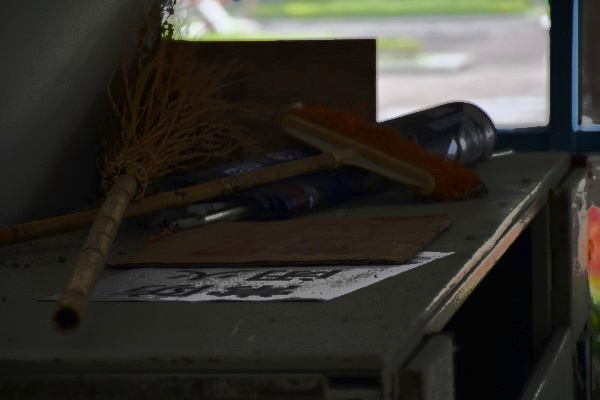}}\\
    \vspace{-1em}
\subfloat{\includegraphics[width=1.45cm, height=0.97cm]{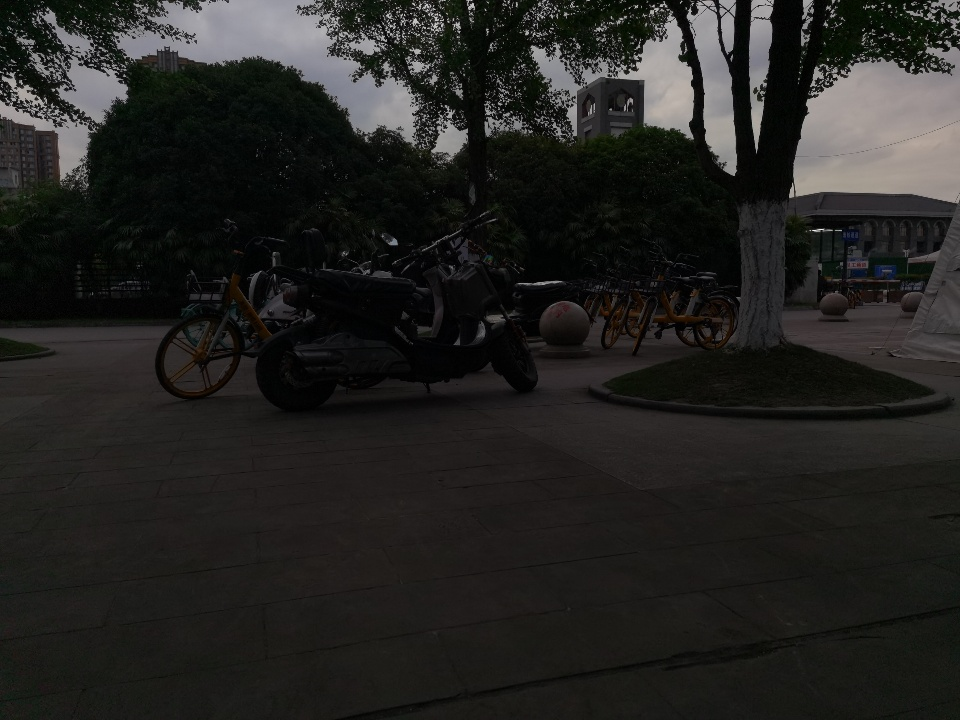}}%
\subfloat{\includegraphics[width=1.45cm, height=0.97cm]{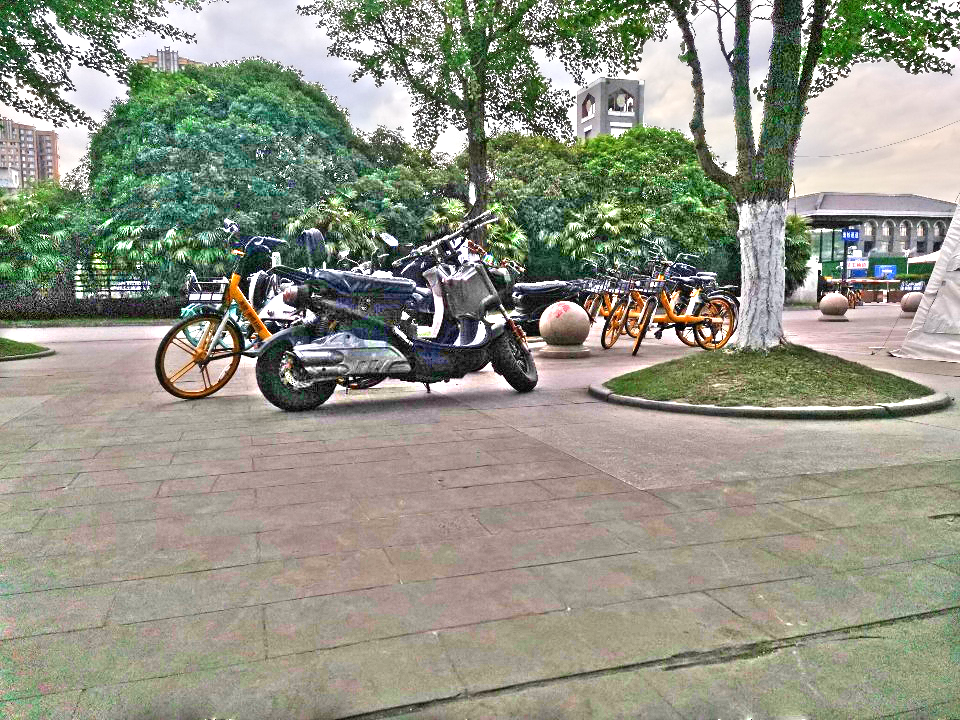}}%
\subfloat{\includegraphics[width=1.45cm, height=0.97cm]{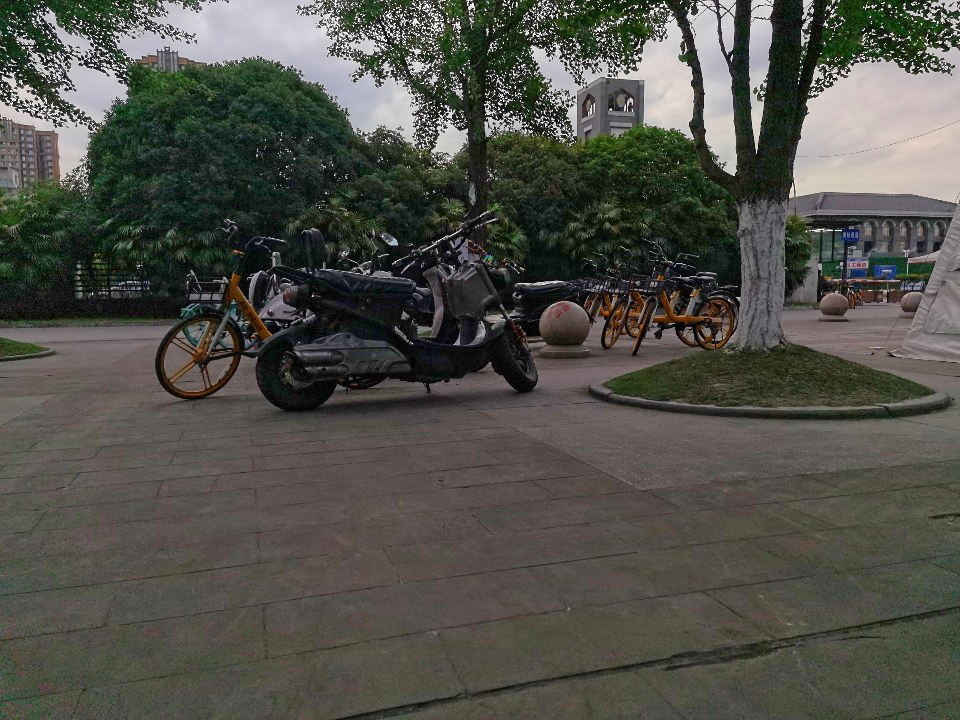}}%
\subfloat{\includegraphics[width=1.45cm, height=0.97cm]{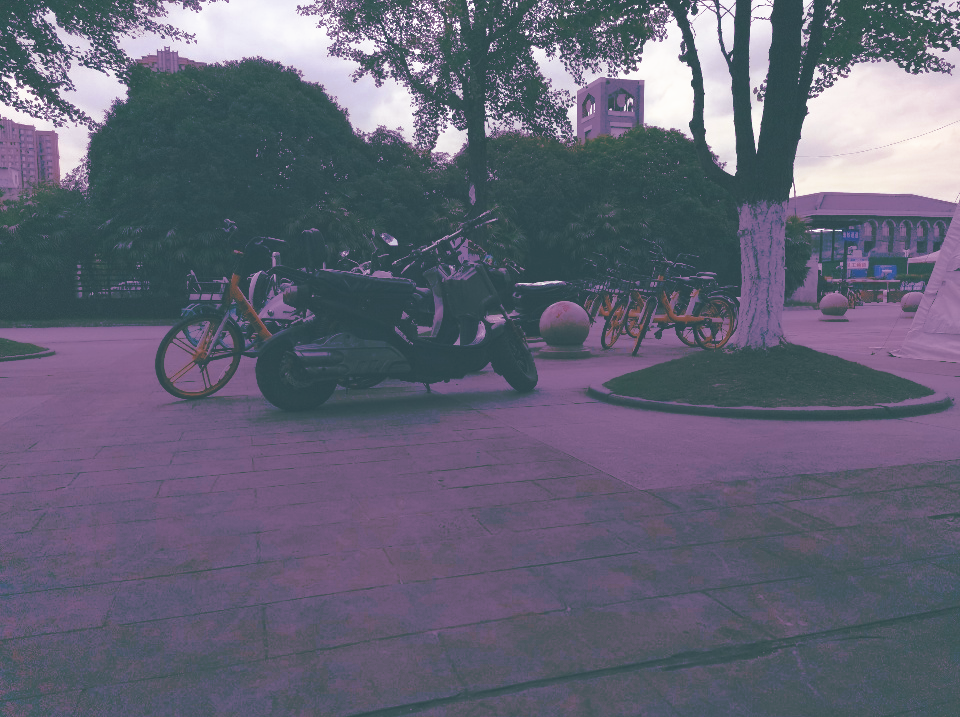}}%
\subfloat{\includegraphics[width=1.45cm, height=0.97cm]{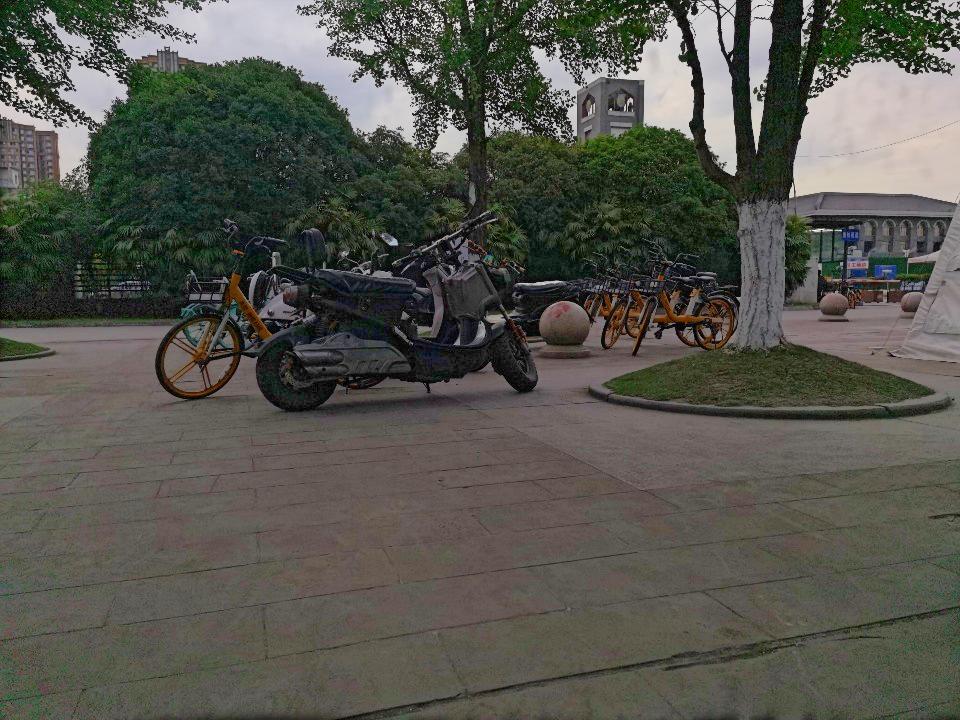}}%
\subfloat{\includegraphics[width=1.45cm, height=0.97cm]{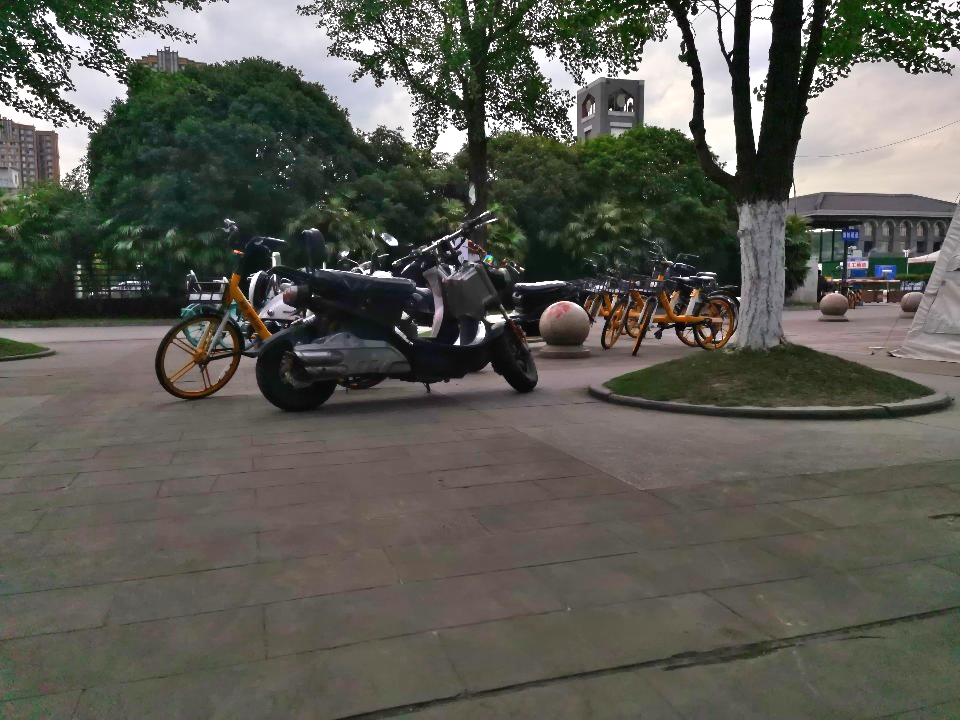}}%
\subfloat{\includegraphics[width=1.45cm, height=0.97cm]{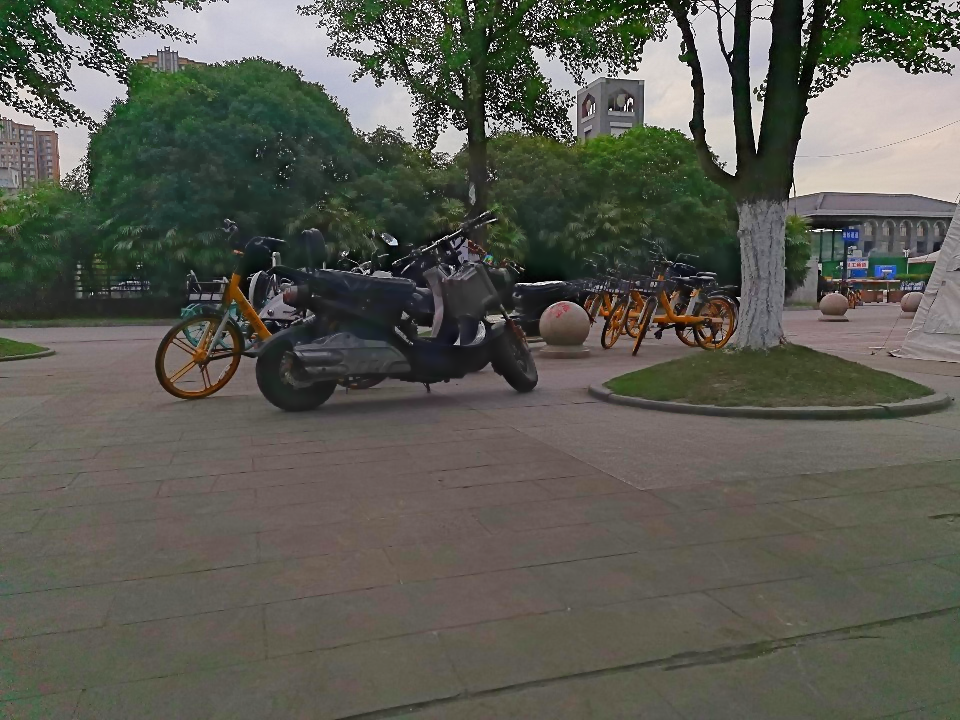}}%
\subfloat{\includegraphics[width=1.45cm, height=0.97cm]{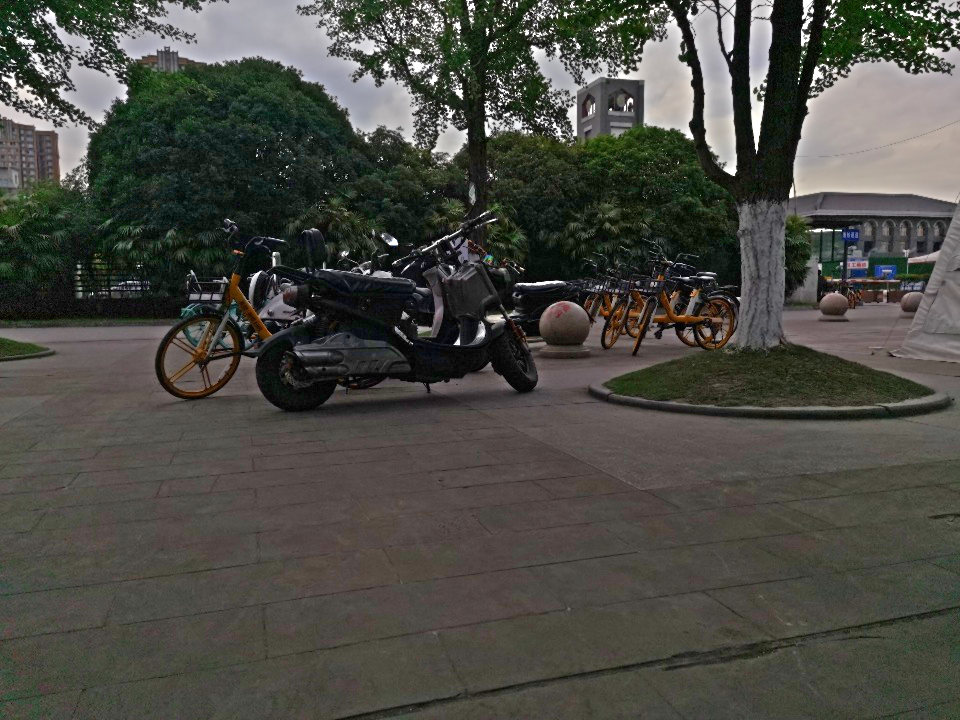}}%
\subfloat{\includegraphics[width=1.45cm, height=0.97cm]{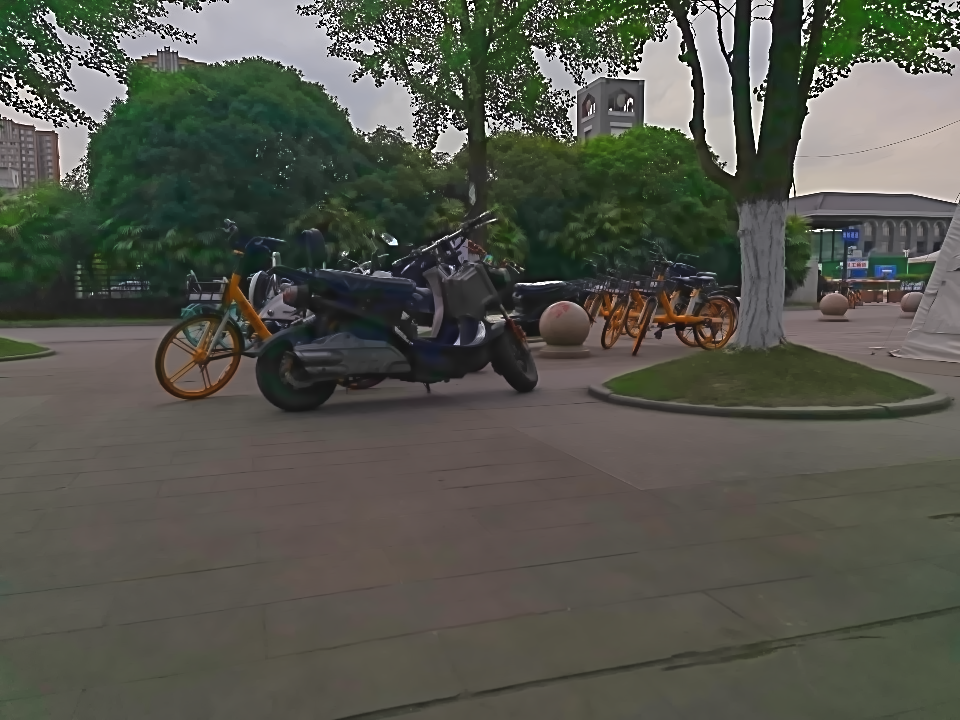}}%
\subfloat{\includegraphics[width=1.45cm, height=0.97cm]{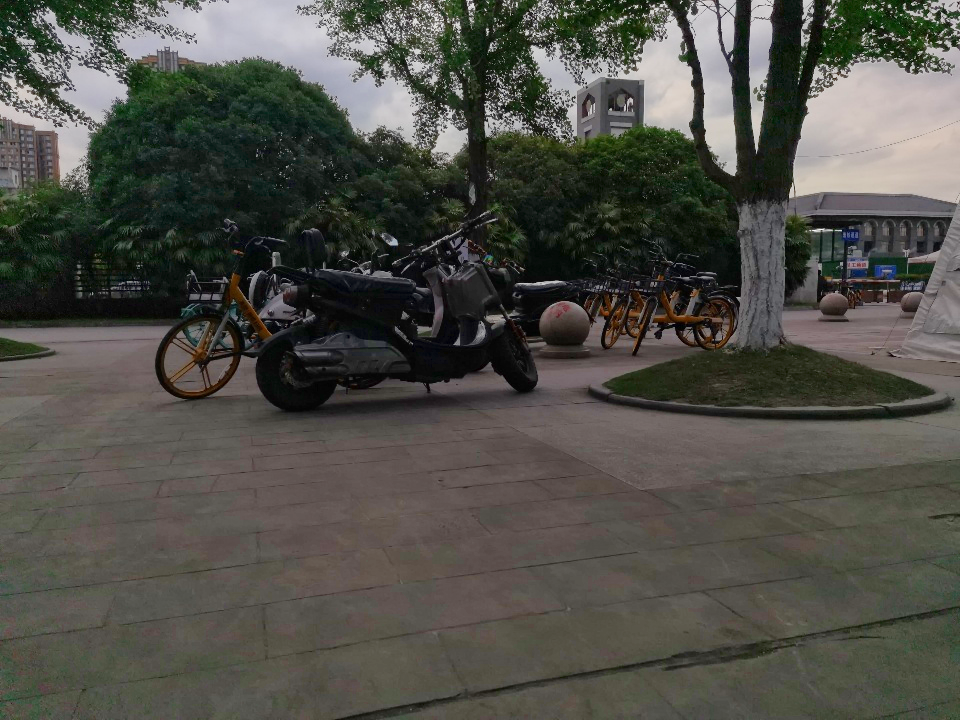}}%
\subfloat{\includegraphics[width=1.45cm, height=0.97cm]{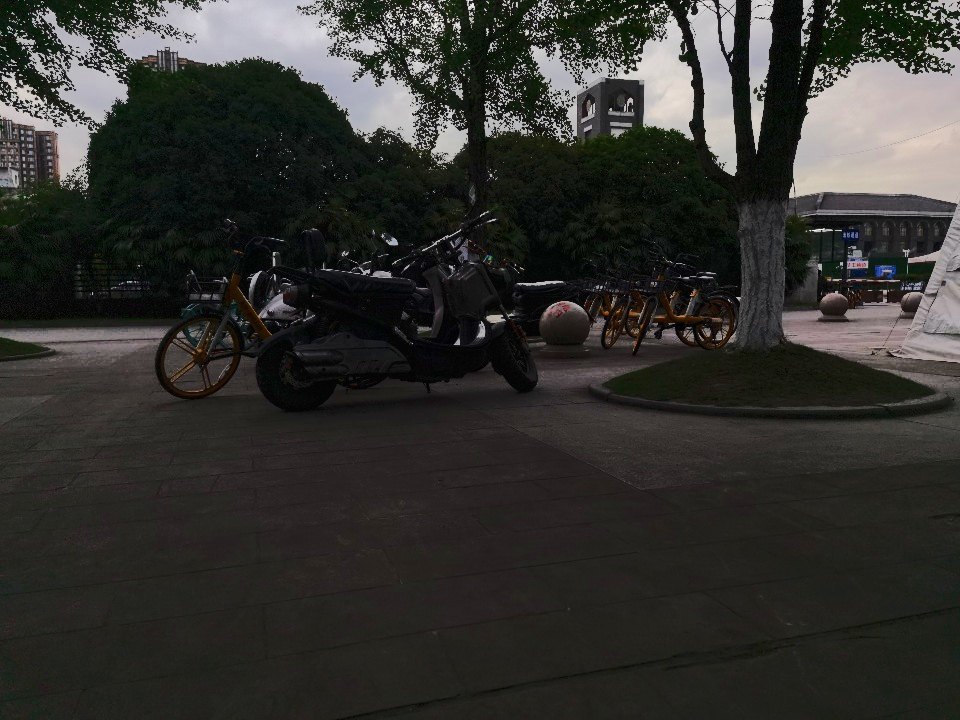}}\\
    \vspace{-1em}
 \subfloat{\includegraphics[width=1.45cm, height=0.97cm]{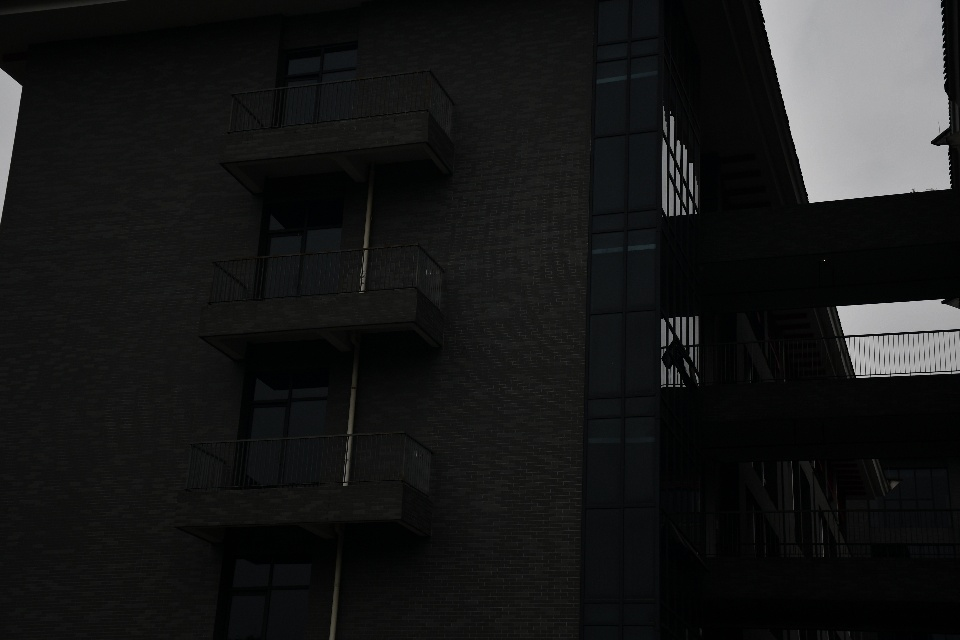}}%
\subfloat{\includegraphics[width=1.45cm, height=0.97cm]{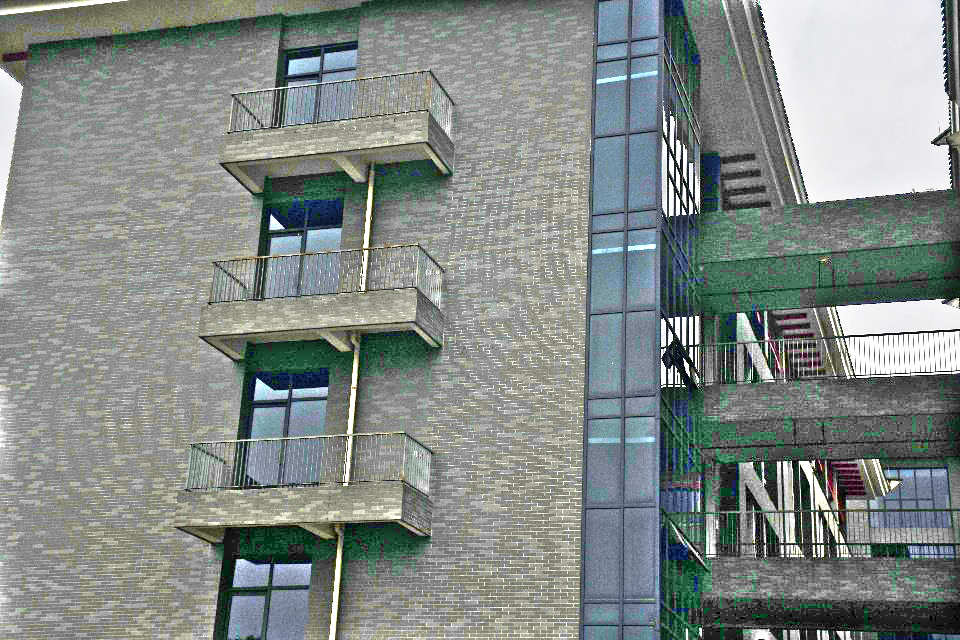}}%
\subfloat{\includegraphics[width=1.45cm, height=0.97cm]{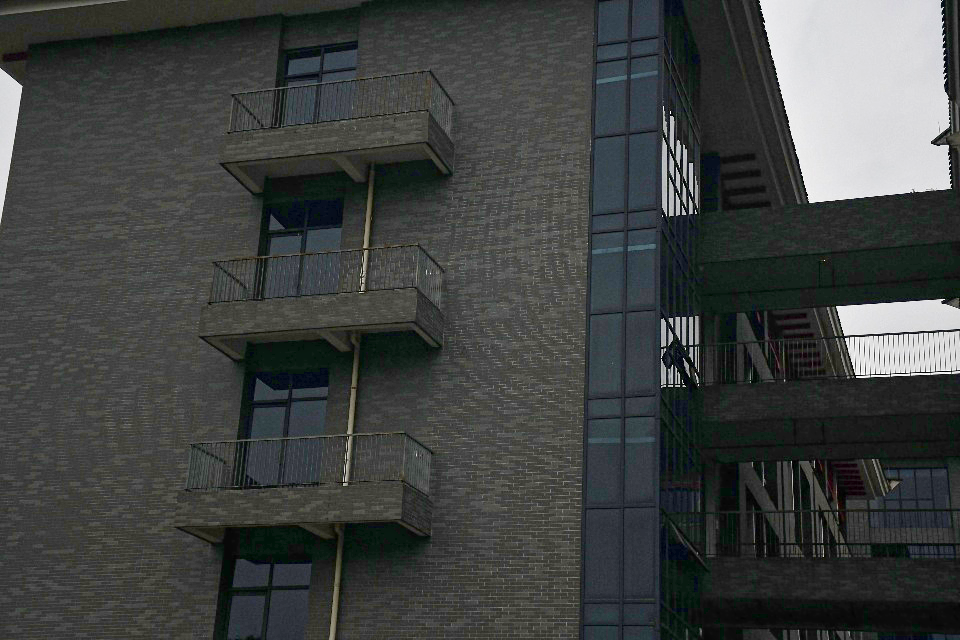}}%
\subfloat{\includegraphics[width=1.45cm, height=0.97cm]{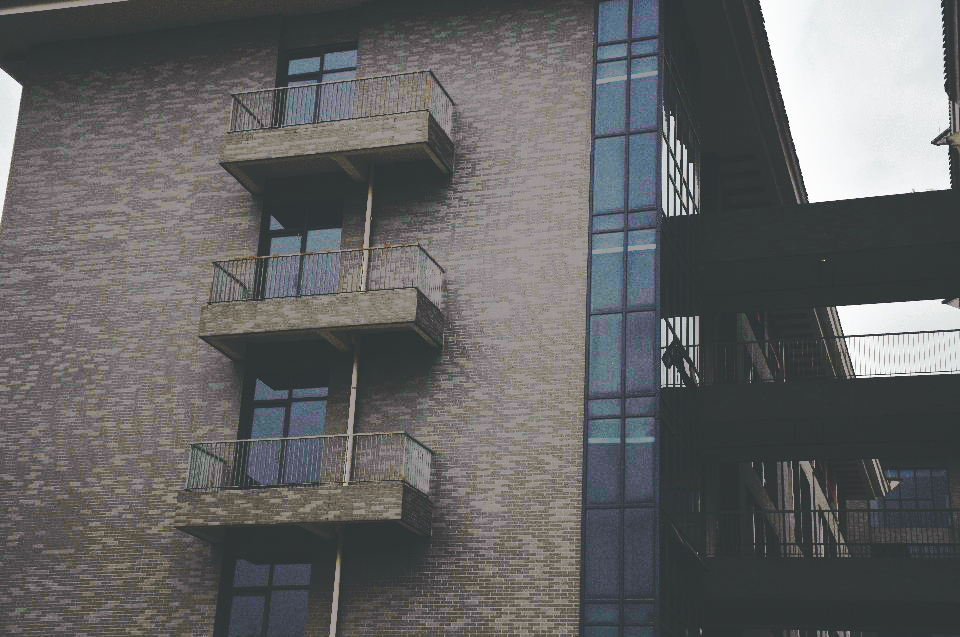}}%
\subfloat{\includegraphics[width=1.45cm, height=0.97cm]{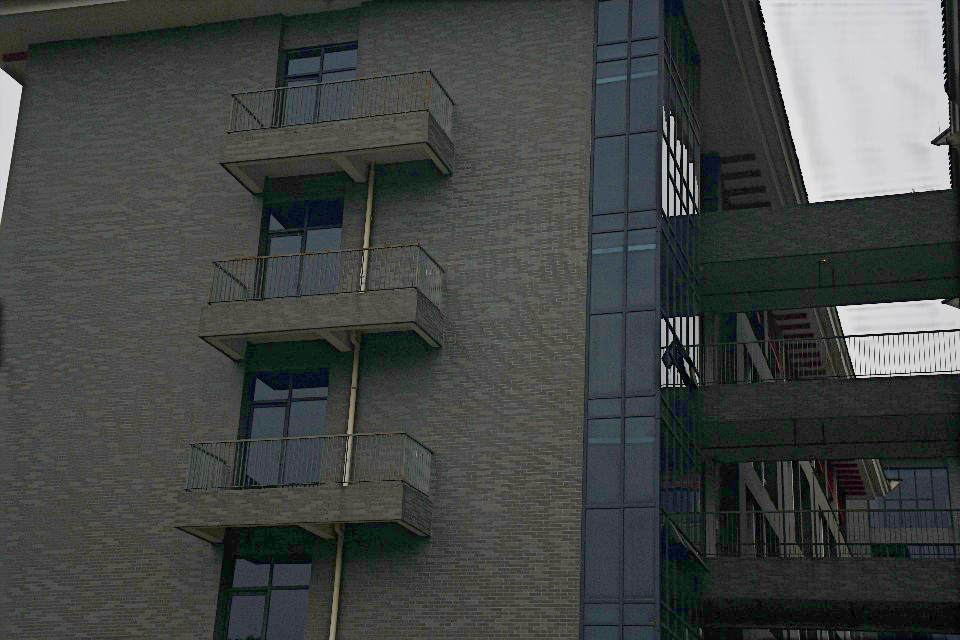}}%
\subfloat{\includegraphics[width=1.45cm, height=0.97cm]{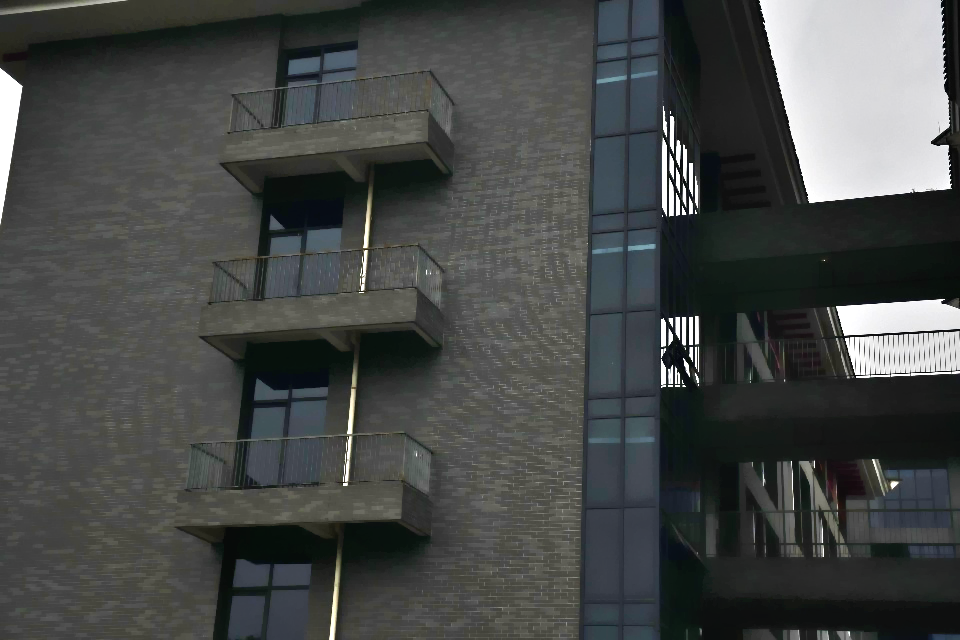}}%
\subfloat{\includegraphics[width=1.45cm, height=0.97cm]{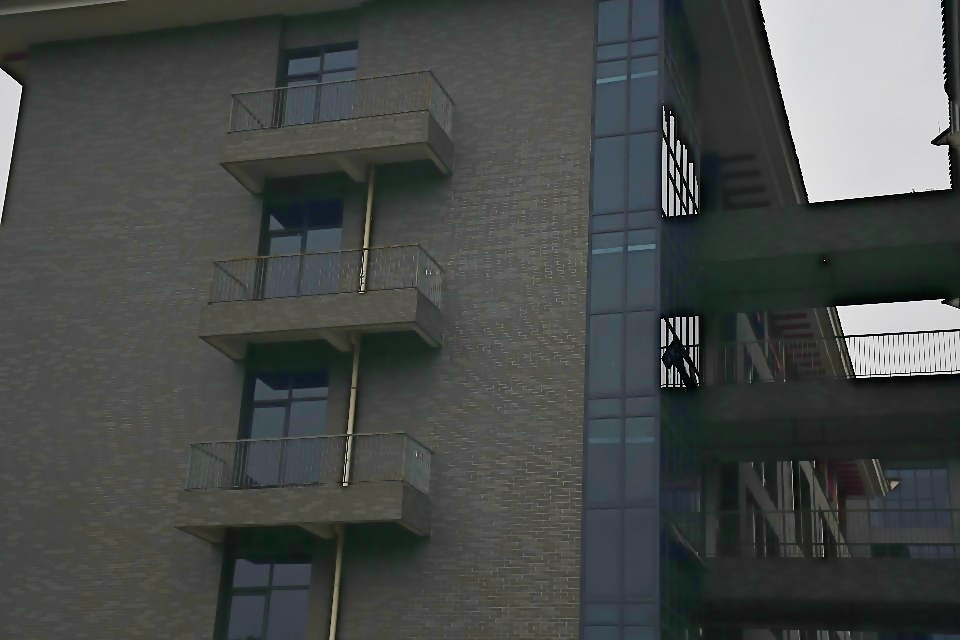}}%
\subfloat{\includegraphics[width=1.45cm, height=0.97cm]{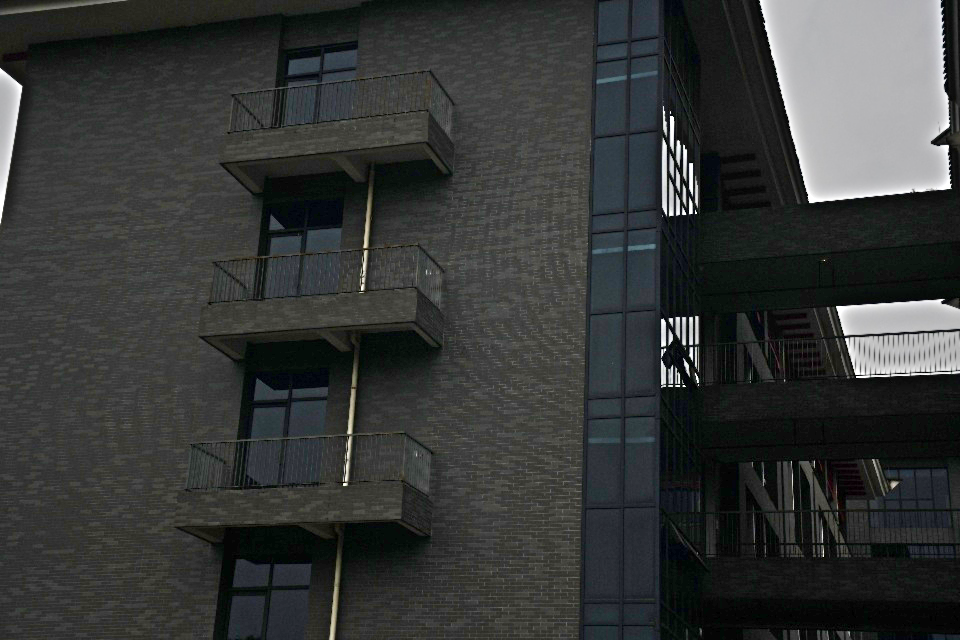}}%
\subfloat{\includegraphics[width=1.45cm, height=0.97cm]{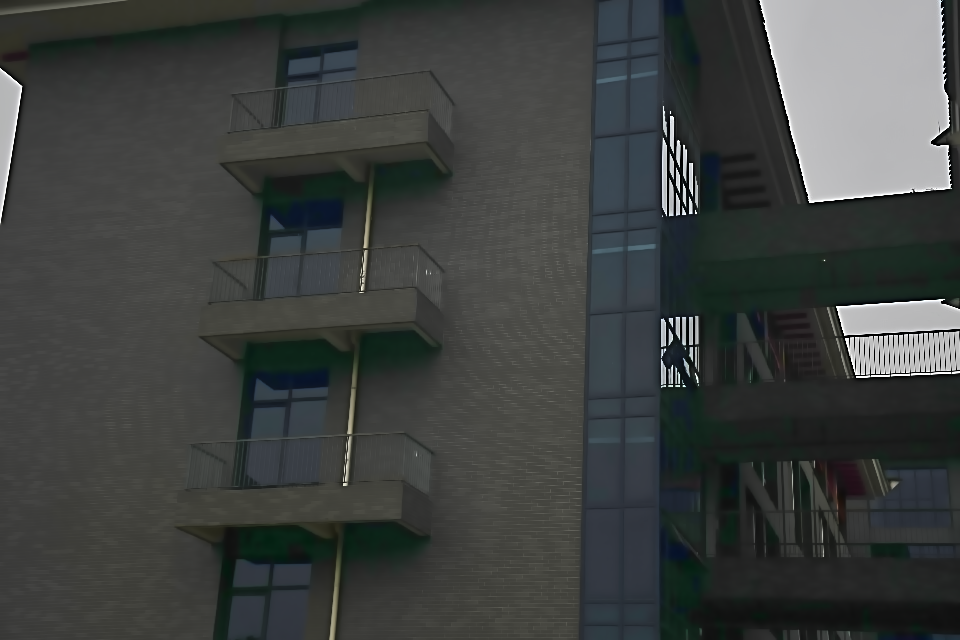}}%
\subfloat{\includegraphics[width=1.45cm, height=0.97cm]{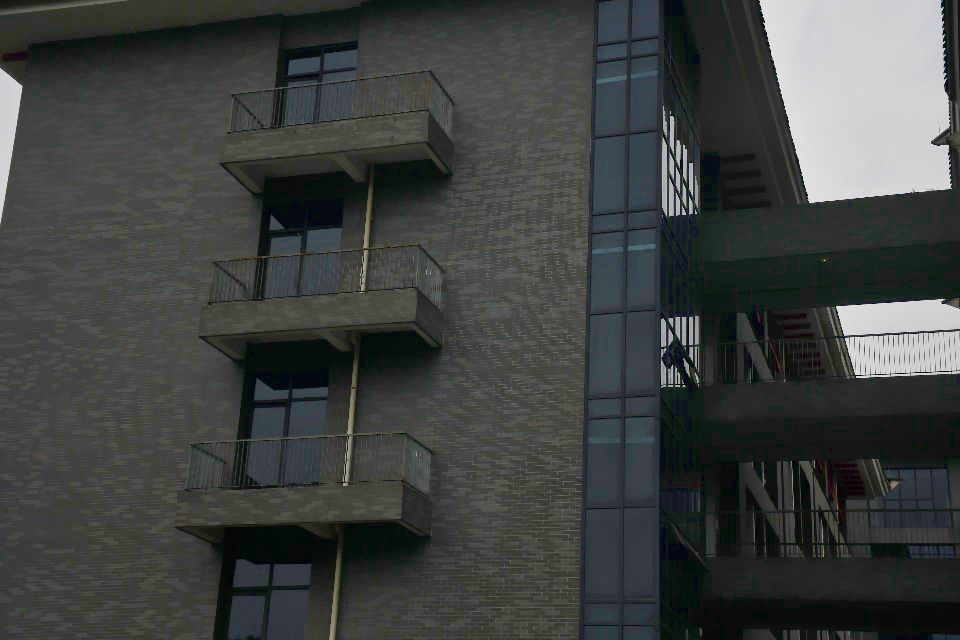}}%
\subfloat{\includegraphics[width=1.45cm, height=0.97cm]{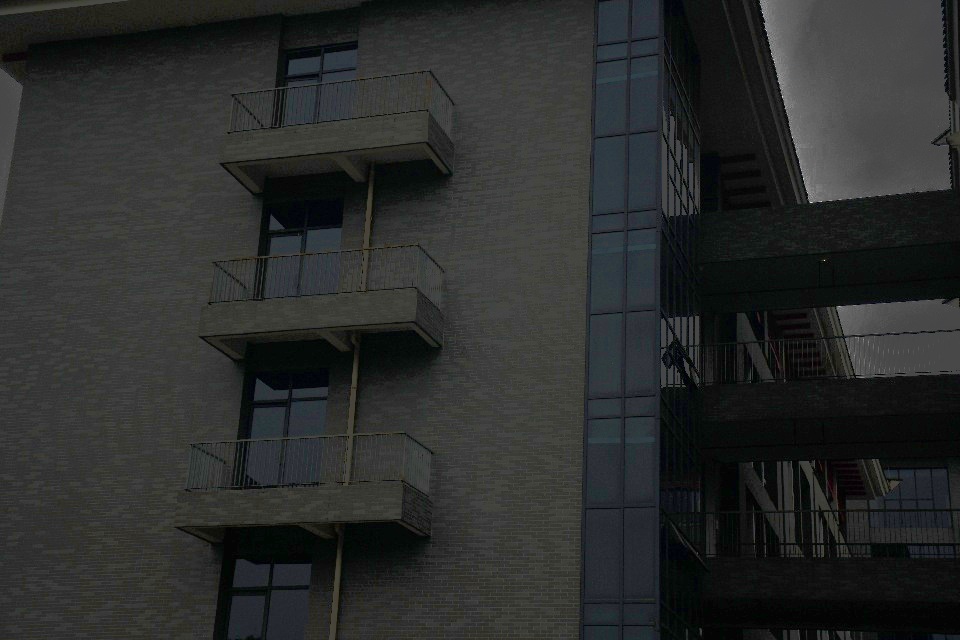}}\\
    \vspace{-1em}
\subfloat{\includegraphics[width=1.45cm, height=0.97cm]{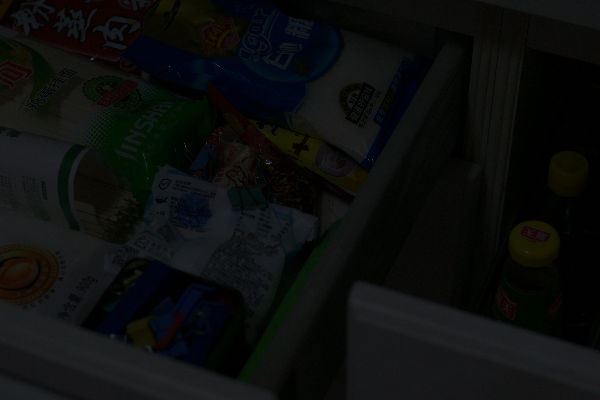}}%
\subfloat{\includegraphics[width=1.45cm, height=0.97cm]{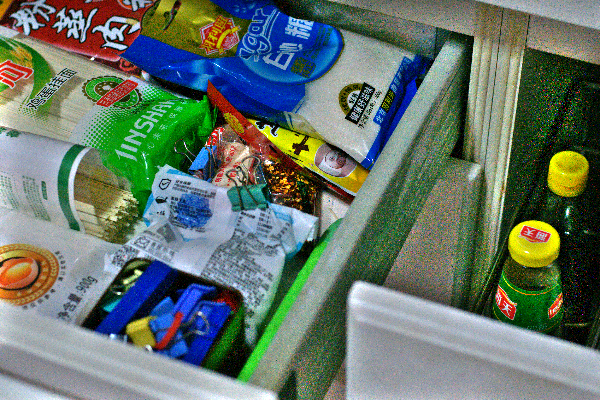}}%
\subfloat{\includegraphics[width=1.45cm, height=0.97cm]{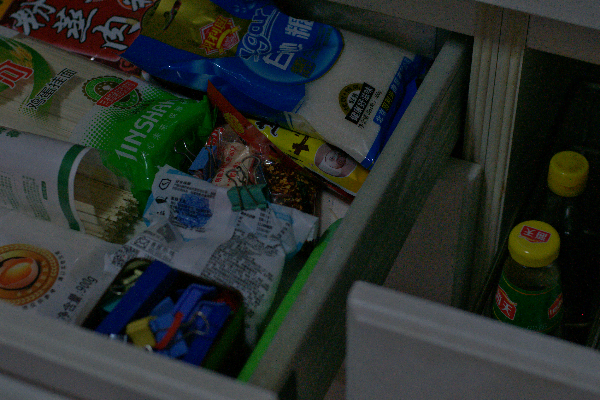}}%
\subfloat{\includegraphics[width=1.45cm, height=0.97cm]{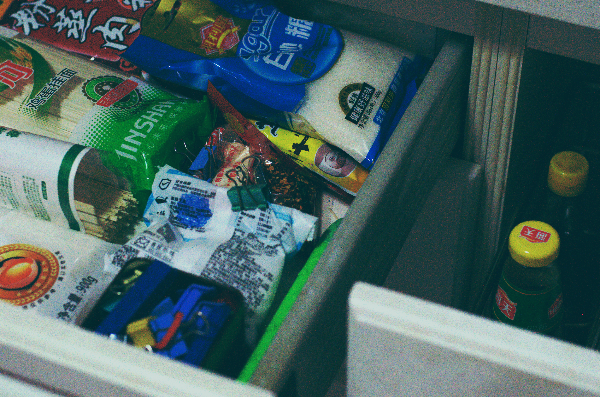}}%
\subfloat{\includegraphics[width=1.45cm, height=0.97cm]{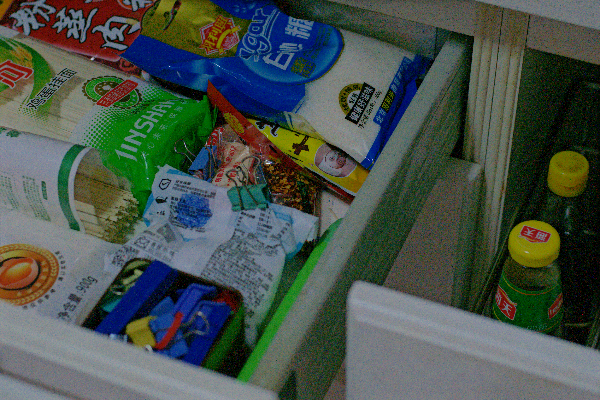}}%
\subfloat{\includegraphics[width=1.45cm, height=0.97cm]{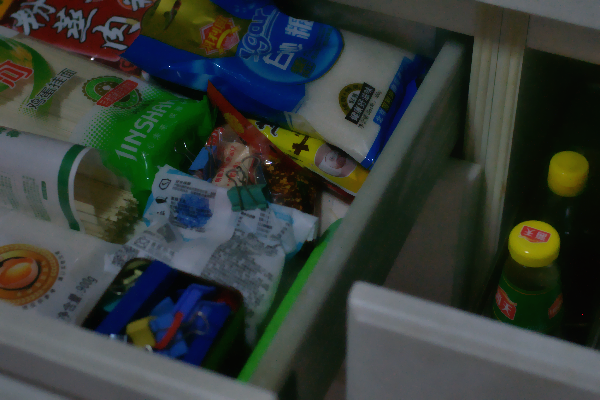}}%
\subfloat{\includegraphics[width=1.45cm, height=0.97cm]{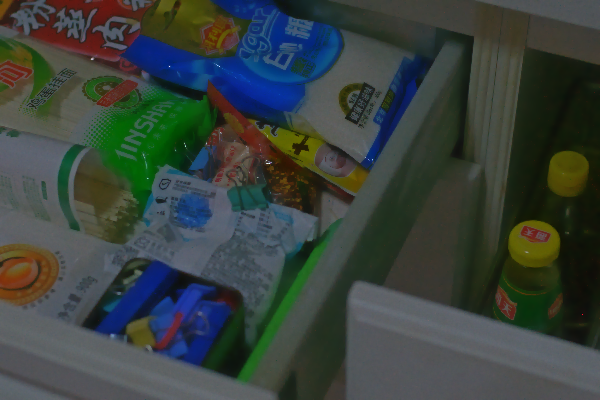}}%
\subfloat{\includegraphics[width=1.45cm, height=0.97cm]{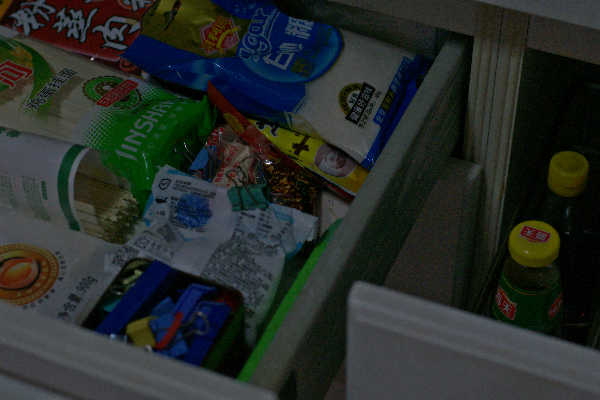}}%
\subfloat{\includegraphics[width=1.45cm, height=0.97cm]{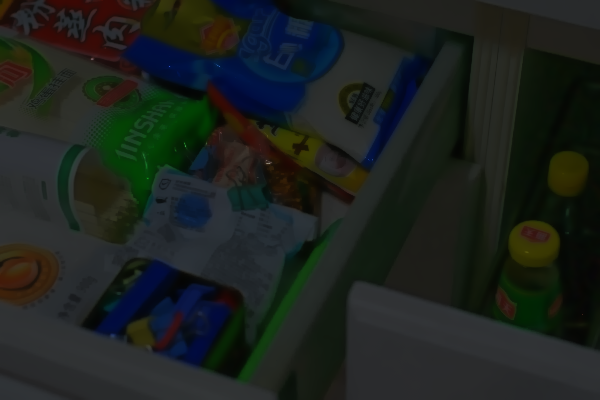}}%
\subfloat{\includegraphics[width=1.45cm, height=0.97cm]{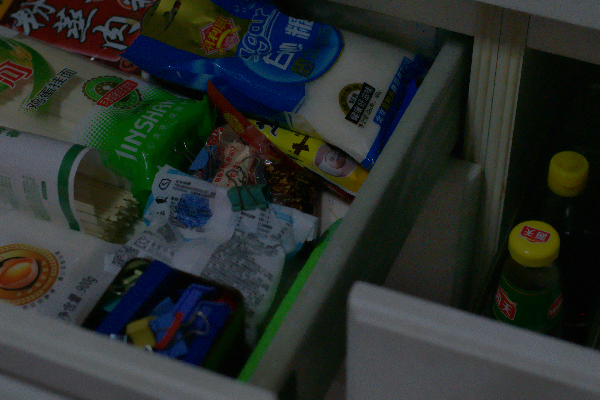}}%
\subfloat{\includegraphics[width=1.45cm, height=0.97cm]{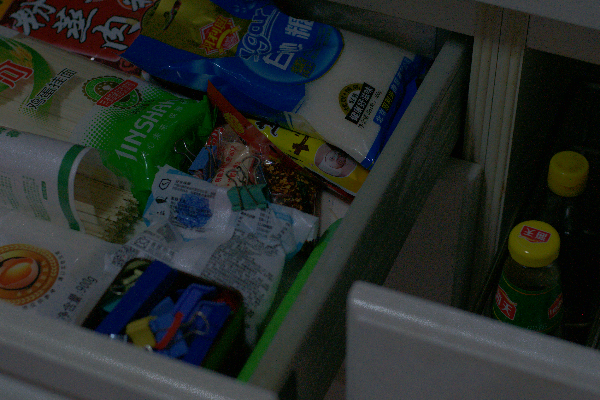}}\\

(a)\hspace{1.1cm}(b)\hspace{1.1cm}(c)\hspace{1.1cm}(d)\hspace{1.1cm}(e)\hspace{1.1cm}(f)\hspace{1.1cm}(g)\hspace{1.1cm}(h)\hspace{1.1cm}(i)\hspace{1.1cm}(j)\hspace{1.1cm}(k)

\vspace{-0.2em}
    \caption{Enhancement results of ten algorithms on LOL and LSRW datasets. (a) low-light image. (b) LIME; (c) STAR; (d) AHPCE; (e) WVM; (f) SSD; (g) RRM; (h) BRAIN; (i) LR3M; (j) FSTAR; (k) proposed.}
    \label{fig:grid}
\end{figure*}

\begin{figure*}[htbp]
    \centering
    \subfloat{\includegraphics[width=2.9cm, height=1.8cm]{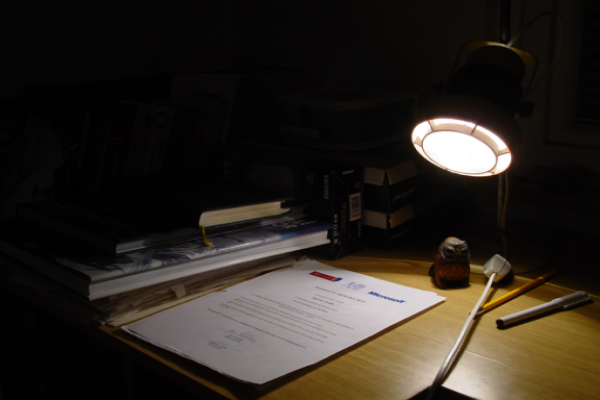}}%
    \subfloat{\includegraphics[width=2.9cm, height=1.8cm]{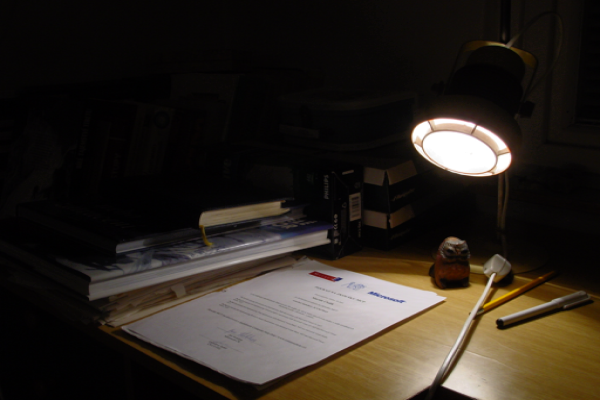}}%
    \subfloat{\includegraphics[width=2.9cm, height=1.8cm]{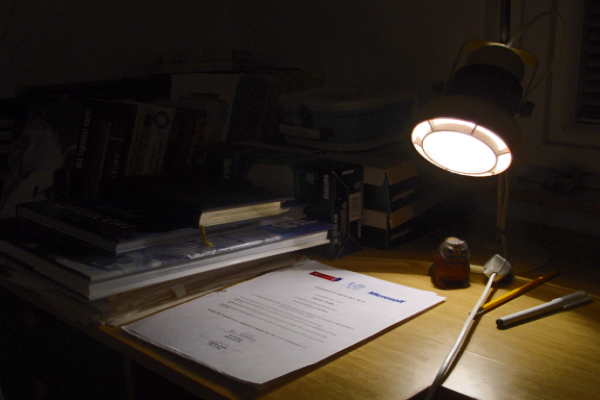}}%
    \subfloat{\includegraphics[width=2.9cm, height=1.8cm]{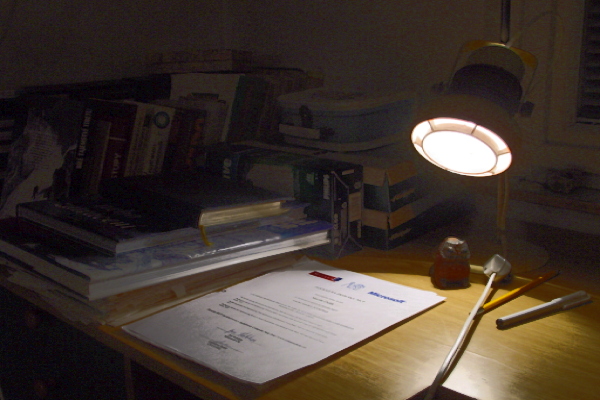}}%
    \subfloat{\includegraphics[width=2.9cm, height=1.8cm]{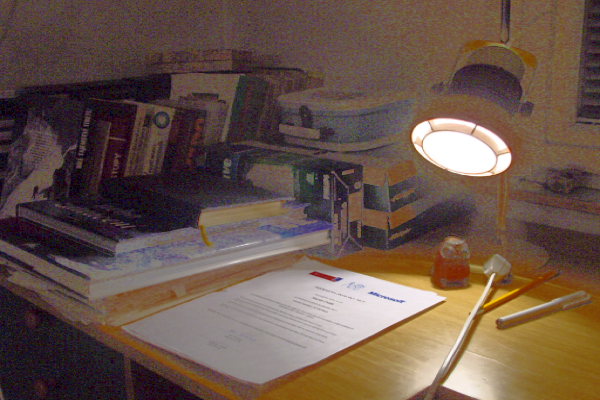}}%
    \subfloat{\includegraphics[width=2.9cm, height=1.8cm]{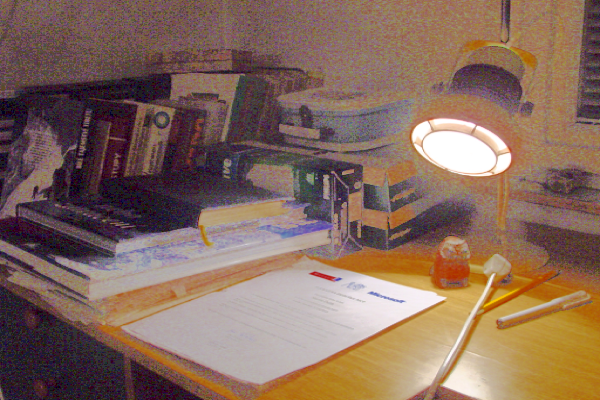}}\\
    \vspace{-1em}
    \subfloat{\includegraphics[width=2.9cm, height=1.8cm]{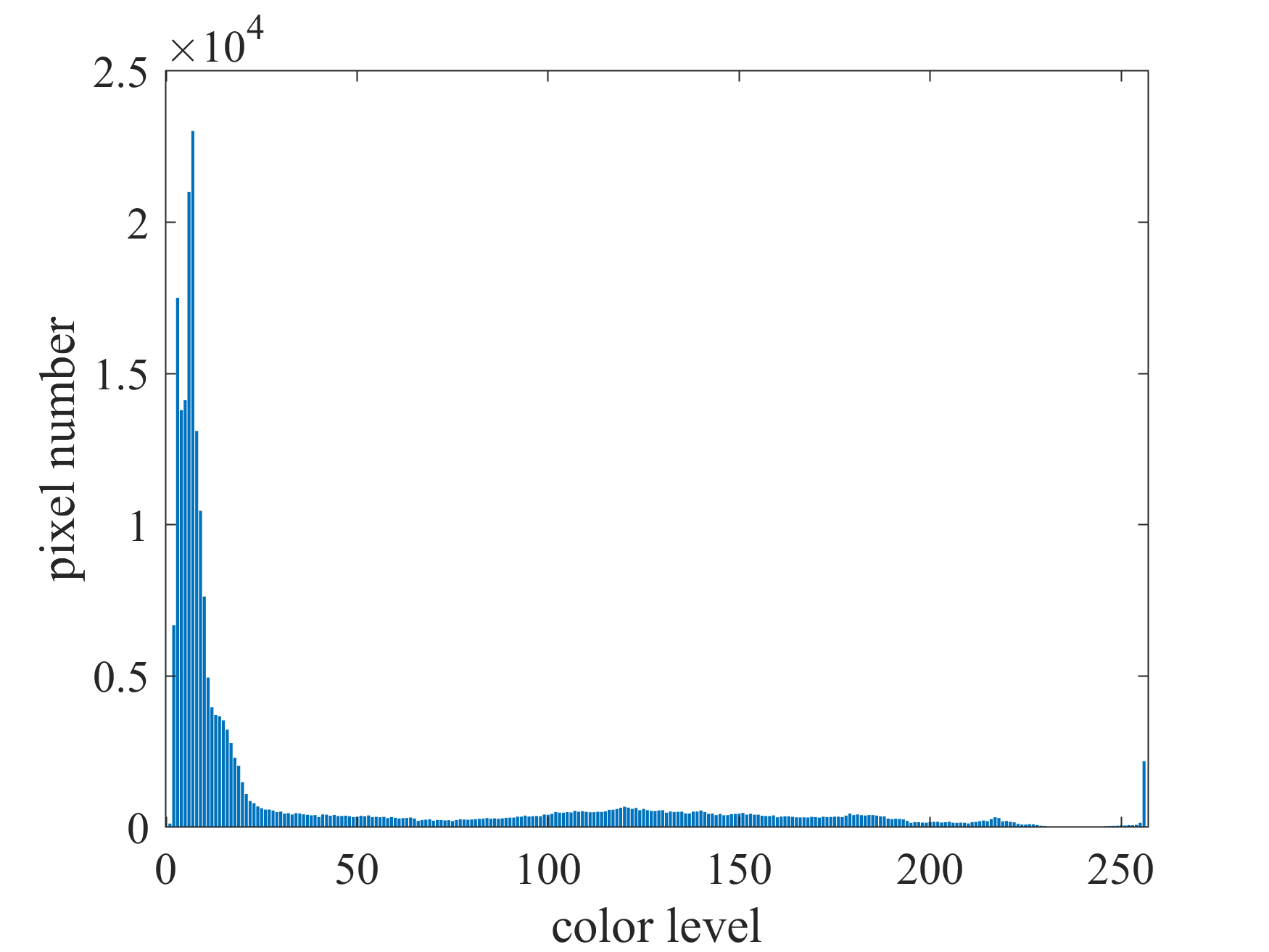}}%
    \subfloat{\includegraphics[width=2.9cm, height=1.8cm]{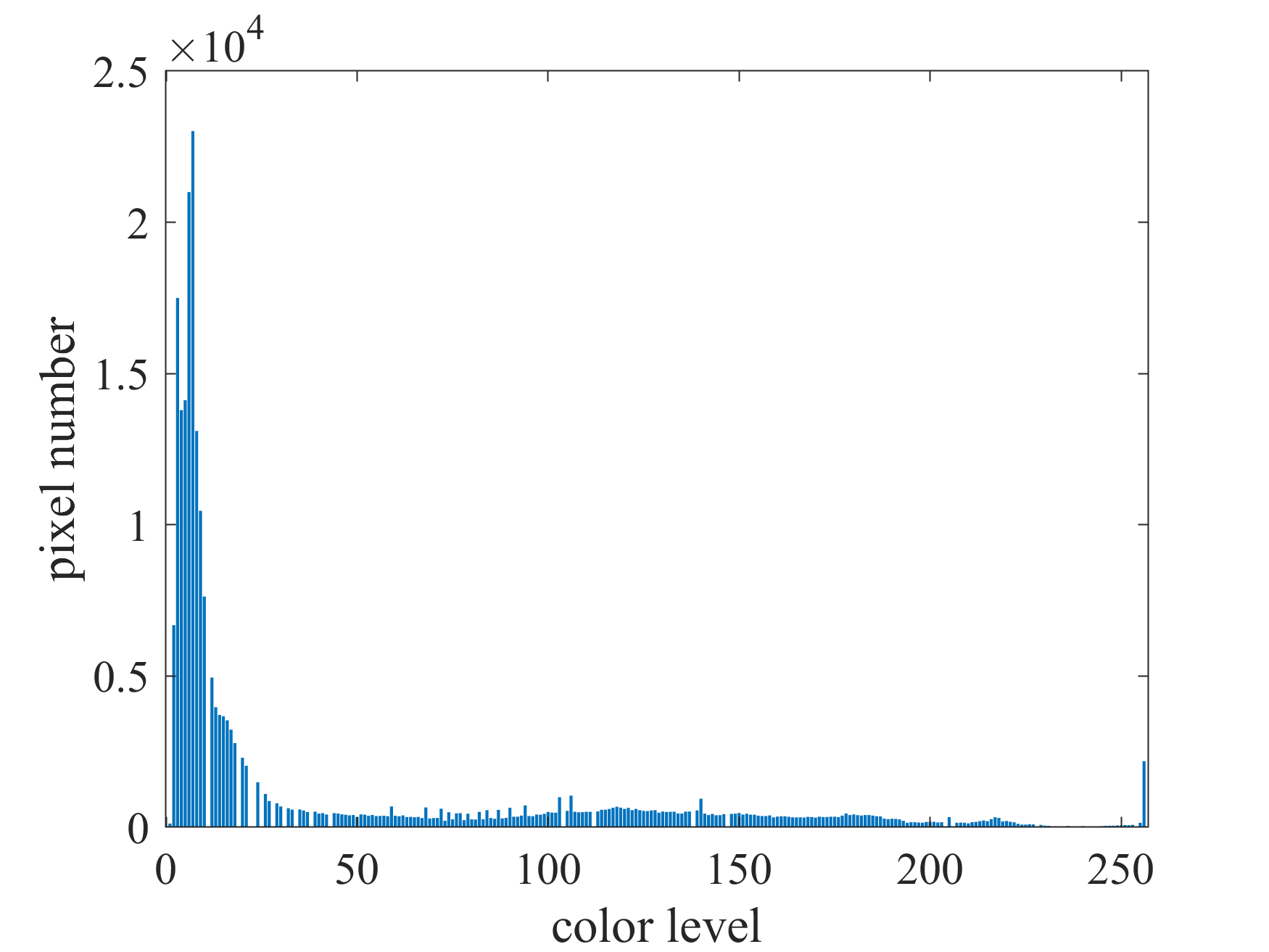}}%
    \subfloat{\includegraphics[width=2.9cm, height=1.8cm]{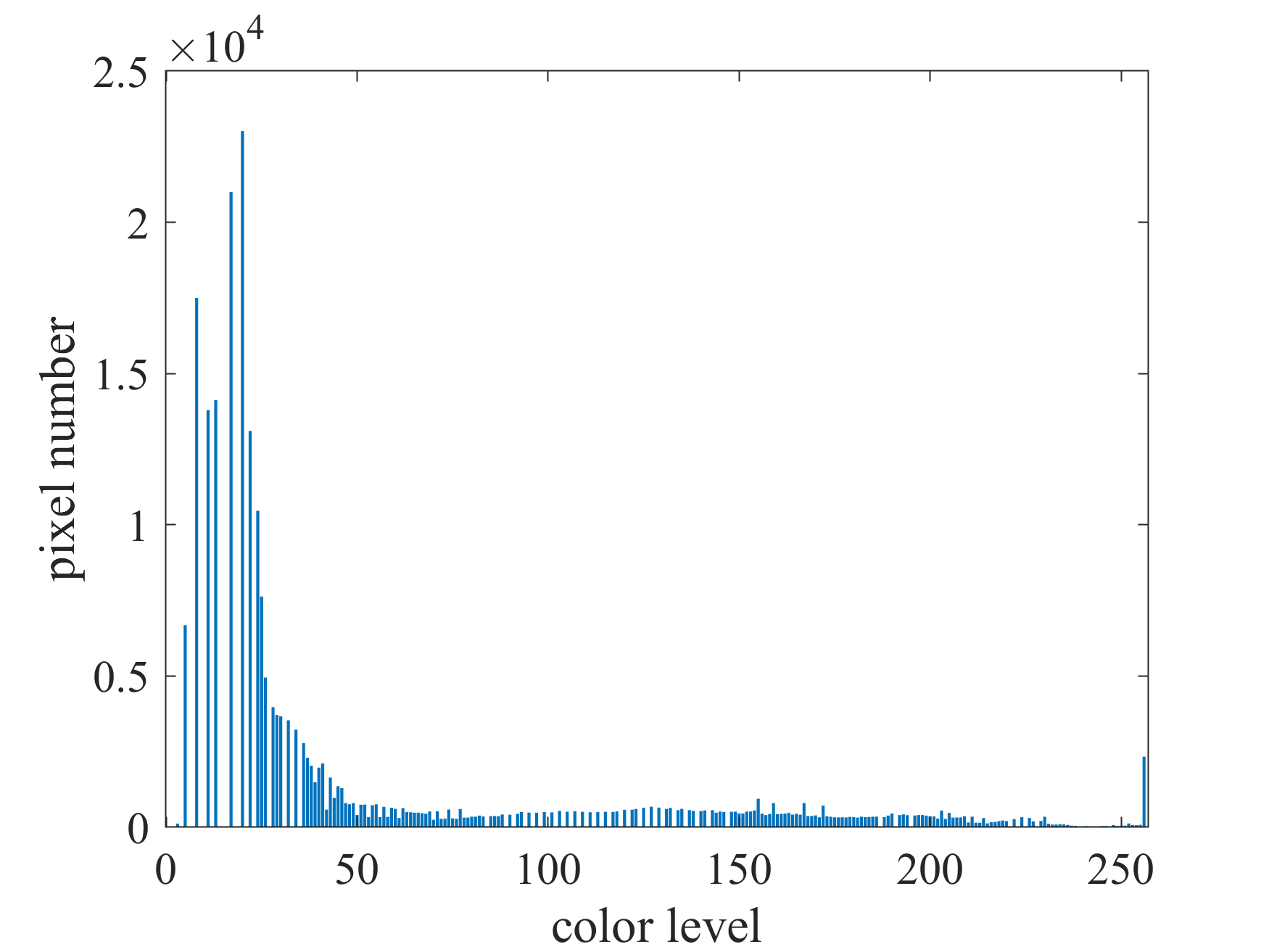}}%
    \subfloat{\includegraphics[width=2.9cm, height=1.8cm]{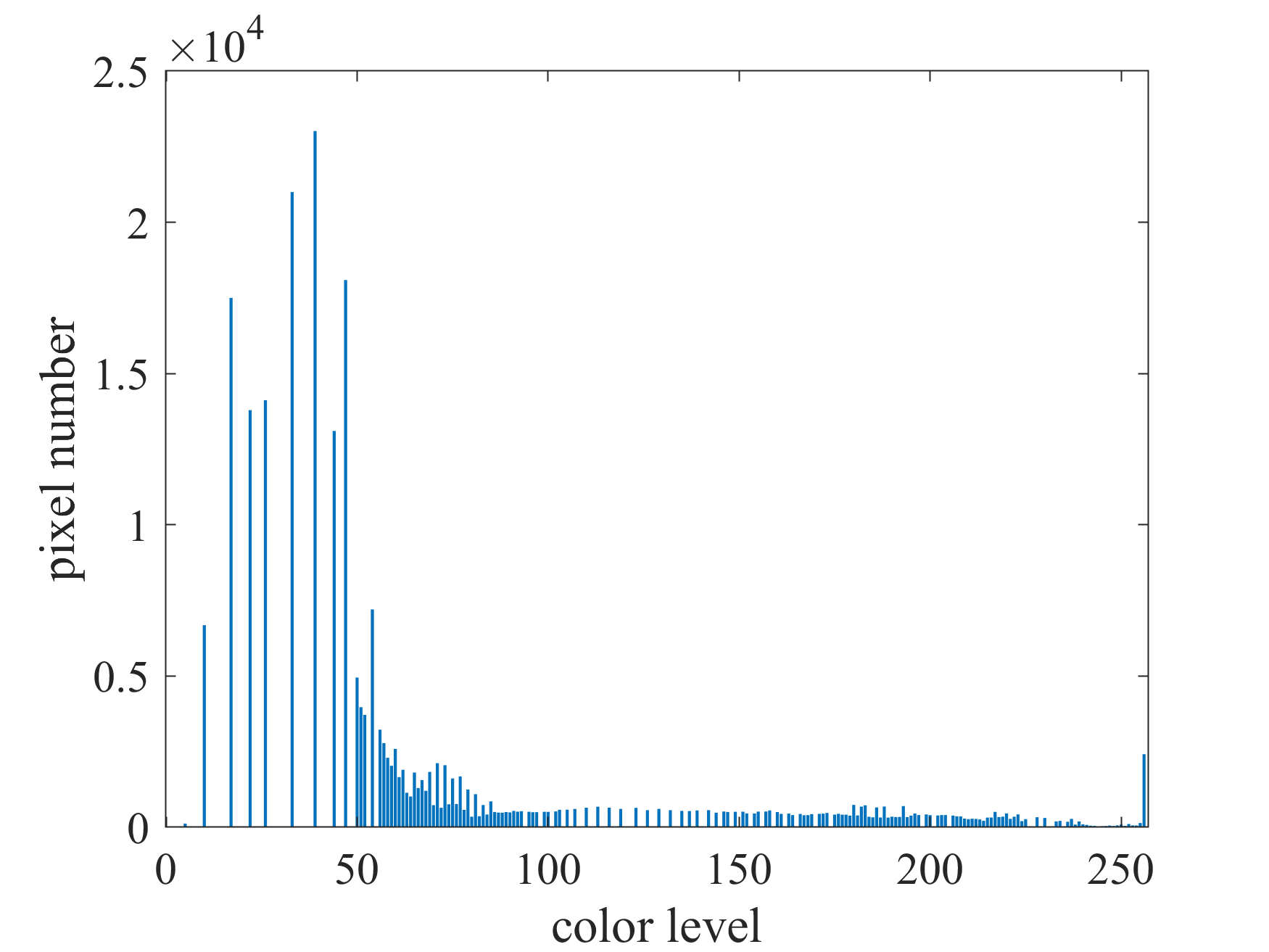}}%
    \subfloat{\includegraphics[width=2.9cm, height=1.8cm]{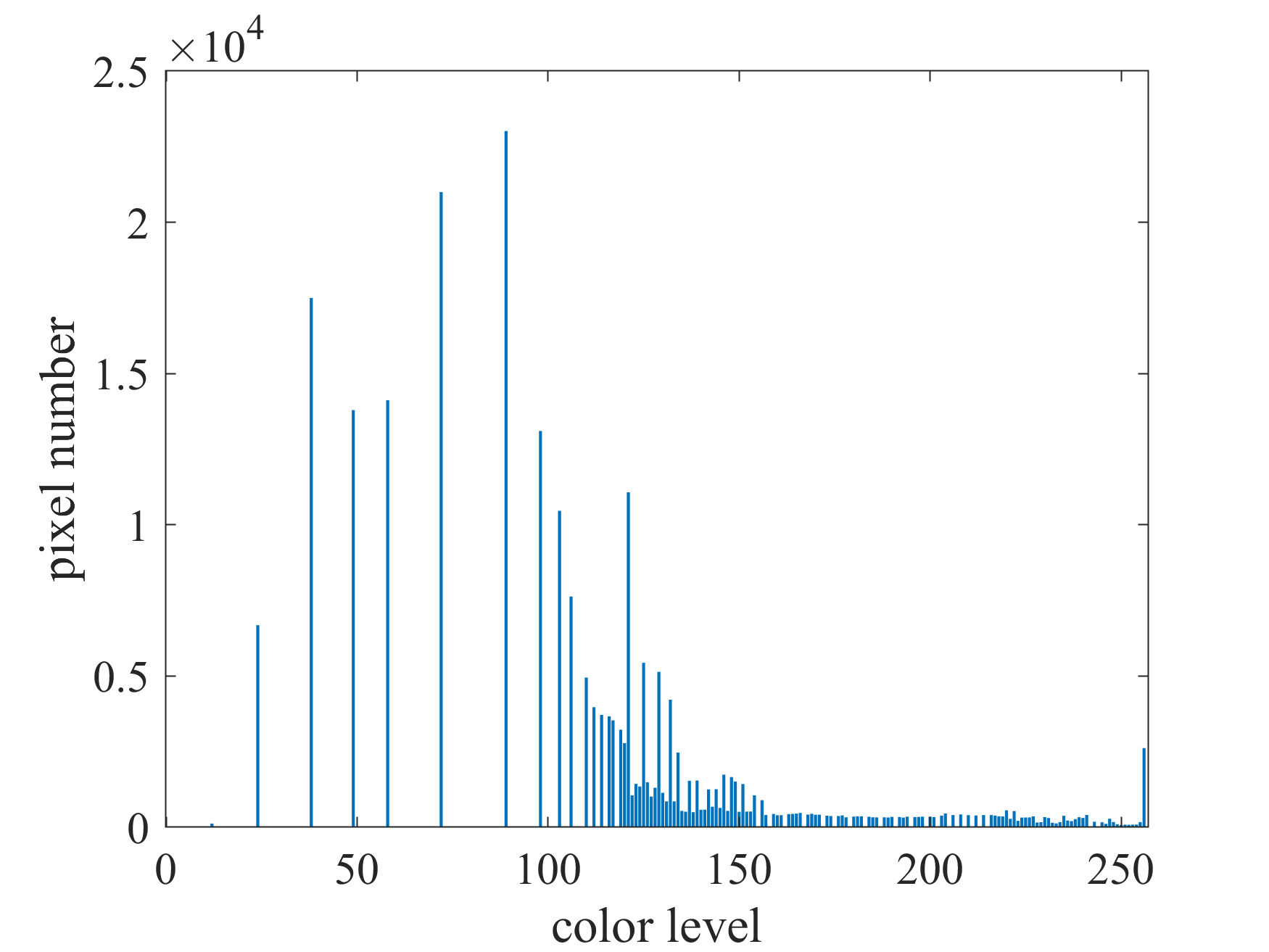}}%
    \subfloat{\includegraphics[width=2.9cm, height=1.8cm]{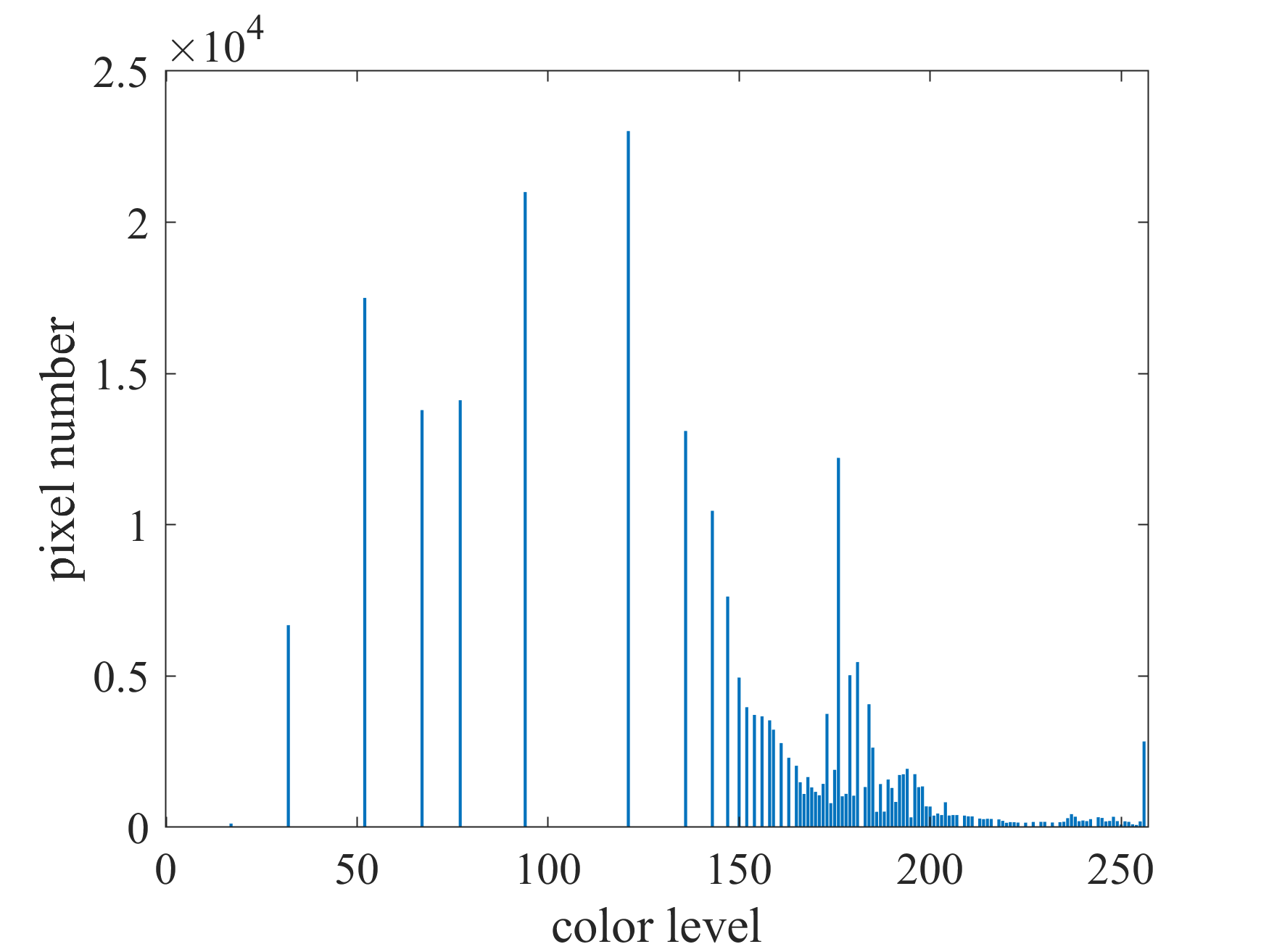}}\\

(a)\hspace{2.5cm}(b)\hspace{2.5cm}(c)\hspace{2.5cm}(d)\hspace{2.5cm}(e)\hspace{2.5cm}(f)
\vspace{-0.2em}
    \caption{Enhancement results and histogram of proposed algorithm with different $\gamma $. (a) low-light image. (b) $\gamma =1$; (c) $\gamma =1.5$; (d) $\gamma =2.2$; (e) $\gamma =5$; (f) $\gamma =10$. *The A1, B1 represent the enhanced results and corresponding histograms, respectively.}
    \label{fig:grid}
\end{figure*}

\begin{table*}[!t]
\caption{Execution Time of Ten Algorithms on ExDark Dataset with Different Resolutions \label{tab mma_table}}
\centering
\begin{tabular}{ccccccccccc}
\toprule
Resolution & LIME & STAR & AHPCE & WVM & SSD & RRM & BRIAN & LR3M & FSTAR & proposed \\
\midrule
336*508 & 1.6228 & 1.7581 & 1.9871 & 5.5222 & 2.6393 & 8.8342 & 6.8111 & 62.7343 & 5.1888 & \textbf{1.5686} \\
400*600 & 2.5349 & 4.8501 & 5.1221 & 9.0034 & 10.1967 & 14.2675 & 6.3718 & 167.7491 & 40.38658 & \textbf{1.6709} \\
1000*664 & 9.7184 & 8.5359 & 18.6841 & 14.6312 & 14.0738 & 27.7620 & 18.9552 & 275.1139 & 97.5191 & \textbf{1.8645} \\
\bottomrule
\end{tabular}
\end{table*}

\begin{table*}[!t]
\caption{Quantitative Results of Ten Algorithms on Image \#3011 with Different Resolutions \label{tab mma_table}}
\centering
\begin{tabular}{cccccccccccc}
\toprule
Size & Index & LIME & STAR & AHPCE & WVM & SSD & RRM & BRIAN & LR3M & FSTAR & proposed \\
\midrule
\multirow{2}{*}{100*100} 
    & Time  & 0.5650 & 0.2522 & \textbf{0.0980} & 0.1712 & 0.3570 & 0.4207 & 1.9906 & 4.3658 & 0.8300 & 1.6990 \\
    & PSNR  & 9.3319 & 14.8228 & 13.8660 & 17.8151 & 14.9319 & 11.0580 & 14.9013 & 15.9416 & 16.7032 & \textbf{17.9581} \\
\multirow{2}{*}{200*200} 
    & Time  & 0.2243 & 0.4356 & 0.8552 & 0.3119 & 0.6964 & 1.4686 & 1.2776 & 13.0972 & 1.8846 & \textbf{2.0995} \\
    & PSNR  & 9.1151 & 14.6911 & 12.9122 & 16.5781 & 14.9458 & 10.9497 & 14.9852 & 15.7205 & 15.9554 & \textbf{17.6312} \\
\multirow{2}{*}{300*300} 
    & Time  & \textbf{0.7465} & 0.9806 & 1.1987 & 0.6325 & 1.4540 & 2.4165 & 2.1059 & 34.1423 & 2.8610 & 1.6570 \\
    & PSNR  & 8.9796 & 14.6564 & 14.0240 & 16.1164 & 14.8840 & 10.8964 & 15.0310 & 15.7036 & 15.4770 & \textbf{17.4694} \\
\multirow{2}{*}{400*400} 
    & Time  & 1.7059 & 1.9736 & 2.7938 & 1.0994 & 3.1236 & 4.6309 & 4.4031 & 57.8065 & 4.7162 & \textbf{1.6803} \\
    & PSNR  & 8.8922 & 14.6377 & 13.9941 & 15.8672 & 14.8285 & 10.8122 & 15.0558 & 15.6269 & 15.2926 & \textbf{17.3868} \\
\multirow{2}{*}{500*500} 
    & Time  & 2.4641 & 2.9236 & 5.3137 & 1.7286 & 4.9744 & 7.6107 & 6.1643 & 75.5910 & 7.3616 & \textbf{1.6733} \\
    & PSNR  & 8.8340 & 14.6250 & 14.8829 & 15.6854 & 14.7836 & 10.8689 & 15.0468 & 15.6253 & 15.2308 & \textbf{17.3355} \\
\multirow{2}{*}{600*600} 
    & Time  & 4.0290 & 4.3649 & 9.7823 & 2.6414 & 7.4948 & 11.7154 & 9.3963 & 113.7967 & 10.3519 & \textbf{1.7376} \\
    & PSNR  & 8.7913 & 14.6171 & 13.9273 & 15.5802 & 14.7451 & 10.8610 & 15.0464 & 15.5801 & 15.1997 & \textbf{17.3005} \\
\multirow{2}{*}{700*700} 
    & Time  & 5.9222 & 5.8624 & 16.6174 & 3.5155 & 10.4483 & 16.6902 & 12.4774 & 157.9600 & 11.7447 & \textbf{1.7012} \\
    & PSNR  & 8.7596 & 14.6116 & 14.0164 & 15.4783 & 14.7177 & 10.8536 & 15.0518 & 15.4970 & 15.0469 & \textbf{17.2749} \\
\multirow{2}{*}{800*800} 
    & Time  & 8.3776 & 7.6438 & 26.3382 & 4.9134 & 13.4235 & 23.2864 & 16.7376 & 203.2019 & 12.5533 & \textbf{1.7344} \\
    & PSNR  & 8.7338 & 14.6007 & 13.6693 & 15.3974 & 14.6904 & 10.8482 & 15.0507 & 15.3872 & 14.9076 & \textbf{17.2545} \\
\multirow{2}{*}{900*900} 
    & Time  & 10.5687 & 9.8826 & 41.4613 & 6.8356 & 17.2951 & 29.6317 & 20.8368 & 260.4267 & 15.6940 & \textbf{1.7486} \\
    & PSNR  & 8.7150 & 14.5978 & 13.9880 & 15.3142 & 14.6673 & 10.8429 & 15.0506 & 15.4662 & 14.8931 & \textbf{17.2389} \\
\multirow{2}{*}{1000*1000} 
    & Time  & 13.5597 & 12.1963 & 58.6440 & 8.1196 & 20.5881 & 43.0656 & 26.1502 & 322.1790 & 16.3733 & \textbf{1.6960} \\
    & PSNR  & 8.6987 & 14.5949 & 13.9820 & 15.2712 & 14.6482 & 10.8392 & 15.0471 & 15.3774 & 14.7559 & \textbf{17.2269} \\
\bottomrule
\end{tabular}
\end{table*}

\subsection{Test and Analysis of Algorithm Complexity}

In this section, we evaluate the execution time of these algorithms on the ExDark dataset, the test image set at different resolutions, including 336*508, 400*600 and 1000*664. The maximum iteration number of these algorithms is set to 10. The execution time of these algorithms on the ExDark dataset with different resolutions are shown in Table IV.

In Table IV, the LIME is faster than other compared algorithms in the low-light images with three resolutions, except the proposed algorithm, because it adopts a speed-up solver to reduce the computational cost. The STAR is also a fast algorithm due to it only introduces the MLV to compute the texture map. The SSD computes the horizontal and vertical direction weight of illumination in each iteration, which increases the computational cost to some extent. The WVM needs to process each channel of the image separately and consume more time. Although the AHPCE is histogram-based algorithm, it calculates the local contrast weighted distribution in the spatial domain and also processes three channels of the image, which consumes more time. The RRM consumes a long time due to it introduces the noise component and processes each channel of image. The MED of the FSTAR involves computing fractional differences in eight symmetric directions for each pixel, which takes a lot of time for the  large size image. The LR3M requires the longest execution time due to its processing of three channels and the need for block matching in each iteration. The proposed algorithm is very fast and consumes similar time for images with different sizes. Because the number of grey levels provided by the histogram is significantly less than the number of pixels in the image, and the difference is even more noticeable for the large-scale images. In addition, we analyze the computational complexity of the proposed algorithm. For each iteration, the histograms of illumination and reflectance require   to finish the estimation. Thus, the computational complexity of the proposed algorithm is  , where the   is the number of iterations to achieve convergence. It is clear that the computational complexity of the proposed algorithm is not related to the pixels number in the image. Compared to variational Retinex algorithms such as LIME and STAR, it has lower computational complexity.

In addition, to further test and analyze the efficiency and performance of the proposed algorithm, we select the image \#3011 from the LSRW dataset, and set different resolutions from 100*100 to 1000*1000. The quantitative results of ten algorithms on these images are shown in Table V.

As can be seen in Table V, all the compared algorithms significantly increase the execution time as the resolution of the test image increases. The execution time of the proposed algorithm is less than 2s in the image with 1000*1000 resolution, and is insensitive to the change in resolution of the image. It is clear that the execution time of the proposed algorithm is not related to the resolution of the low-light image, and demonstrates that the given computational complexity is correct. On the other hand, as the resolution of the image increases, the performance metrics of all algorithms decrease. The proposed algorithm provides the best performance metrics and has minimum metrics loss as the resolution of the image increases. The PSNR of proposed algorithm only loses 0.7312 in image with 100*100 resolution and 1000*1000 resolution. Through this experiment, the proposed algorithm provides better quantitative results with less time.

\section{Conclusion}
In this paper, we propose a novel histogram-based Retinex model for fast low-light image enhancement. The HistRetinex extends the application scenario from the spatial domain to the histogram domain, simplifying the computational complexity and improving efficiency. At first, the histogram location matrix and histogram count matrix are introduced to describe the matrix multiplication between illumination and reflectance components. Secondly, HistRetinex replaces the matrix multiplication in the spatial domain by processing the histogram position matrix and histogram count matrix and realizes the final transformation in the histogram domain. Based on this, utilizing the prior information, HistRetinex constructs a novel objective function, in which prior terms of the histogram are introduced to constrain the illumination and reflectance components. By solving the objective function, the iterative formulas for histograms of illumination and reflectance are given. Finally, we modify the histogram of illumination and provide the enhanced image using histogram matching. Experimental results show that the proposed algorithm outperforms the current state-of-the-art Retinex-based and histogram-based image enhancement algorithms in terms of effectiveness, generalization and execution time. In future work, we plan to reduce the noise of the enhanced image and consider more image intrinsic information.

\newpage

\section{Biography Section}

\begin{IEEEbiography}[{\includegraphics[width=1in,height=1.25in,clip,keepaspectratio]{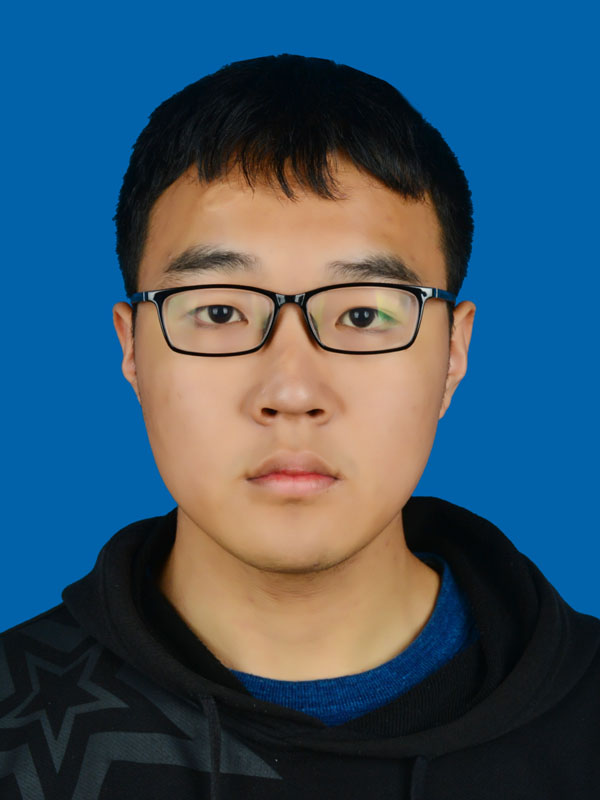}}]{Jingtian  Zhao}
received M.S. degree in Electronic Engineering at Xi’an University of post and telecommunications, Xi’an, China in 2024. He is currently pursuing the Ph.D. degree at College of Missile Engineering, Rocket Force University of Engineering, Xi’an, China. His research interests include digital image processing and intelligent information processing.
\end{IEEEbiography}

\vspace{5pt}

\begin{IEEEbiography}[{\includegraphics[width=1in,height=1.25in,clip,keepaspectratio]{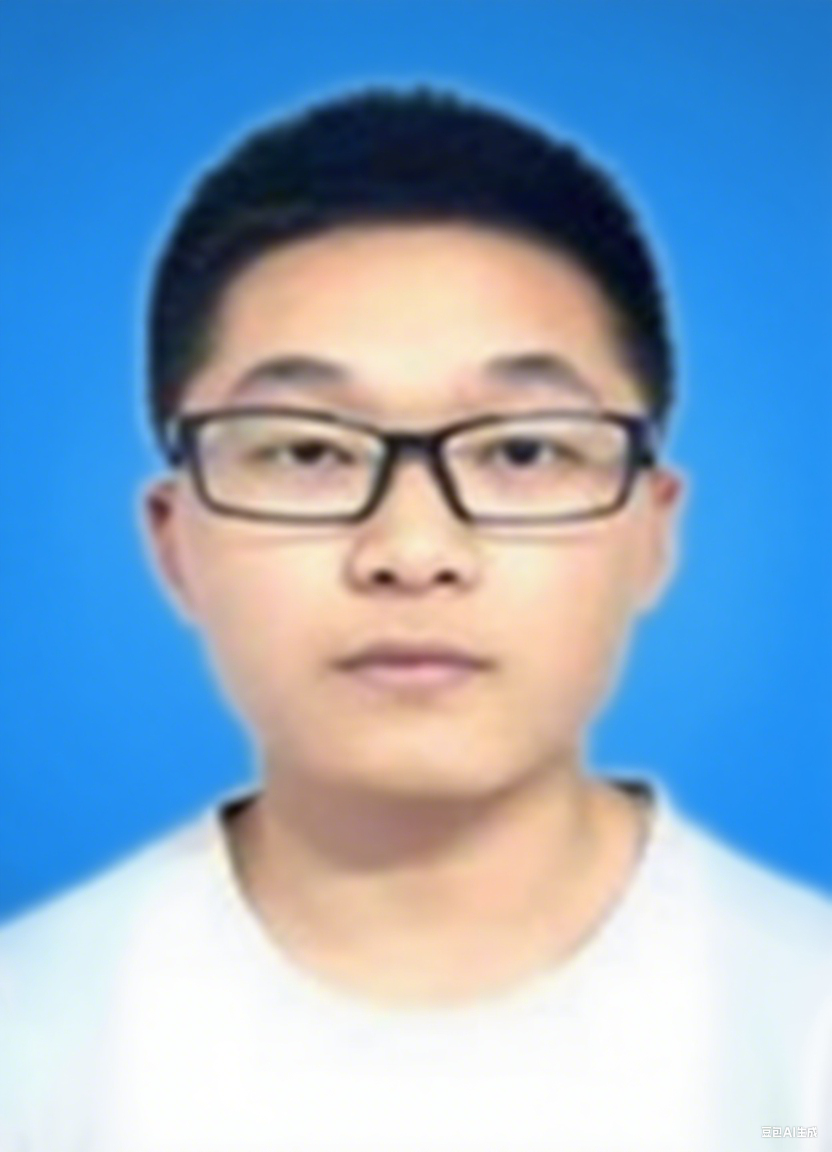}}]{Xueli  Xie}
received the B.E. degree, master’s degree and Ph.D. degree from College of Missile Engineering, Rocket Force University of Engineering, Xi’an, China, in 2024, where he is a teacher in Rocket Force University of Engineering. His research interests include computer vision, deep learning, and object detection.
\end{IEEEbiography}

\vspace{5pt}

\begin{IEEEbiography}[{\includegraphics[width=1in,height=1.25in,clip,keepaspectratio]{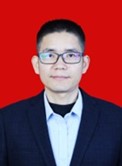}}]{Jianxiang Xi}
received the B.S. and M.S. degrees from High-Tech Institute of Xi’an, China in 2004 and 2007, respectively. He received the Ph.D. degree in control science and engineering from Rocket Force University of Engineering, Xi’an, China in 2012 by a coalition form with Tsinghua University. He is currently a professor at the control science and engineering of Rocket Force University of Engineering, China. His research interests include complex systems control, switched systems and swarm systems.
\end{IEEEbiography}

\vspace{5pt}
\begin{IEEEbiography}[{\includegraphics[width=1in,height=1.25in,clip,keepaspectratio]{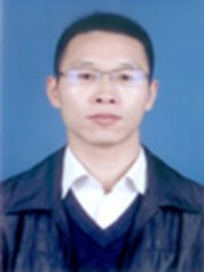}}]{Xiaogang Yang}
was born in Xi’an, Shaanxi, China, in 1978. He received the Ph.D. degree in control science from Rocket Force University of Engineering, Shaanxi, China, in 2006. He is currently a Faculty Member with the Department of Control Engineering, Rocket Force University of Engineering. He is the author of 90 articles and 25 inventions. His research interests include precision guidance and image processing. 
\end{IEEEbiography}

\vspace{5pt}
\begin{IEEEbiography}[{\includegraphics[width=1in,height=1.25in,clip,keepaspectratio]{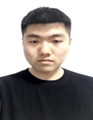}}]{Haoxuan Sun}
was born in Dalian, Liaoning, China, in 1997. He received the master degree in Liaoning University of Technology, Jinzhou, China, in 2024. He is currently pursuing the Ph.D. degree at College of Missile Engineering, Rocket Force University of Engineering, Xi’an, China. His research interests include digital image processing and multi-modal learning. 
\end{IEEEbiography}

\vspace{5pt}

\vfill

\end{document}